%% file: main.tex
\documentclass[preprint,12pt]{elsarticle}

\usepackage{amssymb}
\usepackage{amsmath}
\usepackage{amsthm}
\usepackage{url}
\usepackage{booktabs}
\usepackage{chemformula}
\usepackage[version=3]{mhchem}
\usepackage{longtable}
\usepackage{tabularx}
\usepackage{makecell}
\usepackage{placeins}
\usepackage{pifont}
\usepackage{caption}
\usepackage{hyperref}

\usepackage{graphicx}
\usepackage{chngcntr}
\usepackage{geometry}
\usepackage{rotating}
\usepackage{lineno}

\usepackage{xspace}
\usepackage{amsfonts}
\usepackage{algorithm}
\usepackage{algpseudocode}
\usepackage{subcaption}
\usepackage{booktabs}
\usepackage{longtable}
\usepackage{pdfpages}

\newcommand{\FigurePath}[0]{Figures/}

\newcommand{\speciesindex}{\ensuremath{b}\xspace}
\newcommand{\truncationconst}{\ensuremath{C}\xspace}
\newcommand{\vardim}{\ensuremath{d}\xspace}
\newcommand{\distance}{\ensuremath{\mathbf{d}}\xspace}
\newcommand{\latentfxn}{\ensuremath{f}\xspace}
\newcommand{\latentfxnvec}{\ensuremath{\mathbf{\latentfxn}}\xspace}
\newcommand{\logprobs}{\ensuremath{g}\xspace}
\newcommand{\mechanisticmodel}{\ensuremath{h}\xspace}
\newcommand{\entropy}{\ensuremath{\mathcal{H}}\xspace}
\newcommand{\obsind}{\ensuremath{i}\xspace}
\newcommand{\identity}{\ensuremath{\mathbf{I}}\xspace}
\newcommand{\information}{\ensuremath{\mathcal{I}}\xspace}
\newcommand{\obsindprime}{\ensuremath{j}\xspace}
\newcommand{\taylorterms}{\ensuremath{J}\xspace}
\newcommand{\covfunc}{\ensuremath{k}\xspace}
\newcommand{\covfuncvec}{\ensuremath{\mathbf{\covfunc}\xspace}}
\newcommand{\covfuncmatrixmember}{\ensuremath{K}\xspace}
\newcommand{\covfuncmatrix}{\ensuremath{\mathbf{\covfuncmatrixmember}}\xspace}
\newcommand{\lengthscale}{\ensuremath{\ell}\xspace}
\newcommand{\lengthscalematrix}{\ensuremath{\mathbf{L}}\xspace}
\newcommand{\meanfunc}{\ensuremath{m}\xspace}
\newcommand{\meanfuncvec}{\ensuremath{\mathbf{\meanfunc}}\xspace}
\newcommand{\obsdim}{\ensuremath{n}\xspace}
\newcommand{\latentparamdim}{\ensuremath{p}\xspace}
\newcommand{\pressure}{\ensuremath{P}\xspace}
\newcommand{\taylorpolynomial}{\ensuremath{P}\xspace}
\newcommand{\varinsmechonlydims}{\ensuremath{q}\xspace}
\newcommand{\pwsamples}{\ensuremath{Q}\xspace}
\newcommand{\lagrangeremainder}{\ensuremath{R}\xspace}
\newcommand{\sampleindex}{\ensuremath{s}\xspace}
\newcommand{\nosamples}{\ensuremath{S}\xspace}
\newcommand{\temp}{\ensuremath{T}\xspace}
\newcommand{\usub}{\ensuremath{u}\xspace}
\newcommand{\varinmechonly}{\ensuremath{v}\xspace}
\newcommand{\varinmechonlydom}{\ensuremath{\mathcal{V}}\xspace}
\newcommand{\varinsmechonly}{\ensuremath{\mathbf{\varinmechonly}}\xspace}
\newcommand{\testinind}{\ensuremath{w}\xspace}
\newcommand{\testindim}{\ensuremath{W}\xspace}
\newcommand{\varin}{\ensuremath{x}\xspace}
\newcommand{\varins}{\ensuremath{\mathbf{\varin}}\xspace}
\newcommand{\vardom}{\ensuremath{\mathcal{X}}\xspace}
\newcommand{\observation}{\ensuremath{y}\xspace}
\newcommand{\observations}{\ensuremath{\mathbf{\observation}}\xspace}
\newcommand{\molefrac}{\ensuremath{z}\xspace}
\newcommand{\stdnormrv}{\ensuremath{z}\xspace}

\newcommand{\scalemix}{\ensuremath{\alpha}\xspace}
\newcommand{\actcoeff}{\ensuremath{\gamma}\xspace}
\newcommand{\gammafunc}{\ensuremath{\Gamma}\xspace}
\newcommand{\obserr}{\ensuremath{\varepsilon}\xspace}
\newcommand{\largrangeterm}{\ensuremath{\zeta}\xspace}
\newcommand{\hyperparam}{\ensuremath{\theta}\xspace}
\newcommand{\hyperparams}{\ensuremath{\boldsymbol{\hyperparam}}\xspace}
\newcommand{\hyperparamdom}{\ensuremath{\Theta}\xspace}

\newcommand{\posteriormean}{\ensuremath{\mu}\xspace}
\newcommand{\maternsmooth}{\ensuremath{\nu}\xspace}
\newcommand{\crossoverlap}{\ensuremath{\xi}\xspace}
\newcommand{\obserrstddev}{\ensuremath{\sigma_\obserr}\xspace}
\newcommand{\posteriorstd}{\ensuremath{\sigma}\xspace}
\newcommand{\gpprecision}{\ensuremath{\tau}\xspace}

\newcommand\smallO{
  \mathchoice
    {{\scriptstyle\mathcal{O}}}
    {{\scriptstyle\mathcal{O}}}
    {{\scriptscriptstyle\mathcal{O}}}
    {\scalebox{.7}{$\scriptscriptstyle\mathcal{O}$}}
  }
\DeclareMathOperator{\E}{\mathbb{E}}
\newcommand{\Exp}[2][]{\E_{#1}\!\left[#2\right]}
\newcommand{\reals}{\ensuremath{\mathbb{R}}\xspace}

\renewcommand{\d}{\ensuremath{\text{d}}\xspace}
\newcommand{\diag}{\ensuremath{\text{diag}}\xspace}
\newcommand{\water}{\ensuremath{\text{H}_2\text{O}}\xspace}
\newcommand{\proh}{\ensuremath{\text{PrOH}}\xspace}
\newcommand{\clapeyron}{\texttt{Clapeyron}\xspace}
\newcommand{\python}{\texttt{Python}\xspace}
\newcommand{\julia}{\texttt{Julia}\xspace}
\newcommand{\gpflow}{\texttt{GPFlow}\xspace}
\newcommand{\matern}{Mat\'{e}rn\xspace}

\journal{Elsevier}

\begin{document}

\newproof{pf}{Proof}
\newtheorem{thm}{Theorem}
\newtheorem*{theorem*}{Theorem}
\newtheorem*{prop*}{Proposition}
\newtheorem{remark}{Remark} 

\begin{frontmatter}


\author{Kyla D. Jones}
\author{Alexander W. Dowling\corref{cor1}}
\ead{adowling@nd.edu}
\cortext[cor1]{Corresponding author}
\address{Department of Chemical and Biomolecular Engineering, University of Notre Dame, Notre Dame, IN 46556, USA}

\title{BITS for GAPS: Bayesian Information-Theoretic Sampling for hierarchical GAussian Process Surrogates}

\begin{abstract}
We introduce Bayesian Information-Theoretic Sampling for hierarchical GAussian Process Surrogates (BITS for GAPS), a framework enabling information-theoretic experimental design of Gaussian process-based surrogate models. Unlike standard methods, which use fixed or point-estimated hyperparameters in acquisition functions, our approach propagates hyperparameter uncertainty into the sampling criterion through Bayesian hierarchical modeling. In this framework, a latent function receives a Gaussian process prior, while hyperparameters are assigned additional priors to capture the modeler’s knowledge of the governing physical phenomena. Consequently, the acquisition function incorporates uncertainties from both the latent function and its hyperparameters, ensuring that sampling is guided by both data scarcity and model uncertainty. We further establish theoretical results in this context: a closed-form approximation and a lower bound of the posterior differential entropy.

We demonstrate the framework’s utility for hybrid modeling with a vapor–liquid equilibrium case study. Specifically, we build a surrogate model for latent activity coefficients in a binary mixture. We construct a hybrid model by embedding the surrogate into an extended form of Raoult’s law. This hybrid model then informs distillation design. This case study shows how partial physical knowledge can be translated into a hierarchical Gaussian process surrogate. It also shows that using BITS for GAPS increases expected information gain and predictive accuracy by targeting high-uncertainty regions of the Wilson activity model. Overall, BITS for GAPS is a generalized uncertainty-aware framework for adaptive data acquisition in complex physical systems.
\end{abstract}




\end{frontmatter}


\section{Introduction}

Hybrid modeling is a flexible and expressive paradigm for describing the behavior of complex systems in science and engineering~\cite{psichogios1992, vonstosch2014, sansana2021, agi2024}. A hybrid (i.e., grey-box) model integrates first-principles (white-box) components with data-driven (black-box) elements, combining theoretical knowledge with empirical evidence. By fusing \textit{a priori} information such as conservation laws or heuristics with \textit{a posteriori} data from experiments or simulations, hybrid models provide a flexible representation of system behavior. Consequently, hybrid modeling attracts growing attention across chemical engineering applications~\cite{Narayanan2021, Sitapure2023, POLAK2023, Kay2025}.

The effectiveness of hybrid modeling depends critically on the availability and quality of data. Data acquisition, whether from physical experiments or high-fidelity simulations, is essential for calibrating and validating hybrid models. However, practical constraints such as time, cost, and computational resources often limit data collection~\cite{BIEGLER2014, Ngo2020, kieckhefen2020, Timmerman2024, agbodekhe2025}. These limitations are particularly pronounced in design optimization, multiscale modeling, turbulent combustion, and materials mechanics~\cite{BANERJEE2010}. Developing efficient and principled strategies for data acquisition is therefore essential to advance the practical use of hybrid models~\cite{eyke2020, SCHWEIDTMANN2024, Gargalo2024, barberi2024, DAOUTIDIS2024}. 

A broad range of methodologies supports experiment design in model-driven systems. At one end of this spectrum, optimal design of regression models, or model-based design of experiments (MBDoE), selects sampling points that maximize Fisher information or minimize parameter uncertainty within a known physical model~\cite{asprey2002, franceschini2008, Greenhill2020}. At the other end, Bayesian optimization (BO) performs sequential sampling for purely black-box functions using surrogate models such as Gaussian processes (GPs), exploring the design space without relying on explicit physical structure~\cite{mockus1975, brochu2010, mockus2012, snoek2012, shahriari2016}. MBDoE exploits mechanistic understanding to guide data collection, whereas BO leverages statistical learning and uncertainty quantification to adaptively improve predictions~\cite{wu2024, LIMA2025, shen2025, Carlozo2025}. 

Bridging these paradigms creates an opportunity to combine complementary strengths of model-based experimental design and Bayesian optimization through information-theoretic sequential design of Bayesian hierarchical GPs~\cite{kennedy2001, higdon2008, sauer2021}. In this formulation, priors on GP hyperparameters encode known physical relationships, constraints, or smoothness properties, enabling the surrogate to capture physically meaningful structure while remaining flexible. Sequential design strategies based on maximizing the entropy of the hierarchical GP posterior provide a principled mechanism for targeting uncertainty reduction in both the latent function and its hyperparameters~\cite{MacKay1992, hennig2011}.

Building on this motivation, this work introduces BITS for GAPS: a Bayesian Information-Theoretic Sampling framework for training hierarchical GAussian Process Surrogates. The framework derives a closed-form approximation to the differential entropy of the predictive posterior of a hierarchical GP, enabling practical entropy-guided acquisition. Whereas conventional BO strategies are typically developed for non-hierarchical GP surrogates, BITS for GAPS is formulated specifically for information-theoretic sequential design in hierarchical GP models with physically informed priors.The objective of this work is to provide a principled framework for information-theoretic sequential design when hierarchical modeling and physically informed priors are desired.

The remainder of this paper is organized as follows. Section~\ref{sec: litrev} reviews relevant literature. Section~\ref{sec: hybridmodels} formulates the maximum-entropy sequential design problem. Section~\ref{sec: methods} presents the mathematical foundations of the BITS for GAPS framework and the entropy formulations it employs. Section~\ref{sec: casestudy} demonstrates the approach through a numerical case study on hybrid model-based distillation system design. Finally, Section~\ref{sec: conclusions} summarizes the main findings and discusses implications and directions for future research.

\section{Literature Review}
\label{sec: litrev}

Recent years have seen significant methodological developments in BO and MBDoE across a variety of chemical engineering applications~\cite{jialuwang2022, ke_bo_coce}. Several contributions aim to improve the efficiency and flexibility of BO by incorporating prior knowledge or structure. For instance, \citet{MAHBOUBI2025} propose a transfer learning-based BO framework that integrates mixtures of Gaussians to enhance performance across related tasks. \citet{LEE2024} reduce the computational cost of CFD-based optimization by enveloping BO with historical data. \citet{SAVAGE2024} present a novel approach that integrates human input into high-throughput BO through discrete decision theory.

Another line of research focuses on hybrid or constrained modeling settings. \citet{WINZ2025} develop an upper confidence bound acquisition function tailored to constrained gray-box optimization. \citet{paulson2022} present a constrained expected utility approach combining multivariate GPs. \citet{lu2023} and \citet{LU2025} explore BO in hybrid and inverse optimization contexts, respectively, the latter using BO to estimate unknown parameters from observed decision data. In a similar spirit, \citet{mitsos2023} couple BO with COMSOL simulations using Thompson sampling for efficient multi-objective reactor optimization.

Other studies target hyperparameter tuning and model learning. In recent work, \citet{NGUYEN2025} apply BO for optimizing neural network hyperparameters in molecular property prediction. Similarly, \citet{lee2022} propose a reinforcement learning-based method for multi-step lookahead BO. Finally, \citet{QIU2025} introduce model-inherited trust-region BO for online controller tuning, highlighting BO’s role in adaptive control systems.

There is also growing interest in multi-objective and batch optimization~\cite{Cheng2024, ke2025}. \citet{ye2022} propose a multi-objective prediction framework using BO with eXtreme Gradient Boosting, while \citet{cao2023} apply multi-objective BO to formulation design under ingredient constraints. \citet{misener2023} combine multi-fidelity and asynchronous batch BO to address computational bottlenecks. \citet{zavala2023} introduce a parallel BO method that incorporates expert knowledge by partitioning the design space based on desirable output regions. Finally, and \citet{coutinho2023} develop a systematic method for defining optimization domains in multi-loop PID tuning using BO. 

Beyond BO, Bayesian hierarchical modeling has been increasingly adopted for model calibration problems~\cite{BEHMANESH2016, wu2019, sedehi2020, ping2022, jia2023, ping2023, ping2024}. For example, hierarchical formulations have been used to learn spatially and temporally varying prediction error models, enabling improved uncertainty quantification in settings where model discrepancy is input-dependent rather than homoskedastic. Some representative contributions include hierarchical Bayesian approaches for learning non-stationary prediction errors and model discrepancy fields~\cite{ping2025}, as well as variational inference frameworks that leverage hierarchical structure to update surrogate models under evolving uncertainty conditions~\cite{PING2026}. These works demonstrate that hierarchical Bayesian inference provides a principled mechanism for propagating uncertainty in hyperparameters and latent processes, leading to more expressive predictive distributions than point-estimated alternatives.

While such hierarchical techniques have been successfully applied to uncertainty modeling and model updating, their integration with information-theoretic sequential experimental design remains limited, particularly in chemical engineering contexts. Existing BO and MBDoE approaches typically rely on fixed or point-estimated GP hyperparameters when constructing acquisition functions~\cite{Greenhill2020, ke_bo_coce}, thereby neglecting posterior uncertainty in the surrogate model itself. The present work builds on the hierarchical Bayesian modeling literature by explicitly coupling fully Bayesian GP surrogates with an entropy-based acquisition strategy, enabling sequential design decisions that account for both predictive uncertainty and hyperparameter uncertainty in a unified framework.

To the best of the authors’ knowledge, information-theoretic approaches to BO have seen limited application in chemical engineering contexts, despite their potential to guide data acquisition by directly targeting uncertainty. Moreover, while numerous studies have focused on improving surrogate modeling and acquisition strategies (e.g.,~\citet{paulson2022},~\citet{misener2023},~\citet{zavala2023}), these approaches typically assume fixed or point-estimated GP hyperparameters. In contrast, applications that combine information-theoretic sequential design with fully Bayesian hierarchical GP models, an important framework for incorporating physical insight and parameter uncertainty into black-box models, remain scarce~\cite{jones2023}. The present work addresses this gap by developing an entropy-based acquisition strategy tailored to hierarchical GP surrogates in chemical engineering applications.

\section{Problem Statement}
\label{sec: hybridmodels}
In this work, we focus on serial hybrid models~\cite{psichogios1992, SCHWEIDTMANN2024}, which we define as the function $\mechanisticmodel(\cdot, \cdot)$ with two inputs: a set of known variables $\varinsmechonly \in \varinmechonlydom \subseteq \reals^{\varinsmechonlydims}$, and the output of a function $\latentfxn:\vardom \mapsto \reals$, with inputs $\varins \in \vardom \subseteq \reals^{\vardim}$. The variable domains $\varinmechonlydom$ and $\vardom$ may overlap, be disjoint, or one may be a subset of the other. We encounter such models in settings where part of the system behavior is known and explicitly represented by $\mechanisticmodel(\cdot,\cdot)$, and the remaining unknown or intractable phenomena are captured by the black-box function $\latentfxn(\cdot)$.

To build an accurate surrogate for $\latentfxn(\cdot)$, we assume that the output of $\latentfxn(\cdot)$ is stochastic and adopt a sequential design strategy. At each iteration, we identify the following input $\varins_* \in \vardom$ that most improves the surrogate model according to a defined acquisition criterion. Starting with an initial dataset $\{\varins_1, \dots, \varins_\obsdim\} \subset \vardom$, we iteratively select new inputs, collect data, update the surrogate, and repeat the process until we meet a stopping criterion, such as an experimental budget.
 
We pursue an information-theoretic approach to sequential design. In this context, the acquisition criterion is the statistical information of the surrogate model. For a continuous random variable $\latentfxn(\cdot)$, the differential entropy, $\entropy\{\latentfxn(\cdot)\}$, quantifies statistical information, $\information(\cdot)$. Moreover, information is the negative entropy; $\information(\cdot) := -\entropy\{\latentfxn(\cdot)\}$. For a candidate point $\varins_*$, the differential entropy of the surrogate $\latentfxn(\cdot)$ with probability density function $p(\cdot)$ is
\begin{gather}
    \entropy\{\latentfxn(\varins_*)\} := \mathbb{E}\left[-\log p\{\latentfxn(\varins_*)\}\right]. \label{eq: diffent}
\end{gather}
High entropy corresponds to high uncertainty in the model prediction, indicating regions where new data are most informative. Thus, we seek to solve the acquisition problem:
\begin{gather}
    \max_{\varins_* \in \vardom} \entropy\{\latentfxn(\varins_*)\} = \min_{\varins_* \in \vardom} \information\{\varins_*\}. \label{eq: optimizationprob}
\end{gather}

This paper extends information-theoretic sequential design to cases where $\latentfxn(\cdot)$ is modeled using a Bayesian hierarchical GP. In this formulation, prior physical knowledge such as expected smoothness, surrogate range, or asymptotic behavior is encoded through prior distributions on the model hyperparameters rather than through an explicit parametric form~\cite{Rasmussen2006}. These hierarchical priors bias the surrogate toward physically consistent behavior while retaining flexibility in regimes where the underlying functional relationship is unknown or difficult to interpret directly.

\section{BITS for GAPS Framework}
\label{sec: methods}
\begin{figure}
    \centering
    \includegraphics[width=\linewidth]{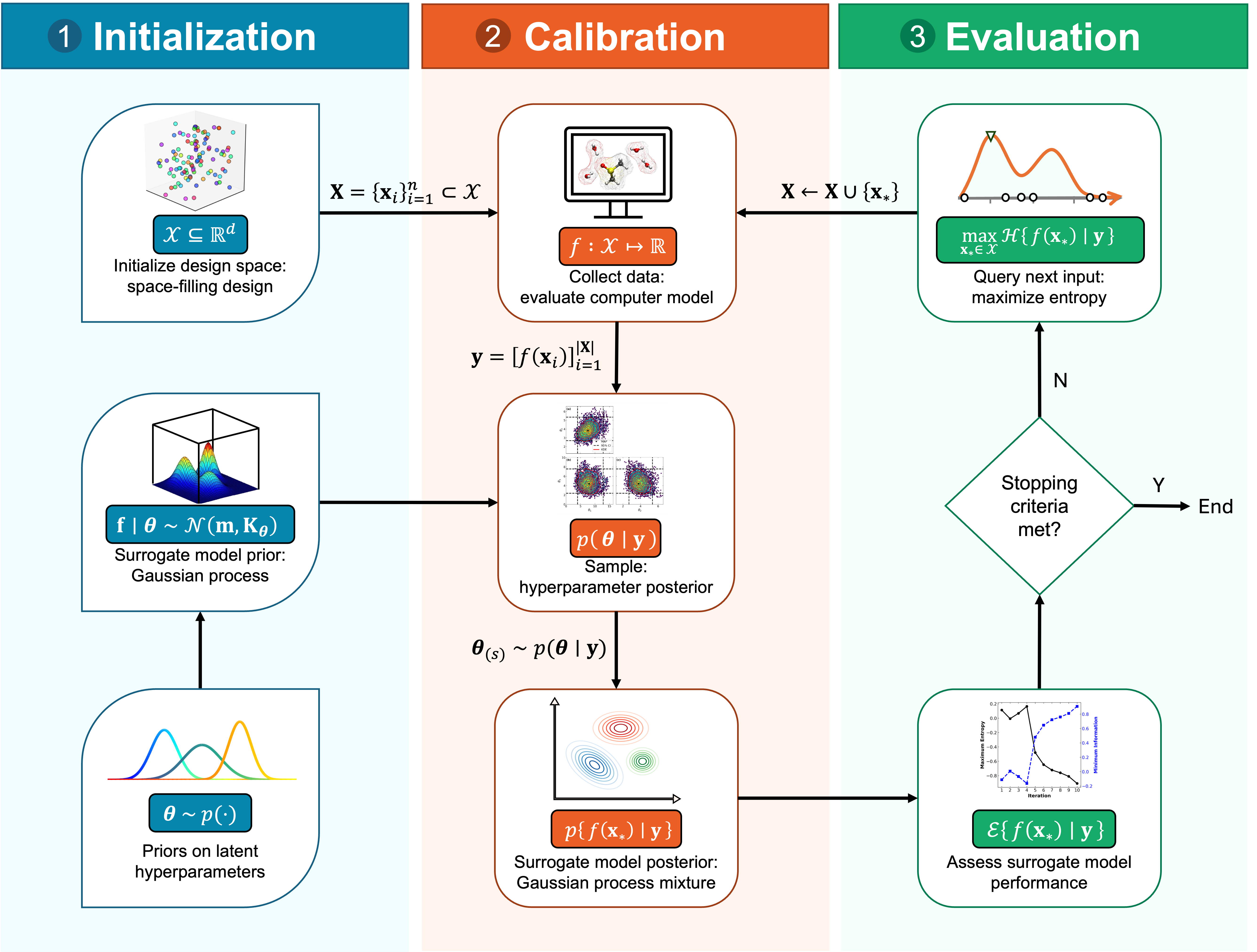}
    \caption{Overview of the BITS for GAPS framework.}
    \label{fig: propframework}
\end{figure}

Figure~\ref{fig: propframework} presents an overview of the proposed BITS for GAPS framework. In the initialization phase (step 1, blue), a space-filling design is used to specify an initial set of input points over the design space. A GP prior is then placed on the (possibly transformed) computer model output, along with prior distributions on the GP hyperparameters.

In the calibration phase (step 2, orange), we evaluate the computer model or experiment at the selected design points and use Markov chain Monte Carlo (MCMC) to sample from the posterior distribution of the hyperparameters. We then propagate a subset of the posterior samples through the hierarchical GP posterior to obtain the surrogate model output.

In the evaluation phase (step 3, green), we assess the quality of the surrogate model with a user-defined metric and select new input points by maximizing the posterior differential entropy. We use the new data to update the surrogate model and repeat this process until we reach a predefined stopping criterion.

\FloatBarrier
\subsection{Bayesian Inference}
Let $\observation \in \mathbb{R}$ denote a single observation. Let $\hyperparam$ be an unknown parameter governing the distribution of $\observation$, i.e., $\observation \sim p(\observation \mid \hyperparam)$. More generally, $\hyperparams$ may represent a collection of parameters $\hyperparams = (\hyperparam_1, \ldots, \hyperparam_{\latentparamdim})^\top \in \hyperparamdom \subseteq \mathbb{R}^{\latentparamdim}$. Let $\observations = (\observation_1, \ldots, \observation_{\obsdim})^\top \in \reals^\obsdim$ denote a sample of $\obsdim$ independent observations.

The prior distribution $p(\hyperparams)$ encodes our beliefs about the parameters $\hyperparams$ before observing any data. The sampling distribution, or likelihood, is the distribution of the observed data given the parameters, i.e., $p(\observations \mid \hyperparams)$. This captures how the data are generated under the assumed model.

To connect the model with the data without conditioning on specific parameter values, we introduce the marginal likelihood, or evidence, given by
\begin{gather*}
    p(\observations) = \int p(\observations \mid \hyperparams)\, p(\hyperparams)\, \d \hyperparams.
\end{gather*}
The evidence measures the overall agreement between the observed data and prior assumptions. A value of $p(\observations) = 0$ would indicate complete inconsistency between prior knowledge and data, rendering Bayesian inference undefined.

The goal of Bayesian inference is to update our beliefs about the parameters in light of the data. This update is formalized by Bayes' rule, which yields the posterior distribution:
\begin{gather}
    p(\hyperparams \mid \observations) = \frac{p(\observations \mid \hyperparams)\, p(\hyperparams)}{p(\observations)}. \label{eq: bayes}
\end{gather}
The posterior quantifies the uncertainty about $\hyperparams$ after observing the data, and forms the core object of interest in Bayesian inference.

\subsection{Markov Chain Monte Carlo}
This section provides a high-level overview of MCMC methods to motivate the selection of a sampling algorithm and support interpretation of results. For comprehensive treatments, see~\citet{mackay2003}. We emphasize that the BITS for GAPS framework is agnostic to the choice of MCMC algorithm.

Monte Carlo (MC) methods are a class of computational techniques that rely on random sampling. MC methods are widely used to (i) generate samples from a target probability distribution and (ii) estimate expectations of functions under that distribution. These tasks are particularly common in Bayesian inference, where the posterior distribution (Eq.~\eqref{eq: bayes}) is often analytically intractable or difficult to sample from directly~\cite{wasserman2004}. 

MCMC works by constructing a Markov chain whose stationary distribution matches the target distribution~\cite{Brooks2011}. Each sample depends only on the previous one, satisfying the Markov property. Key theoretical foundations include stationarity, convergence (typically asymptotic), and often reversibility via detailed balance, which ensures correctness of the sampling process in the limit.

Notable MCMC algorithms include Metropolis-Hastings~\cite{metropolis,hastings}, Gibbs sampling~\cite{Geman,Gelfand}, Hamiltonian MC (HMC)~\cite{DUANE1987,Neal1996,stanpaper}, the DiffeRential Evolution Adaptive Metropolis (DREAM) algorithm~\cite{dream}, and the No-U-Turn Sampler (NUTS)~\cite{nuts}. In practice, MCMC chains may require many iterations to converge, and poor choices of algorithm or tuning parameters can lead to slow mixing, inefficient exploration, or biased estimates. While Metropolis-Hastings and Gibbs sampling are widely used due to their simplicity and generality, they often suffer from slow exploration due to random walk behavior~\cite{mackay2003}.

HMC addresses limitations of random walk exploration by incorporating gradient information to guide proposals. By introducing auxiliary momentum variables and simulating Hamiltonian dynamics, HMC generates proposals that traverse parameter space more efficiently, often resulting in higher acceptance rates and lower autocorrelation. These properties make HMC particularly well-suited to high-dimensional posteriors or those with strong curvature.

However, the performance of HMC depends on the choice of hyperparameters, especially the integration step size and number of leapfrog steps. Poorly chosen values can lead to unstable simulations or inefficient exploration. Adaptive step-size schemes, which adjust the step size during a warm-up phase to maintain a target acceptance rate~\cite{andrieu2008,hoffman2020}, help mitigate this sensitivity and improve sampler robustness.

\subsection{Gaussian Processes}
A stochastic process is a collection of random variables indexed by time, space, or another domain. A GP is a particular type of stochastic process where any finite subset of these random variables follows a multivariate normal distribution. More formally, a GP defines a joint distribution over an infinite collection of random variables, thereby serving as a natural extension of multivariate Gaussian distributions to function spaces~\cite{Rasmussen2006, Gramacy2020Surrogates:Sciences}.

Let $\latentfxn(\cdot)$ denote a random function such that $\{\latentfxn(\varins) \mid \varins \in \vardom\}$ follows a GP, where $\vardom$ is the indexing set (e.g., time or spatial domain). A GP is fully specified by its mean and covariance function:
\begin{align*}
    \meanfunc(\varins) &:=\E{[\latentfxn(\varins)]}, \\
    \covfunc(\varins, \varins') &:= \E{[\{\latentfxn(\varins) - \meanfunc(\varins)\}\{\latentfxn(\varins') - \meanfunc(\varins')\}]}.
\end{align*}
In vector notation, for a finite collection of inputs $\{\varins_1, \ldots, \varins_\obsdim\}$, the corresponding random variables $\latentfxnvec = [\latentfxn(\varins_1), \ldots, \latentfxn(\varins_\obsdim)]^\top$ follow a multivariate normal distribution:
\begin{gather}
    \latentfxnvec \sim \mathcal{N}(\meanfuncvec, \covfuncmatrix), \label{eq: gpprior}
\end{gather}
where $\meanfuncvec = [\meanfunc(\varins_1), \ldots, \meanfunc(\varins_\obsdim)]^\top$ and
\begin{gather*}
    \covfuncmatrixmember_{\obsind, \obsindprime} = \covfunc(\varins_\obsind, \varins_\obsindprime), \quad \forall \, (\obsind, \obsindprime) \in \{1,\ldots, \obsdim\}^2.
\end{gather*}

\subsection{Gaussian Process Regression}
In Bayesian nonparametric statistics, GPs are commonly employed as priors for modeling real-valued latent functions. Consider the following regression model:
\begin{gather*}
    \observation_\obsind = \latentfxn(\varins_\obsind) + \obserr_\obsind, \quad \obsind=1,\ldots,\obsdim,
\end{gather*}
where $\obserr_\obsind$ represents measurement noise. For simplicity, we assume the errors are independent and identically distributed (i.i.d.) Gaussian with a mean of zero and constant variance:
\begin{gather*}
    \obserr_1, \ldots, \obserr_\obsdim \stackrel{\text{i.i.d.}}{\sim} \mathcal{N}(0, \obserrstddev^2).
\end{gather*}
Note that the assumption of homoskedasticity is not generally required, but is adopted here for simplicity.

To infer the latent function $\latentfxn(\cdot)$ from the observed data $\{(\varins_\obsind, \observation_\obsind)\}_{\obsind=1}^\obsdim$, we place a GP prior on $\latentfxn(\cdot)$ as specified in Eq.~\eqref{eq: gpprior}. Thus, in the Bayesian sense, $\latentfxnvec$ can be thought of as a parameter subject to a multivariate Gaussian prior. Moreover, the modeler selects the mean and covariance functions based on prior beliefs about the latent function, which we cover in more detail in Section~\ref{subsec: meanandcovselect}.

Since the observations are corrupted by Gaussian noise, the vector of observations \observations also follows a multivariate normal distribution:
\begin{gather*}
    \observations \mid \latentfxnvec \sim \mathcal{N}\left(\latentfxnvec,\obserrstddev^2\, \identity \right),
\end{gather*}
where \identity is the $\obsdim \times \obsdim$ identity matrix.

An important property of GPs is that any finite set of latent function values, including both observed and unobserved inputs, forms a joint Gaussian distribution. In particular, the latent function evaluated at a new input $\varins_*$ is jointly Gaussian with the noisy observations:
\begin{gather*}
    \begin{bmatrix}
        \observations \\ \latentfxn(\varins_*)
    \end{bmatrix}
    \sim \mathcal{N}\left(
    \begin{bmatrix}
        \meanfuncvec \\ \meanfunc(\varins_*)
    \end{bmatrix},
    \begin{bmatrix}
        \covfuncmatrix + \obserrstddev^2\,\identity & \covfuncvec_* \\
        \covfuncvec_*^\top & \covfunc(\varins_*, \varins_*)
    \end{bmatrix}
    \right),
\end{gather*}
where
\begin{gather*}
    \covfuncvec_* = [\covfunc(\varins_1, \varins_*), \ldots, \covfunc(\varins_\obsdim, \varins_*)]^\top.
\end{gather*}

By conditioning on the observations, the predictive distribution for $\latentfxn(\varins_*)$ is Gaussian:
\begin{gather*}
    \latentfxn(\varins_*) \mid \observations \sim \mathcal{N}\{ \posteriormean(\varins_*), \posteriorstd^2(\varins_*) \},
\end{gather*}
where
\begin{subequations}
    \begin{align}
        \posteriormean(\varins_*) &= \meanfunc(\varins_*) + \covfuncvec_*^\top \left( \covfuncmatrix + \obserrstddev^2\, \identity \right)^{-1} \left( \observations - \meanfuncvec \right), \label{eq: gppredmean}\\
        \posteriorstd^2(\varins_*) &= \covfunc(\varins_*, \varins_*) - \covfuncvec_*^\top \left( \covfuncmatrix + \obserrstddev^2\, \identity \right)^{-1} \covfuncvec_*. \label{eq: gppredcov}
    \end{align}
    \label{eq: predictivegp}
\end{subequations}
Eqs.~\eqref{eq: gppredmean}) and~\eqref{eq: gppredcov} are the standard expressions for the predictive mean and variance, respectively~\cite{Rasmussen2006}.
\subsection{Mean \& Covariance Specification}
\label{subsec: meanandcovselect}
When using GPs for regression, prior beliefs (or lack thereof) about the latent function are introduced through the specification of the mean and covariance functions. The mean function encodes assumptions about the function's expected value in the absence of observations, thereby establishing the prior baseline of the GP. The covariance function, on the other hand, determines the shape, smoothness, generalization behavior, and prior assumptions of the GP.

This section discusses the selection of stationary kernel functions commonly used in practice to model the GP's covariance. For a broader discussion on classes of kernels, see~\citet{Genton2002}.

Mathematically, a kernel defines a measure of similarity between input points. In the GP framework, a kernel is a symmetric, positive-definite function that governs the structure of the covariance matrix. A classic example is the squared exponential (SE), or Gaussian kernel:
\begin{gather}
     \covfunc_{\text{SE}}(\varins_\obsind, \varins_\obsindprime) = \gpprecision^{-1} \exp \left(-\frac{1}{2}\, \distance^\top \,\boldsymbol{\lengthscalematrix}^{-1} \,\distance \right), \label{eq: sekernel}
 \end{gather}
where $\distance = \varins_\obsind -\varins_\obsindprime$ is the input difference, $\gpprecision$ denotes the process precision, and $\lengthscalematrix \in \reals^{\vardim \times \vardim}$ encodes the characteristic length scales that influence smoothness. The SE kernel models functions that are infinitely differentiable, making it a popular choice for capturing smooth behavior.

The structure of $\lengthscalematrix$ can be adapted to impose further assumptions on the smoothness of the underlying process. These choices are elaborated at the end of this section.

A generalization of Eq.~\eqref{eq: sekernel} is the rational quadratic (RQ) kernel:
\begin{gather*}
     \covfunc_{\text{RQ}}(\varins_\obsind, \varins_\obsindprime) = \gpprecision^{-1} \left(1 +\frac{1}{2 \scalemix} \,\distance^\top \,\lengthscalematrix^{-1} \,\distance \right)^{-\scalemix}.
\end{gather*}
The RQ kernel introduces an additional parameter $\scalemix$ that controls the distribution of length scales. As $\scalemix \rightarrow \infty$, the RQ kernel converges to the SE kernel (Eq.~\eqref{eq: sekernel})~\cite{xiuyong2022}. This flexibility allows the RQ kernel to better model functions that vary across multiple characteristic length scales.

Another widely used covariance function is the \matern kernel, defined as
\begin{gather*}
     \covfunc_{\text{\matern}}
     (\varins_\obsind, \varins_\obsindprime) = \gpprecision^{-1}\,\frac{2^{1-\maternsmooth}}{\gammafunc(\maternsmooth)} \, \left(\sqrt{2\maternsmooth \, \distance^\top \,\lengthscalematrix^{-1} \,\distance}\, \right)^{\maternsmooth}\, K_\maternsmooth  \left(\sqrt{2\maternsmooth \, \distance^\top \,\lengthscalematrix^{-1} \,\distance}\, \right).
 \end{gather*}
Here, $\maternsmooth$ is a smoothness parameter, $\gammafunc(\cdot)$ is the Gamma function, and $K_\maternsmooth(\cdot)$ denotes the modified Bessel function of the second kind. Similar to the RQ kernel, the \matern kernel generalizes the SE kernel and approaches it as $\maternsmooth \rightarrow \infty$~\cite{porcu2023}. However, unlike the SE kernel, which assumes infinite differentiability, the \matern kernel enables control over the function’s smoothness through $\maternsmooth$. A GP with \matern covariance is $\maternsmooth - 1$ times differentiable in the mean-square sense. This property makes it especially useful for applications where the smoothness level is unknown or expected to vary. Standard choices for $\maternsmooth$ in GP modeling include $1/2$, $3/2$, and $5/2$~\cite{Rasmussen2006}.

The choice of $\lengthscalematrix$ plays a crucial role in determining how prior knowledge is encoded in the GP. In the isotropic case, a single length scale is applied across all input dimensions, such that $\lengthscalematrix = \lengthscale \, \identity$. This formulation assumes equal importance and effect of each input feature. When dimension-specific behavior is expected, one may adopt anisotropic kernels that support automatic relevance determination (ARD). Under ARD, the length scale matrix is diagonal, i.e., $\lengthscalematrix = \diag(\lengthscale_1,\dots,\lengthscale_{\vardim})$, allowing each input dimension to be scaled independently. This captures the varying influence of individual features on the model’s predictions. 

To unify notation, we represent all kernel hyperparameters as a vector $\hyperparams = (\hyperparam_1,\dots,\hyperparam_\latentparamdim)^\top \in \hyperparamdom \subseteq \reals^\latentparamdim$.

\subsection{Bayesian Hierarchical Gaussian Process Regression}
We now generalize GP regression to settings where priors are placed on the hyperparameters $\hyperparams$ of the GP kernel. This leads to the hierarchical Bayesian model:
\begin{align*}
    & \hyperparams \sim p(\cdot),\\
    & \latentfxnvec \mid \hyperparams \sim \mathcal{N}(\meanfuncvec, \covfuncmatrix),\\
    & \observations \mid \latentfxnvec \sim \mathcal{N}(\latentfxnvec, \obserrstddev^2 \, \identity).
\end{align*}
We are interested in obtaining the predictive distribution for a new input $\varins_*$, obtained by marginalizing over the joint posterior:
\begin{align*}
    p\{\latentfxn(\varins_*) \mid \observations\} &= \iint p\{\latentfxn(\varins_*) \mid \latentfxnvec, \hyperparams\} \, p(\latentfxnvec, \hyperparams \mid \observations) \, \d\latentfxnvec \, \d\hyperparams\\
    &= \iint p\{\latentfxn(\varins_*) \mid \latentfxnvec, \hyperparams\} \, p(\latentfxnvec \mid \observations, \hyperparams) \, p(\hyperparams \mid \observations) \, \d\latentfxnvec \, \d\hyperparams.
\end{align*}
The inner integral yields:
\begin{gather*}
    p\{\latentfxn(\varins_*)\mid \observations, \hyperparams\} = \int p\{\latentfxn(\varins_*)\mid \latentfxnvec, \hyperparams\} \, p(\latentfxnvec \mid \observations, \hyperparams) \, \d \latentfxnvec,
\end{gather*}
which is the standard GP predictive distribution with mean and variance given in Eq.~\eqref{eq: predictivegp} at fixed hyperparameters $\hyperparams$ under the Gaussian noise setting. 

The hierarchical predictive posterior can be written as an expectation with respect to the hyperparameter posterior:
\begin{gather}
    p\{\latentfxn(\varins_*) \mid \observations\} = \int p\{\latentfxn(\varins_*) \mid \observations, \hyperparams\} \, p(\hyperparams \mid \observations) \, \d\hyperparams = \Exp{p\{\latentfxn(\varins_*) \mid \observations, \hyperparams\}}. \label{eq: posteriortoapprox}
\end{gather}
Although the integral in Eq.~\eqref{eq: posteriortoapprox} is analytically intractable in general, it reveals an important structural property: the hierarchical predictive posterior can be interpreted as a mixture distribution over the conditional GP posteriors indexed by the hyperparameters.

This representation suggests a natural Monte Carlo approximation. Drawing samples $\{\hyperparams^{(\sampleindex)}\}_{\sampleindex=1}^\nosamples$ from the hyperparameter posterior $p(\hyperparams\mid \observations)$, we approximate
\begin{gather}
    p\{\latentfxn(\varins_*) \mid \observations\} \simeq  \frac{1}{\nosamples} \sum_{\sampleindex=1}^\nosamples p\{\latentfxn(\varins_*) \mid \observations, \hyperparams^{(\sampleindex)}\}. \label{eq: gmmposteriordist}
\end{gather}
\begin{remark}
    Under standard assumptions on the sampling procedure (e.g., i.i.d. sampling or stationary, ergodic Markov chain with invariant distribution $p(\hyperparams\mid \observations)$), the pointwise Monte Carlo estimator in Eq.~\eqref{eq: gmmposteriordist} converges almost surely to the true hierarchical posterior:
\begin{gather*}
    \frac{1}{\nosamples} \sum_{\sampleindex=1}^\nosamples p\{ \latentfxn(\varins_*)\mid \observations, \hyperparams^{(\sampleindex)}\} \xrightarrow{a.s.} p\{\latentfxn(\varins_*) \mid \observations\}.
\end{gather*}
\end{remark}

As noted by~\citet{lalchand2020}, the approximation in Eq.~\eqref{eq: gmmposteriordist} implies the predictive posterior takes the form of a finite Gaussian mixture model (GMM), where each component corresponds to a conditional GP predictive distribution evaluated at a sampled hyperparameter setting $\hyperparams^{(\sampleindex)}$. Specifically, for the \sampleindex-th component,
\begin{gather*}
    \latentfxn(\varins_*) \mid \observations, \hyperparams^{(\sampleindex)} \sim \mathcal{N}\left\{\posteriormean_{\sampleindex}(\varins_*), \posteriorstd^2_{\sampleindex}(\varins_*)\right\}.    
\end{gather*}
Using standard results on moments of GMMs, the first and second moments of the predictive posterior are~\citet{pereira2022}:
\begin{subequations}
    \begin{align}
    &\E{[\latentfxn(\varins_*)\mid \observations]} = \posteriormean(\varins_*) = \frac{1}{\nosamples} \sum_{\sampleindex=1}^\nosamples \posteriormean_{\sampleindex}(\varins_*),\\
    &\E{[\left\{ \latentfxn(\varins_*) \mid \observations - \posteriormean(\varins_*)\right\}^2]} = \frac{1}{\nosamples} \sum_{\sampleindex=1}^\nosamples  \posteriorstd^2_{\sampleindex}(\varins_*) + \frac{1}{\nosamples} \sum_{\sampleindex=1}^\nosamples \left\{ \posteriormean_{\sampleindex}(\varins_*) - \posteriormean(\varins_*) \right\}^2. \label{eq: gmmvar}
\end{align}
\end{subequations}

\subsection{Approximate Inference of Credible Regions}
\label{subsection: approxinf}
The hierarchical predictive distribution (Eq.~\eqref{eq: gmmposteriordist}) is a GMM. Since GMMs do not admit closed-form expressions for their quantiles, the variance expression in Eq.~\eqref{eq: gmmvar} cannot be applied. Consequently, credible regions of the predictive posterior must be estimated empirically, following the approach of~\citet{lalchand2020}, as illustrated below. Because the predictive posterior is asymptotically equivalent, the posterior credible regions derived from the empirical predictive distribution converge to their true counterparts under mild regularity conditions.

In this section, we refer to the hierarchical posterior $\latentfxn(\varins_*) \mid \observations$ as $\latentfxn(\varins_*)$. That is, we drop conditioning of the observations for convenience.

\begin{algorithm}
\caption*{Pointwise Credible Region for Hierarchical Gaussian Process Predictive Posterior}
\begin{algorithmic}
    \State \textbf{Given:} \testindim test inputs $\{\varins_{*_\testinind}\}_{\testinind=1}^\testindim$
    \State \textbf{for} $\testinind = 1, \dots, \testindim$:
        \State \hspace{1em} Draw \pwsamples samples from the univariate hierarchical posterior:
        \[
        \sampleindex \sim \mathcal{U}(1,\nosamples), \quad \stdnormrv\sim\mathcal{N}(0,1)\]
        \[\latentfxn(\varins_{*_\testinind}) = \posteriormean_{\sampleindex}(\varins_{*_\testinind}) + \posteriorstd_\sampleindex(\varins_{*_\testinind}) \,\stdnormrv
        \]
        \State \hspace{1em} Sort samples in ascending order 
        $\latentfxn(\varins_{*_\testinind})^{(1)} \leq \hdots \leq \latentfxn(\varins_{*_\testinind})^{(\pwsamples)}$
        \State \hspace{1em} Extract $\alpha/2\times100^{th}$ percentile 
        $\Rightarrow \latentfxn(\varins_{*_\testinind})^{(\ell)}$ where 
        $\ell = \left[ \frac{\alpha}{2} \times \pwsamples \right]$
        \State \hspace{1em} Extract $(1-\alpha/2)\times100^{th}$ percentile 
        $\Rightarrow \latentfxn(\varins_{*_\testinind})^{(u)}$ where 
        $u = \left[ \left(1 - \frac{\alpha}{2} \right) \times \pwsamples \right]$
    \State \textbf{return} 
    \State \hspace{1em} $\latentfxnvec_*^{(\ell)} = \{ \latentfxn(\varins_{*_\testinind})^{(\ell)}\}_{\testinind=1}^\testindim$
    \State \hspace{1em} $\latentfxnvec_*^{(u)} = \{\latentfxn(\varins_{*_\testinind})^{(u)}\}_{\testinind=1}^\testindim$
\end{algorithmic}
\end{algorithm}
\FloatBarrier

\subsection{Approximation of the Hierarchical Gaussian Process Posterior Entropy}
Let $\latentfxn_*:= \latentfxn(\varins_*)\mid \observations$ denote the predictive random variable at input $\varins_*$. The differential entropy of the hierarchical GP posterior is
\begin{gather}
    \entropy(\latentfxn_*) = -\int p(\latentfxn_*) \log p(\latentfxn_*) \, \d\latentfxn_*. \label{eq: gppostentropy}
\end{gather}
Let the predictive posterior be approximated as a uniformly weighted mixture of \nosamples Gaussian components (Eq.~\eqref{eq: gmmposteriordist}),
\begin{gather*}
    p(\latentfxn_*) \simeq \frac{1}{\nosamples} \sum_{\sampleindex=1}^\nosamples p_\sampleindex(\latentfxn_*), \quad p_\sampleindex(\latentfxn_*) = p(\latentfxn_* \mid \hyperparams^{(\sampleindex)}).
\end{gather*}
Then the entropy can be expressed as
\begin{gather*}
    \entropy(\latentfxn_*) \simeq -\frac{1}{\nosamples} \sum_{\sampleindex=1}^\nosamples \int p_\sampleindex(\latentfxn_*) \log p(\latentfxn_*) \,\d \latentfxn_*.
\end{gather*}

\begin{remark}
    Regarding the impact of Monte Carlo approximation on the differential entropy, the entropy is a deterministic functional of the predictive posterior distribution. In this work, the predictive posterior is approximated as a finite GMM with well-behaved component means and variances. As the number of posterior samples \nosamples increases, the Monte Carlo approximation of the predictive distribution converges almost surely to the exact posterior predictive distribution. Since the entropy depends smoothly on the predictive density, this implies that the corresponding entropy estimate also converges almost surely.
\end{remark}

The logarithm in Eq.~\eqref{eq: gppostentropy} acts on a finite sum of Gaussian component densities, yielding a log-sum expression that admits no closed form expression. Following~\citet{huber2008}, we approximate the log-density by a Taylor expansion around the predictive mean of each mixture component. Let $\posteriormean_\sampleindex:= \posteriormean_\sampleindex(\varins_*)$ and $\posteriorstd^2_\sampleindex:= \posteriorstd^2_\sampleindex(\varins_*)$. Define $\logprobs(\latentfxn_*):=\log p(\latentfxn_*)$. By Taylor's theorem, for expansion around $\posteriormean_\sampleindex$,
\begin{gather*}
    \logprobs(\latentfxn_*) = \taylorpolynomial_\taylorterms(\latentfxn_*) + \lagrangeremainder_\taylorterms(\latentfxn_*),
\end{gather*}
where $\taylorpolynomial_\taylorterms(\cdot)$ is the $\taylorterms^{th}$ order Taylor polynomial
\begin{gather*}
    \taylorpolynomial_\taylorterms(\latentfxn_*) = \logprobs(\posteriormean_\sampleindex) + \logprobs'(\posteriormean_\sampleindex)(\latentfxn_* - \posteriormean_\sampleindex) + \frac{\logprobs''(\posteriormean_\sampleindex)}{2!}(\latentfxn_* - \posteriormean_\sampleindex)^2+\dots+\frac{\logprobs^{(\taylorterms)}(\posteriormean_\sampleindex)}{\taylorterms!}(\latentfxn_* - \posteriormean_\sampleindex)^\taylorterms,
\end{gather*}
and $\lagrangeremainder_\taylorterms(\cdot)$ is the truncation error,
\begin{gather*}
    \lagrangeremainder_\taylorterms(\latentfxn_*) = \logprobs(\latentfxn_*) - \taylorpolynomial_\taylorterms(\latentfxn_*).
\end{gather*} 
The computational cost of this approximation is dominated by the evaluation of the posterior predictive variance for each mixture component, resulting in a complexity of $\mathcal{O}(\nosamples\obsdim^2)$.
\begin{prop*}
    Assume that, in a neighborhood of $\posteriormean_\sampleindex$, $\logprobs^{(\taylorterms+1)}(\cdot)$ is uniformly bounded, i.e.,
    \begin{gather*}
        \sup_{\largrangeterm} |\logprobs^{(\taylorterms+1)}(\largrangeterm)| \leq M_{\taylorterms+1}.
    \end{gather*}
    Then the expected absolute truncation error of the $\taylorterms^{th}$ order Taylor approximation satisfies 
    \begin{gather*}
        \Exp{|\lagrangeremainder_\taylorterms(\latentfxn_*)|} \leq \truncationconst_{\taylorterms+1} \, \posteriorstd_\sampleindex^{\taylorterms + 1},
    \end{gather*}
    where $\truncationconst_{\taylorterms+1}$ is a constant independent of $\posteriorstd_\sampleindex$.
\end{prop*}

\begin{pf}
    The proof follows from standard results on Taylor's theorem. The Lagrange form of the remainder for expansion about $\posteriormean_\sampleindex$, is given by
    \begin{gather*}
        \lagrangeremainder_\taylorterms(\latentfxn_*) = \frac{\logprobs^{(\taylorterms+1)}(\largrangeterm)}{(\taylorterms+1)!}(\latentfxn_* - \posteriormean_\sampleindex)^{\taylorterms+1}, \quad \largrangeterm \in (\min \{ \latentfxn_*, \posteriormean_\sampleindex\},\max \{ \latentfxn_*, \posteriormean_\sampleindex\}).
    \end{gather*}
    Using the uniform bound on $\logprobs^{(\taylorterms+1)}(\cdot)$, we obtain
    \begin{gather*}
        |\lagrangeremainder_\taylorterms(\latentfxn_*)| \leq \frac{M_{\taylorterms+1}}{(\taylorterms+1)!} |\latentfxn_* - \posteriormean_\sampleindex|^{(\taylorterms+1)}.
    \end{gather*}
    Taking expectations with respect to $\latentfxn_* \sim p_\sampleindex(\cdot)$ yields
    \begin{gather*}
        \Exp{|\lagrangeremainder_\taylorterms(\latentfxn_*)|} \leq \frac{M_{\taylorterms+1}}{(\taylorterms+1)!} \Exp{|\latentfxn_* - \posteriormean_\sampleindex|^{\taylorterms+1}}.
    \end{gather*}
    Since $\latentfxn_* - \posteriormean_\sampleindex = \posteriorstd_s \, \stdnormrv_*$, with $\stdnormrv_* \sim \mathcal{N}(0,1)$,
    \begin{gather*}
        \Exp{|\latentfxn_* - \posteriormean_\sampleindex|^{\taylorterms+1}} = \posteriorstd_\sampleindex^{\taylorterms+1}\Exp{|\stdnormrv_*|^{\taylorterms+1}}.
    \end{gather*}
    The absolute moment $\Exp{|\stdnormrv_*|^{\taylorterms+1}}$ is finite and admits the closed-form expression
    \begin{gather*}
        \Exp{|\stdnormrv_*|^{\taylorterms+1}} = \frac{2^{\taylorterms/2 + 1}}{\sqrt{2\pi}} \Gamma\left(\frac{\taylorterms+2}{2}\right),
    \end{gather*}
    which is derived in~\ref{subsection: propositionproofsi}. Combining the above expression with the expected absolute truncation error yields
    \begin{gather*}
        \pushQED{\qed}
        \Exp{|\lagrangeremainder(\latentfxn_*)|} \leq \truncationconst_{\taylorterms+1} \posteriorstd_\sampleindex^{\taylorterms+1},
    \end{gather*}
    where
    \begin{gather*}
        \truncationconst_{\taylorterms+1} = \frac{M_{\taylorterms+1}}{(\taylorterms+1)!} \frac{2^{\taylorterms/2 +1}}{\sqrt{2\pi}}\Gamma\left(\frac{\taylorterms+2}{2}\right).\qedhere
\popQED
    \end{gather*}
\end{pf}
The proposition establishes that the expected absolute truncation error of the $\taylorterms^{th}$-order Taylor approximation of the log-density scales as a polynomial funciton of the posterior standard deviation, specifically $\smallO(\posteriorstd_\sampleindex^{\taylorterms+1})$. Practically, this result provides a quantitative justification for using for using low-order Taylor expansions to approximate the log-density of each Gaussian mixture component when the posterior variance is small.

In the context of sequential design (Eq.~\eqref{eq: optimizationprob}), the goal is to select the input \( \varins_* \) that maximizes the differential entropy. Because the entropy of the Gaussian mixture predictive distribution does not admit a closed-form expression, we instead optimize the Taylor approximation derived in this work. In addition, we derive a closed-form lower bound on the entropy. This bound provides a theoretically grounded reference value for the true entropy of the mixture distribution and can be evaluated analytically, offering insight into the behavior of the predictive uncertainty.

\begin{theorem*}    
    A lower bound $\entropy_{\text{LB}}(\cdot)$ of Eq.~\eqref{eq: gppostentropy} is given by 
    \begin{gather*}
        \entropy_{\text{LB}}(\latentfxn_*) = -\frac{1}{\nosamples}\sum_{\sampleindex=1}^\nosamples \log \left(\frac{1}{\nosamples}\sum_{\sampleindex'=1}^\nosamples \crossoverlap_{\sampleindex, \sampleindex'} \right),
    \end{gather*}
    where \( \crossoverlap_{\sampleindex, \sampleindex'} \) denotes the cross-overlap between predictive components:
    \begin{gather*}
        \crossoverlap_{\sampleindex, \sampleindex'} = p(\posteriormean_{\sampleindex}), \quad \posteriormean_{\sampleindex} \sim \mathcal{N}( \posteriormean_{\sampleindex'}, \posteriorstd_\sampleindex^2 + \posteriorstd_{\sampleindex'}^2).
    \end{gather*}
    That is, $\crossoverlap_{\sampleindex, \sampleindex'}$ is Gaussian density for the random variable $\posteriormean_{\sampleindex}$ with mean $\posteriormean_{\sampleindex'}$ and variance $\posteriorstd_\sampleindex^2 + \posteriorstd_{\sampleindex'}^2$.\\
\end{theorem*}
\begin{pf}
    Since  $-\log p(\latentfxn_*)$ is convex in $p(\latentfxn_*)$, Jensen's inequality can be employed. Thus, with $-\log \E{[p(\latentfxn_*)]} \leq \E{[-\log{p(\latentfxn_*)]}}$ the lower bound of the differential entropy (Eq.~\eqref{eq: gppostentropy}) is obtained:
    \begin{align*}
        \entropy(\latentfxn_*)
        &= -\frac{1}{\nosamples} \sum_{\sampleindex=1}^\nosamples 
        \int p_\sampleindex(\latentfxn_*) \log p(\latentfxn_*) \, \d \latentfxn_* \\
        &= -\frac{1}{\nosamples} \sum_{\sampleindex=1}^\nosamples 
        \Exp[\latentfxn_* \sim p_\sampleindex(\cdot)]{\log p(\latentfxn_*)}\\
        &\geq -\frac{1}{\nosamples} \sum_{\sampleindex=1}^\nosamples 
        \log \left( \Exp[\latentfxn_* \sim p_\sampleindex(\cdot)]{p(\latentfxn_*)}\right)\\
        &= - \frac{1}{\nosamples} \sum_{\sampleindex=1}^\nosamples 
        \log \left( \int p_\sampleindex(\latentfxn_*) \, p( \latentfxn_*) \, \d \latentfxn_* \right) \\
        &= -\frac{1}{\nosamples}\sum_{\sampleindex=1}^\nosamples 
        \log \left( \frac{1}{\nosamples} \sum_{\sampleindex'=1}^\nosamples \crossoverlap_{\sampleindex, \sampleindex'} \right),
    \end{align*}
    with the constant
    \begin{gather*}
        \pushQED{\qed}\crossoverlap_{\sampleindex,\sampleindex'} %
        = \int p_\sampleindex(\latentfxn_*) \, p_{\sampleindex'}(\latentfxn_*)  \, \d \latentfxn_* = p(\posteriormean_\sampleindex), \quad \posteriormean_\sampleindex\sim \mathcal{N}( \posteriormean_{\sampleindex'}, \posteriorstd_\sampleindex^2 + \posteriorstd_{\sampleindex'}^2). \qedhere \popQED
    \end{gather*}
\end{pf}
For completeness,~\ref{subsection: crossoverlapsi} provides a detailed derivation showing that this cross-overlap integral evaluates to a Gaussian density. The lower bound has comparable computational complexity to the Taylor approximation, while avoiding the additional cost associated with evaluating multiple Taylor terms when the expansion order is large.
\FloatBarrier
\section{Numerical Example: Phase Equilibria}
\label{sec: casestudy}

\subsection{Motivation}
Modeling phase equilibria, particularly vapor-liquid equilibrium (VLE), is a fundamental task in chemical engineering. Accurate VLE models are essential for the design and optimization of separation operations such as distillation, absorption, and liquid-liquid extraction~\cite{geankoplis2003}. These processes rely on phase behavior to selectively separate components from mixtures. The ability to predict phase compositions as a function of temperature, pressure, and mixture composition is critical to their design and efficiency.

A foundational approach to modeling VLE is Raoult's law, which provides a simple relationship between the composition of the liquid and vapor phases in equilibrium. For ideal mixtures, Raoult’s law states that the partial pressure of each component in the vapor phase is equal to the product of its mole fraction in the liquid phase and its pure-component vapor pressure. Mathematically, Raoult's law is
\begin{gather*}
    \molefrac^{(v)}_\speciesindex \,\pressure  =  \molefrac^{(\ell)}_\speciesindex \,\pressure_\speciesindex^*,
\end{gather*}
where $\molefrac_\speciesindex$ is the mole fraction of species $\speciesindex$ in the vapor $(v)$ or liquid $(\ell)$ phase, $\pressure$ is the total system pressure [Pa], and $\pressure^*_\speciesindex$ is the equilibrium vapor pressure of the pure component [Pa].

Raoult’s law assumes ideal mixing, which applies when components are chemically similar and the solution is nearly pure. In most real mixtures, deviations arise from intermolecular interactions between dissimilar molecules that differ in strength from those between like molecules~\cite{petrucci2007}. To account for these effects, Raoult’s law can be extended by introducing an activity coefficient, $\actcoeff_\speciesindex$ [~\,], resulting in:
\begin{gather}
    \molefrac^{(v)}_\speciesindex \,\pressure  =  \molefrac^{(\ell)}_\speciesindex \,\actcoeff_\speciesindex \,\pressure_\speciesindex^*. \label{eq: extendedraoults}
\end{gather}
The activity coefficient corrects for non-ideal interactions in the liquid phase and is a key quantity in VLE modeling. It varies with composition, temperature, and pressure, and can be estimated using thermodynamic models such as the Wilson~\cite{wilson1964}, non-random two-liquid (NRTL)~\cite{Renon1968}, or universal quasichemical (UNIQUAC)~\cite{Abrams1975, MAURER1978} models.

Although these models are physically grounded and often accurate, they can be computationally intensive, particularly when their parameters are estimated from molecular dynamics~\cite{Kohns2017}. This complexity can become a bottleneck in process simulations, optimization routines, or real-time control applications.

\subsection{Extension to Binary Systems}
To address the need for surrogate representations of activity coefficients, we use BITS for GAPS (Fig.~\ref{fig: propframework}) to model activity coefficients. We then embed the activity coefficient in extended Raoult’s law (Eq.~\eqref{eq: extendedraoults}) to create a hybrid model for predicting VLE phase data. We leverage this phase data to inform the design of a distillation column.

The GP surrogate predicts the activity coefficient of \proh, $\actcoeff_{\proh}$ [~\,], based on its mole fraction, $\molefrac_{\proh}$ [~\,], and the system temperature, \temp~[K]. The activity coefficient of \water, $\actcoeff_{\water}$ [~\,], is inferred from the activity coefficient of \proh using the Gibbs–Duhem relationship (Eq.~\eqref{eq: gibbsduhem}). Following the notation from Section~\ref{sec: hybridmodels}, the activity coefficient $\actcoeff_{\proh}(\cdot)$ is treated as a black-box function $\latentfxn(\cdot)$, with inputs $\varins = (\molefrac_{\proh}, \temp)^\top$. The hybrid model, denoted $\mechanisticmodel(\cdot,\cdot)$, uses this black-box output along with mechanistic inputs $\varinsmechonly = (\pressure,\pressure_\proh^*)^\top$ in extended Raoult’s law.

We extend this to binary mixtures using the Gibbs-Duhem relationship, which imposes a thermodynamic constraint on the chemical potentials of the components in a solution. Specifically, if the activity coefficient of one component is known as a function of composition, the activity coefficient of the other component can be determined from this relationship.

The differential form of the Gibbs-Duhem equation for a binary system is
\begin{gather*}
    \molefrac_1 \, \d\ln{\actcoeff_1} + \molefrac_2 \, \d\ln{\actcoeff_2} = 0, \quad \molefrac_1 + \molefrac_2 = 1.
\end{gather*}
This expression can be rearranged and treated as a first-order separable equation in terms of \(\molefrac_1\), yielding the following integral form:
\begin{gather*}
    \int_{\ln \actcoeff_2(\molefrac_1^{\text{ref}})}^{\ln \actcoeff_2(\molefrac_1)} \d\ln \actcoeff_2 = -\int_{\ln \actcoeff_1(\molefrac_1^{\text{ref}})}^{\ln \actcoeff_1(\molefrac_1)} \frac{\molefrac_1}{1 - \molefrac_1} \, \d\ln \actcoeff_1.
\end{gather*}
Choosing the reference state as \(\molefrac_1^{\text{ref}} = 0\) simplifies the expression to:
\begin{gather}
    \ln \actcoeff_2(\molefrac_1) = -\int_{\ln \actcoeff_1(0)}^{\ln \actcoeff_1(\molefrac_1)} \frac{\molefrac_1}{1 - \molefrac_1} \, \d\ln \actcoeff_1. \label{eq: gibbsduhem}
\end{gather}
This expression provides a means to compute the activity coefficient of component 2, \(\actcoeff_2\), based on the known behavior of component 1, \(\actcoeff_1\), across composition space~\cite{robny2014}. Eq.~\ref{eq: gibbsduhem} can be evaluated with random samples the surrogate posterior.

It is important to note that this integral diverges as \(\molefrac_1 \rightarrow 1\), due to the singularity in the integrand. Therefore, the upper limit of integration is truncated at \(\molefrac_1 = 1 - \epsilon\), where \(\epsilon\) is a small number (e.g., 10$^{-4}$) ensuring that \(\ln \actcoeff_2(\molefrac_1)\) remains finite and well-approximated.

\subsection{Implementation Details}
We conduct all analyses in \python and \julia.\ref{subsection: softreqs} lists the software requirements, including each package’s version and build.

We generate the training data for the surrogate by evaluating the Wilson model~\cite{wilson1964}, which we choose as a well-established benchmark for validating our framework. To construct the initial dataset, we use Latin hypercube sampling to select ten input combinations, implementing it with the \texttt{LatinHypercube} class from the \python \texttt{scipy.stats.qmc} submodule~\cite{Virtanen2020}. We then randomly split the dataset in half to define the training and test sets.

We select the bounds of the design space to reflect the physical constraints of the system. Specifically, we vary the temperature between 350~K and 367~K, where the upper bound corresponds to the boiling point of pure \proh at atmospheric pressure. We vary the mole fraction of \proh from zero to one to cover the whole compositional range.

We generate activity coefficients using the \texttt{activity\_coefficient} function from the \julia package \clapeyron~\cite{walker2022}, with the Wilson model~\cite{wilson1964}. Because we conduct the analysis in \python, we link \julia using the \texttt{JuliaCall} module. We define a wrapper function in \julia to evaluate the activity coefficients for specified compositions and temperatures, and import it into \python using \texttt{JuliaCall.include}~\cite{JuliaCall}.

We implement the Gaussian process (GP) surrogate model in \python using the \gpflow library~\cite{Matthews2017, vanderWilk2020}. Since activity coefficients are strictly positive, we apply a logarithmic transformation to map them onto the real line, making the GP a suitable prior over the transformed outputs.

We use a zero-mean function for the GP prior, which corresponds to a prior belief that the latent activity coefficient is unity across the design space (i.e., ideal mixing). For the covariance function, we implement a custom anisotropic squared exponential kernel (Eq.~\eqref{eq: sekernel}) to model correlations across the two-dimensional input space. This kernel includes three hyperparameters: a kernel standard deviation $\hyperparam_1$, and two input length scales $\hyperparam_2$ and $\hyperparam_3$, corresponding to the mole fraction of \proh and temperature, respectively.

We apply input transformations to improve numerical conditioning and encode prior structural knowledge. Specifically, we increase the \proh mole fraction input by 0.1 and apply a logarithmic transformation, and we normalize the temperature input. The logarithmic transformation reflects the prior belief that, at small \proh mole fractions, the activity coefficient exhibits stronger non-ideal behavior and more rapid variation, and thus benefits from increased resolution in this regime. Normalizing the temperature input places both inputs on comparable scales, improving kernel conditioning and ensuring that the GP length scale hyperparameters operate on similar orders of magnitude, which facilitates more stable and interpretable inference.

To improve numerical stability, we add a jitter term to the diagonal of the GP covariance matrix. The jitter is chosen to yield an uncertainty band approximately 20\% of the average activity coefficient, consistent with reported discrepancies between experimental data and well-established activity coefficient models in water–alcohol systems~\cite{Rousseau1972}. More generally, the jitter magnitude can be selected based on prior beliefs about surrogate model error.

We incorporate prior beliefs about the smoothness and amplitude of the target function by placing priors on each kernel hyperparameter. Table~\ref{tab: hyperparameterpriors} summarizes these choices. LogNormal and Gamma priors are used to ensure all hyperparameters remain strictly positive, consistent with their physical interpretation as scales or variances. LogNormal priors are assigned to the kernel standard deviation $\hyperparam_1$ and the mole fraction length scale $\hyperparam_2$, since both the model outputs and the mole fraction input are log-transformed. This allows prior beliefs to be specified directly in physical space, which is more intuitive than reasoning about priors in the transformed space.

\begin{table}[]
\centering
\begin{tabular}{|l|c|c|c|}
\hline
Hyperparameter & Variable & Prior & Parameterization \\
\hline
Kernel Std. Dev. & $\hyperparam_1$ & LogNormal & (0, 2.0) \\
Mole Fraction Lengthscale & $\hyperparam_2$ & LogNormal & (log(0.3), 0.5) \\
Temperature Lengthscale & $\hyperparam_3$ & Gamma & (4.0, 2.0) \\
\hline
\end{tabular}
\caption{Priors used for the GP kernel hyperparameters. Parameterization is (location, scale) for LogNormal priors and (shape, rate) for the Gamma prior.}
\label{tab: hyperparameterpriors}
\end{table}

The numerical values of the hyperparameter priors reflect prior beliefs about the scale and smoothness of the underlying physical system. The kernel standard deviation is assigned a LogNormal prior with median one and large log-scale variance, corresponding to a weakly informative prior that allows the overall magnitude of activity coefficients to vary over several orders of magnitude. This reflects substantial uncertainty in the amplitude of non-ideal behavior in binary mixtures.

The mole fraction length scale is assigned a LogNormal prior with median 0.3 and moderate dispersion, encoding the belief that the latent function varies on the order of tens of percent in composition space. This reflects physical intuition that non-ideal behavior changes appreciably over moderate changes in composition, while remaining smooth at very small scales.

The temperature length scale is assigned a Gamma prior with shape four and rate two, with mean two and mode 1.5 on the normalized temperature domain. This encodes the belief that the response varies smoothly with temperature over a length scale comparable to the normalized domain. The Gamma distribution discourages unrealistically short length scales, which would correspond to excessive sensitivity to temperature, while remaining sufficiently flexible to allow data-driven adaptation.

As with any hierarchical Bayesian model, the results obtained using BITS for GAPS depend on the specification of the hyperparameter priors. These priors encode assumptions about the scale, smoothness, and uncertainty structure of the latent function, and therefore influence both posterior inference and the behavior of the entropy-based acquisition function. In this work, the priors were chosen to be physically interpretable, with the intent of regularizing the model while allowing the data to dominate inference. Nevertheless, different prior choices may lead to different posterior geometries, which in turn can affect hyperparameter uncertainty and the resulting sampling decisions. The proposed framework does not rely on any specific parametric form of the priors, and alternative distributions may be substituted to reflect different modeling assumptions or domain knowledge.

We perform inference using the \texttt{HamiltonianMonteCarlo} class from the \texttt{mcmc} submodule of \texttt{TensorFlow Probability}~\cite{dillon2017}. We set the sampler to use a fixed step size of 0.05 with five leapfrog steps per iteration. During the first five iterations, we enable step size adaptation with an adaptation rate of 0.1 and a target acceptance probability of 0.9. Each HMC chain generates 5,000 samples. We run four independent chains in parallel. We assess MCMC performance using the Gelman–Rubin statistic ($\hat{R}$)~\cite{gelman1992} and effective sample size ($\hat{ESS}$). We compute MAP estimates and 95\% credible intervals using the approximation methods described in Section~\ref{subsection: approxinf}. To visualize hyperparameter correlations, we construct pair plots of the joint marginals. We estimate densities using kernel density estimation (KDE), implemented via the \texttt{stats.kde\_gaussian} class from the \texttt{SciPy} library~\cite{Virtanen2020}.

We deploy the BITS for GAPS framework to guide model development. We identify the point of maximum entropy using the \texttt{SciPy.optimize} module in \python with the Limited-memory Broyden–Fletcher–Goldfarb–Shanno with Box constraints (L-BFGS-B) solver~\cite{byrd1995, Zhu1997}. To find a near-global optimum, we run the optimization with ten restarts, using a Sobol sequence to generate initial values. We approximate entropy using a second order Taylor expansion and 15 samples from the hyperparameter posterior. Since each Gaussian component carries weight $1/\nosamples$, the posterior mass associated with each component is concentrated locally, leading to comparatively small component variances and reduced truncation error in the Taylor approximation.

To quantify the accuracy and uncertainty of the predictive surrogate model, we calculate error metrics between the training and test sets. Specifically, we calculate the root mean squared error (RMSE) and mean absolute error (MAE) for the test and train sets using 50 realizations of the surrogate model posterior. We terminate BITS for GAPS once the RMSE and MAE between test and train sets stabilizes.

To inform the design of a distillation column, we embed samples from the surrogate posterior into extended Raoult's law (Eq.~\eqref{eq: extendedraoults}) to generate VLE phase envelopes. Using these phase envelopes, we estimate the theoretical number of stages required for separation with the McCabe–Thiele method detailed in~\ref{subsection: mccabethiele}. We design the column for a bottoms product composition of 1\% \proh, a feed molar flow rate of 100~mol/s, a feed composition of 10\% \proh, a reflux ratio of 1.0, and a distillate composition of 43\% \proh. We introduce the feed on stage three and specify a total of four stages. With these specifications, we obtain the equilibrium curve from the GP-informed VLE data and derive the operating lines using mass balances and the specified reflux ratio. We confirm the number of theoretical stages by stepping off stages between the operating lines and the equilibrium curve, starting at the distillate composition and continuing down to the bottoms composition.

\FloatBarrier
\subsection{Results \& Discussion}
The results and discussion are organized by six key findings (subsection titles).

\subsubsection{Data visualization demonstrates the need for surrogate modeling.}
\begin{figure}
    \centering
    \begin{subfigure}{0.41\textwidth}
        \centering
        \includegraphics[width = \textwidth]{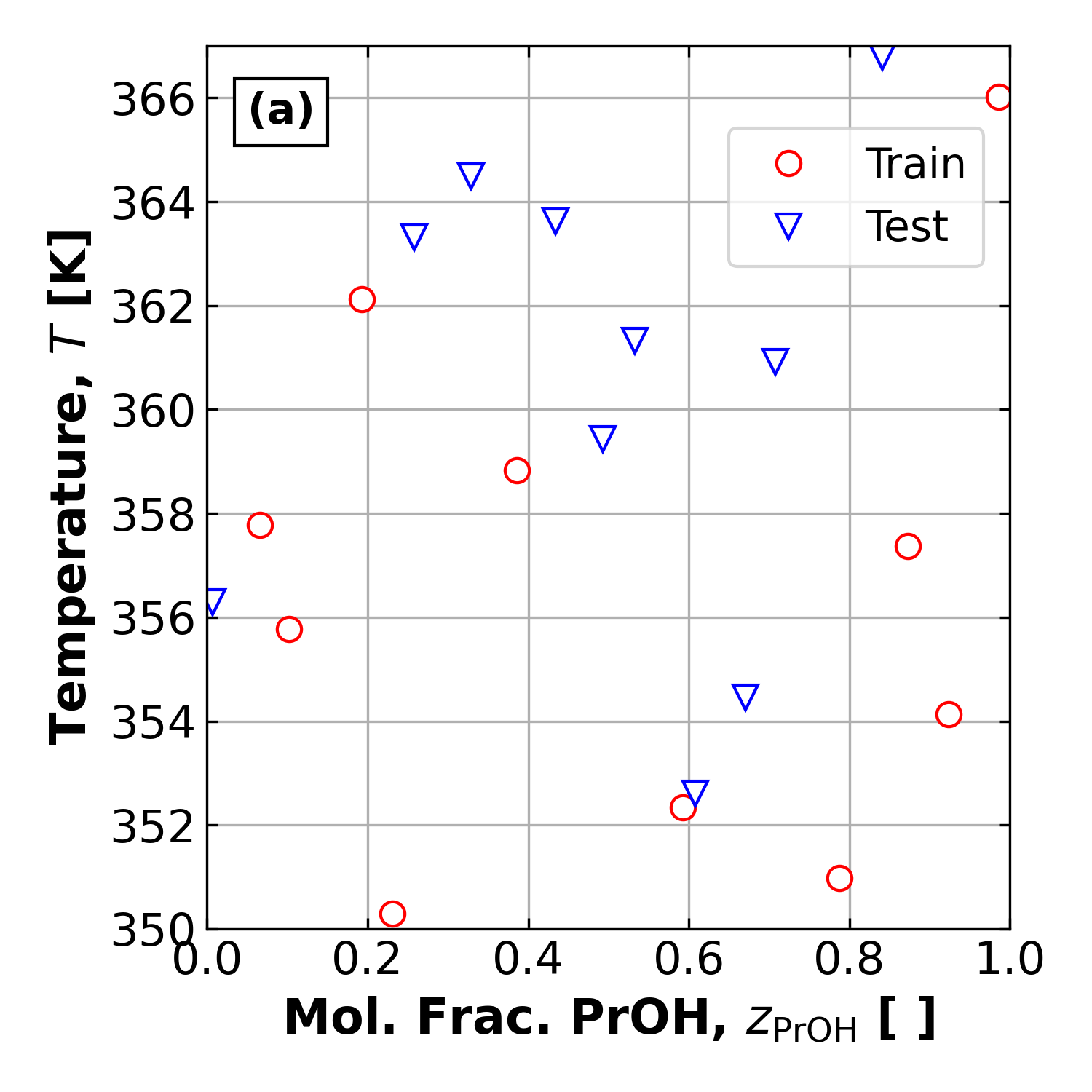}
    \end{subfigure}
    \begin{subfigure}{0.5\textwidth}
        \centering
        \includegraphics[width = \textwidth]{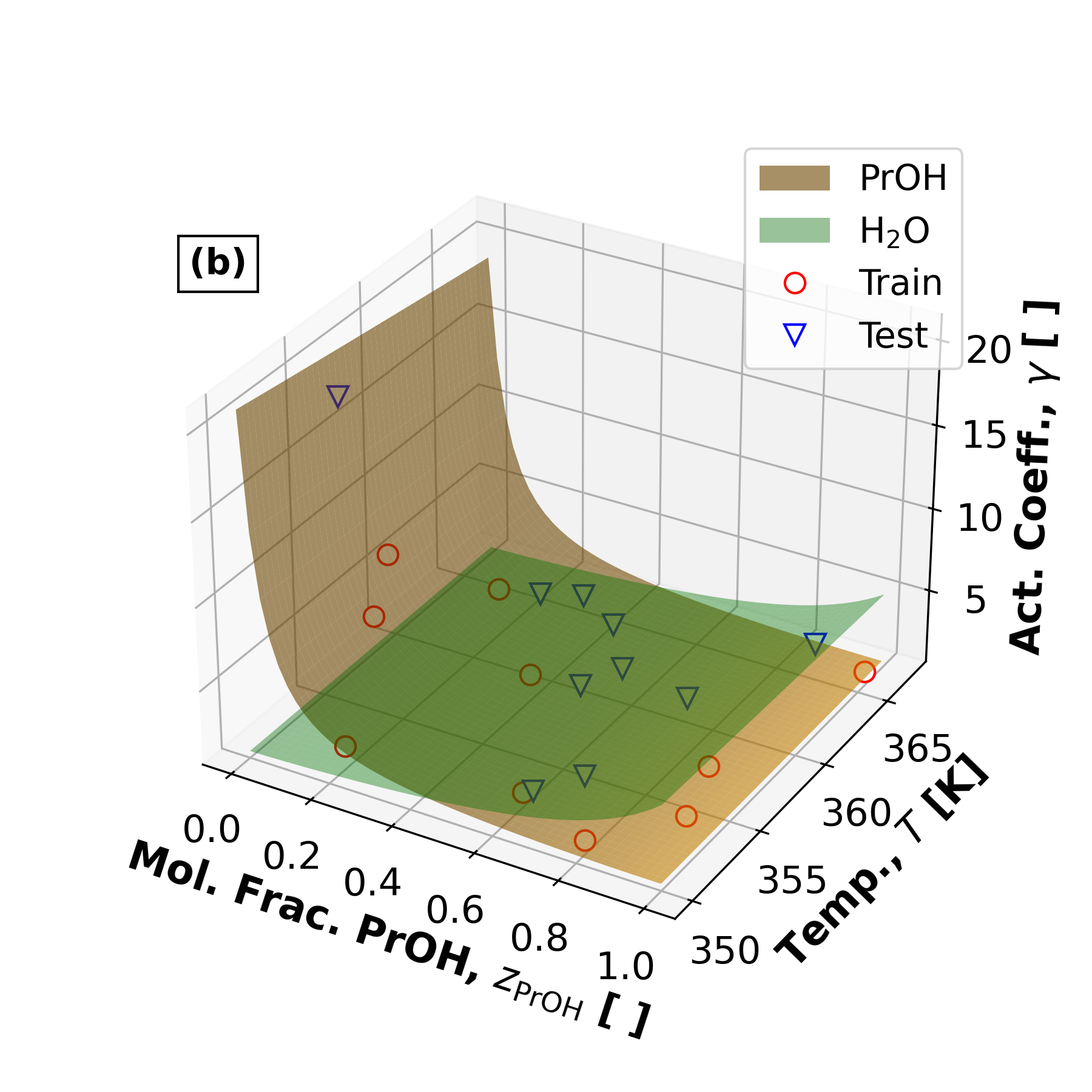}
    \end{subfigure}
    \caption{Training and testing data for the activity coefficient surrogate model. (a) Latin hypercube sample of \proh mole fraction, $\molefrac_{\proh}$~[~\,], and temperature, \temp~[K]. Red circles (blue triangles) indicate selected design points for training (testing). (b) Activity coefficient, \actcoeff~[~\,], as a function of $\molefrac_{\proh}$ and \temp, at atmospheric pressure. \proh is shown in orange; \water in green.}
    \label{fig: trainingdata}
\end{figure}

Figure~\ref{fig: trainingdata}a shows the initial design points for training and testing the activity coefficient surrogate model. Figure~\ref{fig: trainingdata}b shows the latent Wilson (ground truth) activity coefficient model for \water and \proh across the design space. The goal of the GP is to emulate the orange surface in Figure~\ref{fig: trainingdata}b.

Figure~\ref{fig: trainingdata}b highlights the need for an activity coefficient model that captures non-ideal mixing. The activity coefficient of \proh increases significantly as the mixture becomes \water-rich, indicating strong non-ideal behavior. As such, assuming ideal mixing for \proh would lead to inaccurate VLE predictions in this regime. Similarly, the activity coefficient of \water shows non-ideal mixing, especially in \proh-rich phases. While temperature has a less pronounced effect on the activity coefficient than composition, Figure~\ref{fig: trainingdata}b shows minor variations near the dilute limit.

\subsubsection{BITS for GAPS increases model information by identifying optimal designs.}
\begin{figure}
    \centering
    \begin{subfigure}{0.49\textwidth}
        \centering
        \includegraphics[width = \textwidth]{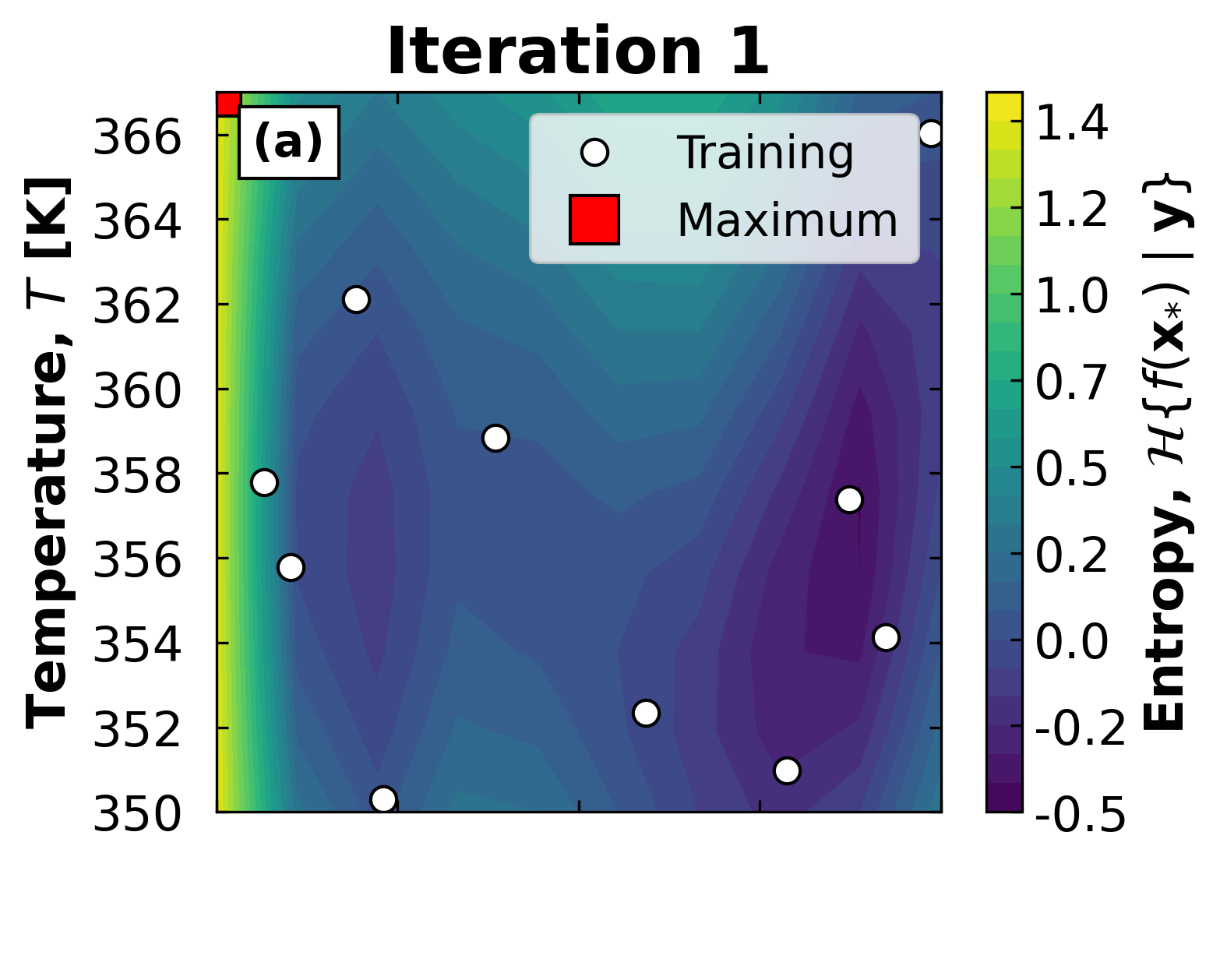}
        \caption*{}
    \end{subfigure}
    \begin{subfigure}{0.49\textwidth}
        \centering
        \includegraphics[width = \textwidth]{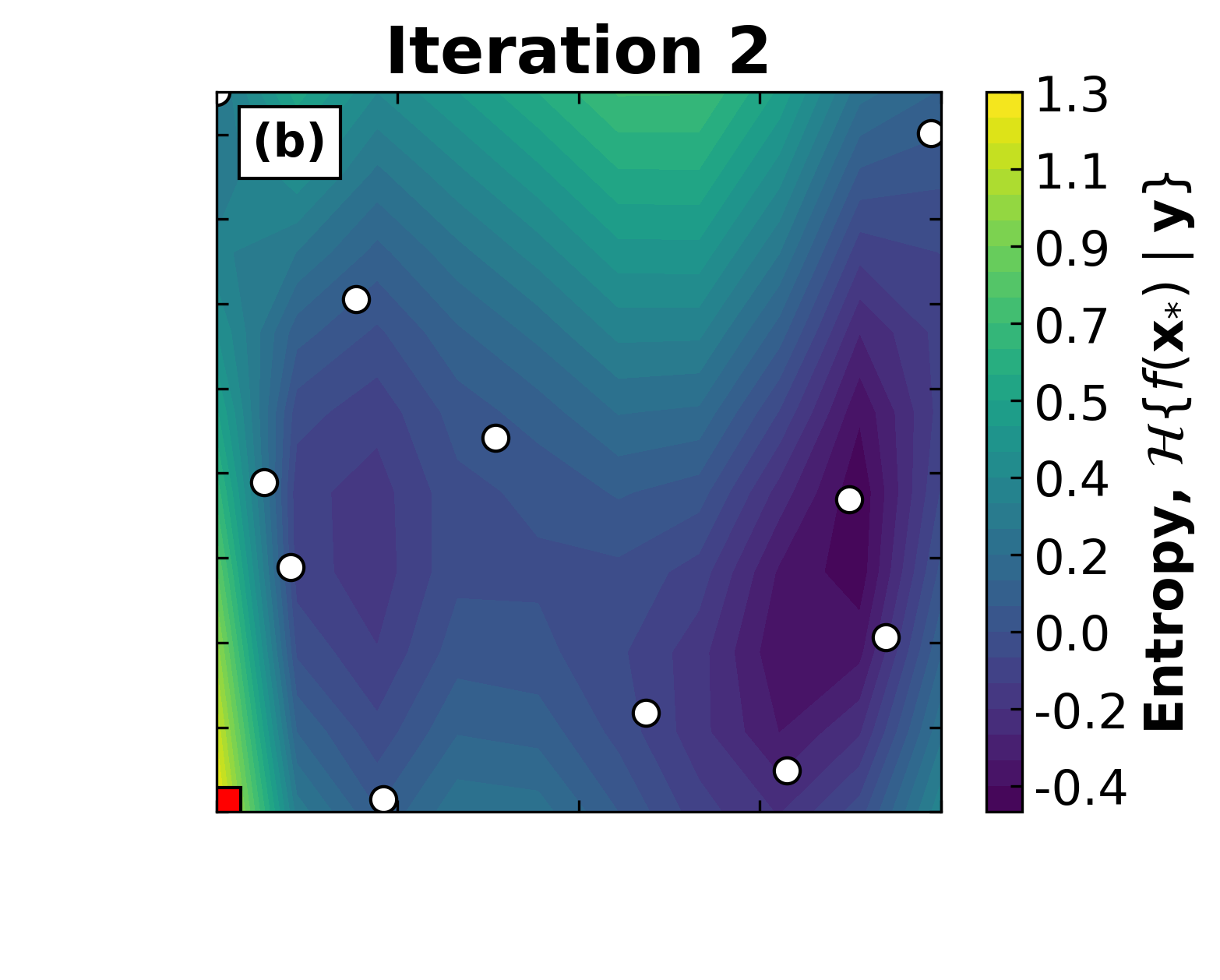}
        \caption*{}
    \end{subfigure}
    \begin{subfigure}{0.49\textwidth}
        \centering
        \includegraphics[width = \textwidth]{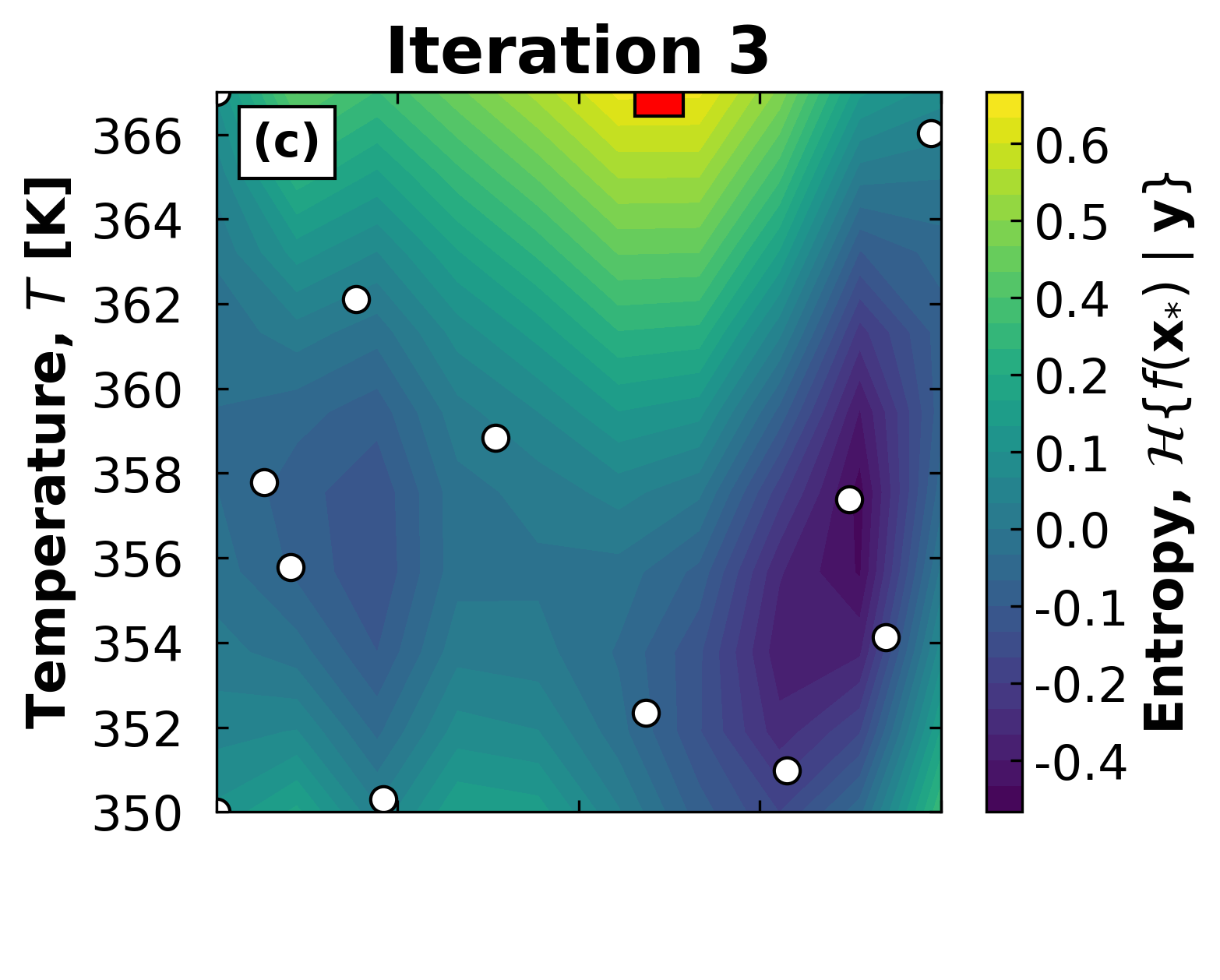}
        \caption*{}
    \end{subfigure}
    \begin{subfigure}{0.49\textwidth}
        \centering
        \includegraphics[width = \textwidth]{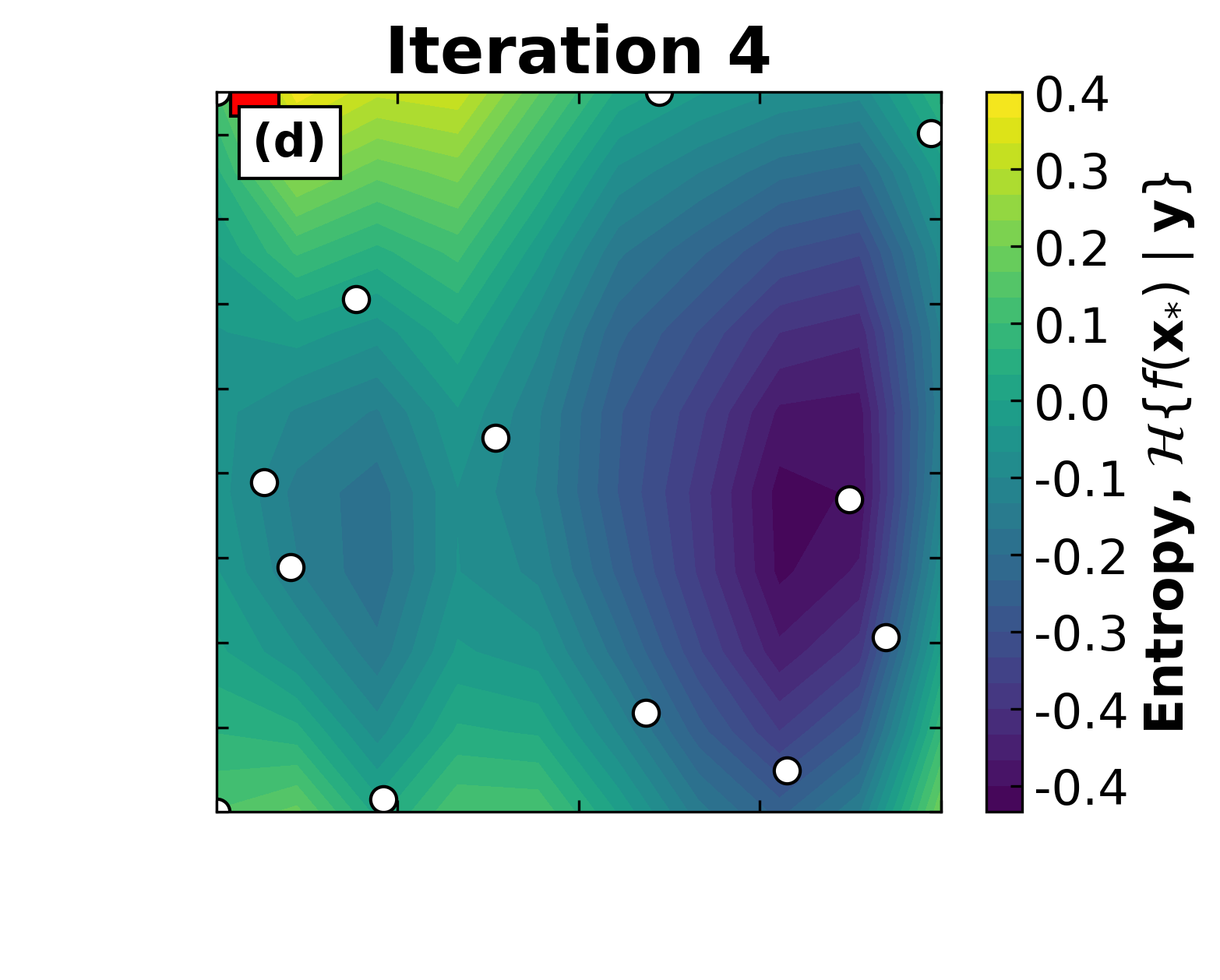}
        \caption*{}
    \end{subfigure}
    \begin{subfigure}{0.49\textwidth}
        \centering
        \includegraphics[width = \textwidth]{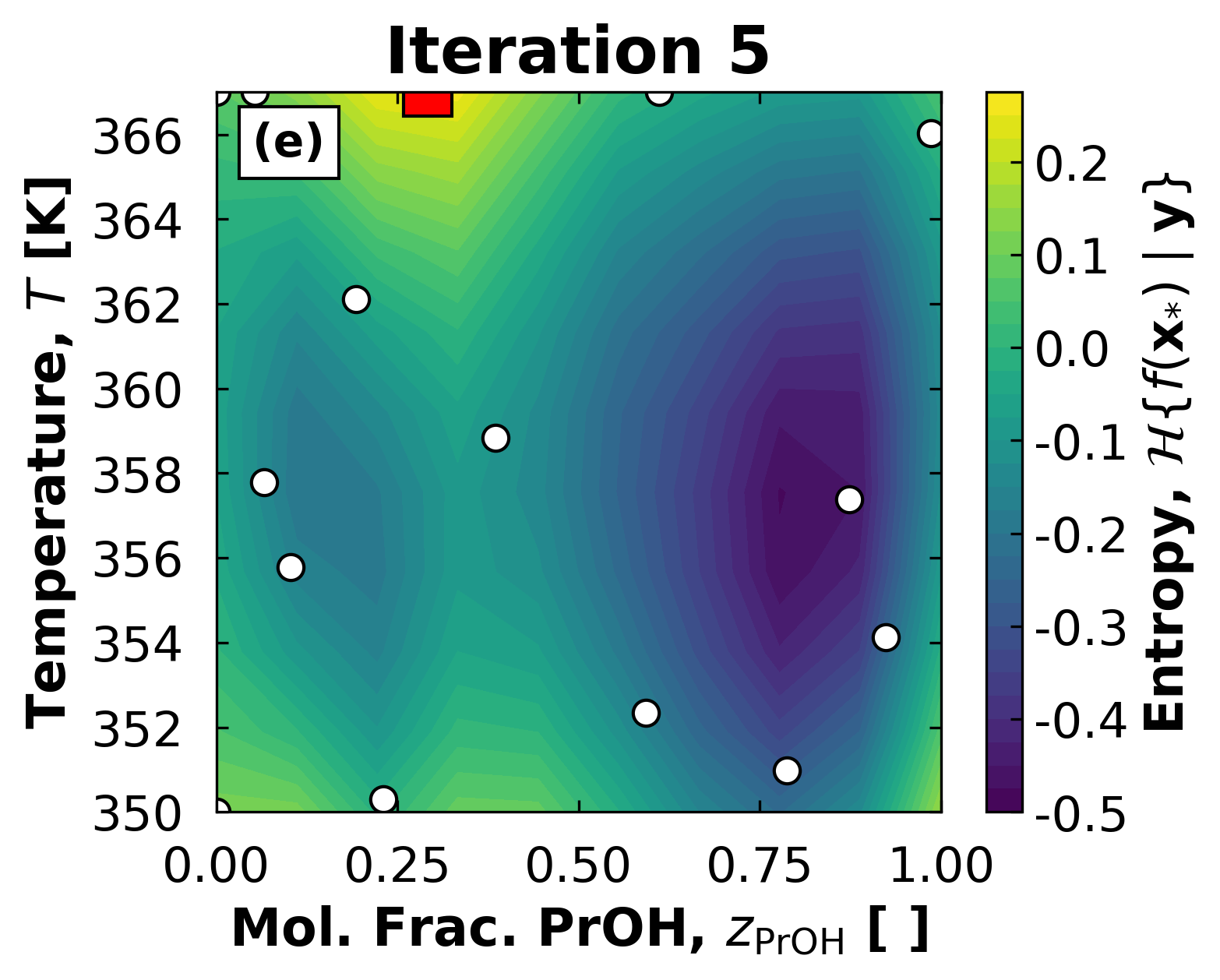}
        \caption*{}
    \end{subfigure}
    \begin{subfigure}{0.49\textwidth}
        \centering
        \includegraphics[width = \textwidth]{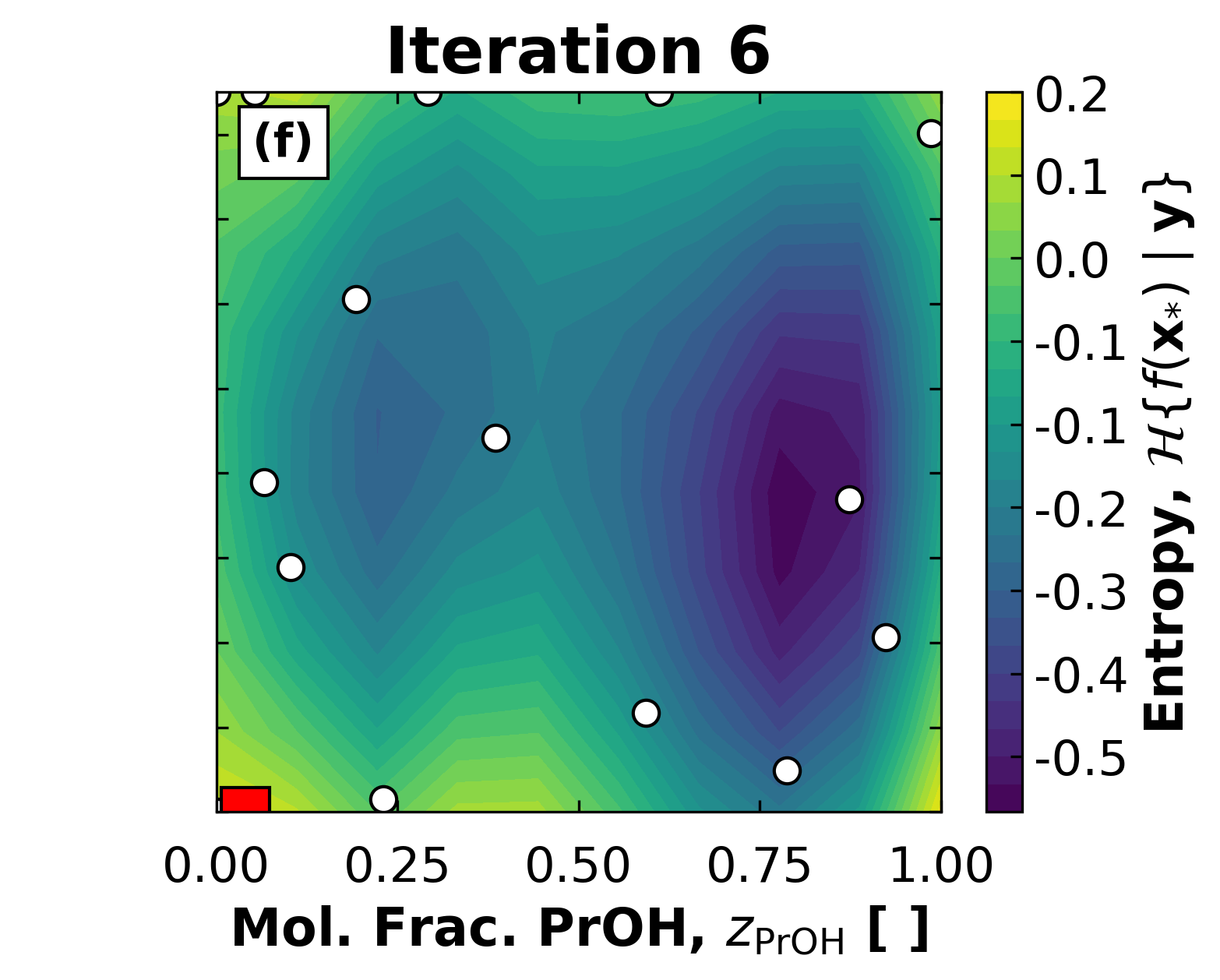}
        \caption*{}
    \end{subfigure}
    \caption{Posterior differential entropy, $\entropy\{\latentfxn(\varins_*)\mid \observations\}$, as a function of temperature, \temp~[K], and mole fraction of \proh, $\molefrac_{\proh}~[~\,]$, at iterations 1-6 (a-f). White circles denote previously sampled (training) points, and red squares denote the locations selected by the optimizer as having maximum posterior entropy. The red squares are sampled and augmented into the training data for subsequent iterations.}
    \label{fig: entropy_surfaces}
\end{figure}

\begin{figure}
    \centering
    \includegraphics[width=0.7\linewidth]{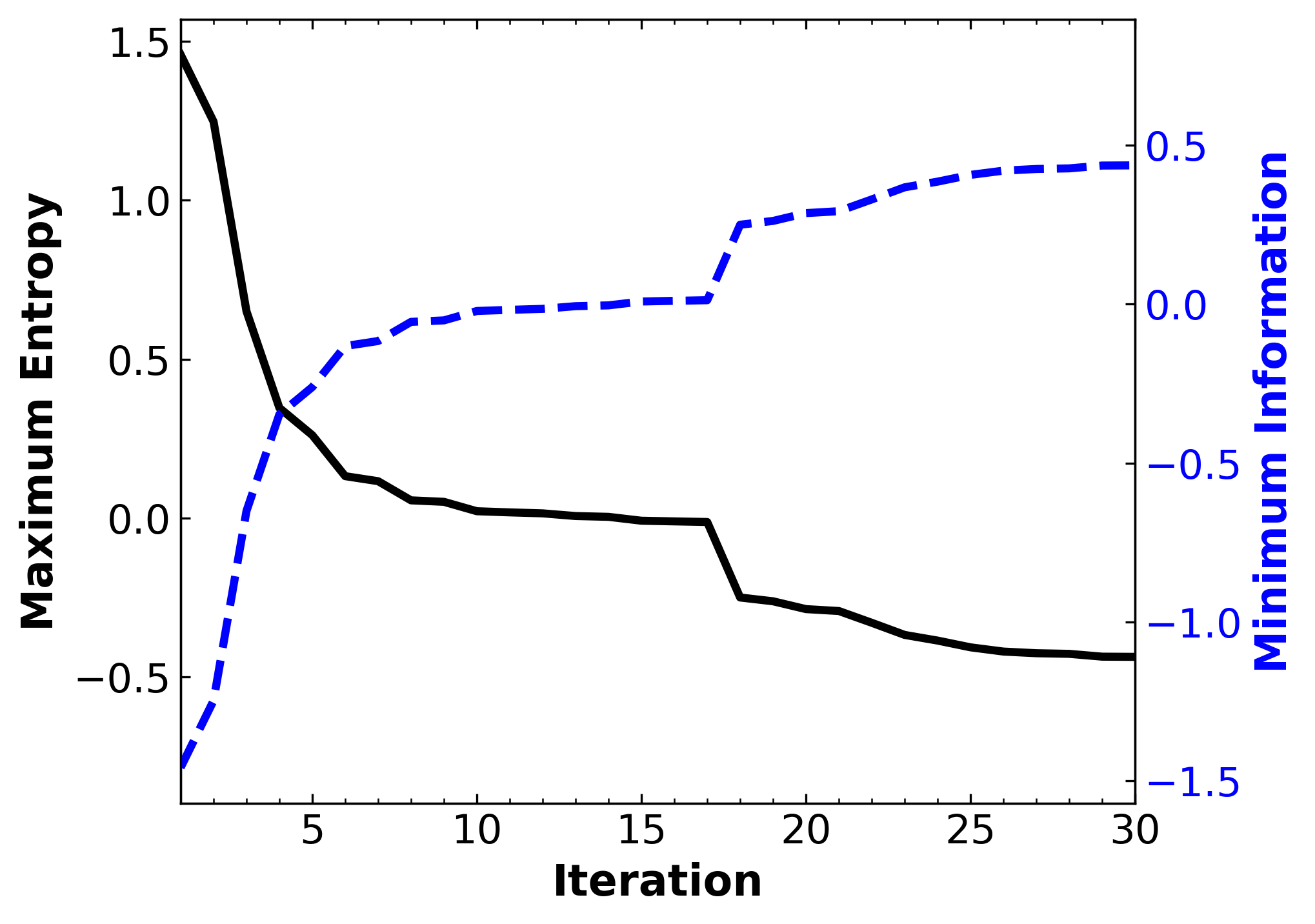}
    \caption{Maximum entropy and minimum information over successive iterations of BITS for GAPS. The black solid line shows the maximum entropy across evaluated candidate models at each iteration. The blue dashed line shows the corresponding minimum information (defined as the negative of maximum entropy).}
    \label{fig: entropyviters}
\end{figure}

To assess how BITS explores and refines the design space, we track the approximated entropy field and its maxima over successive iterations. Figure~\ref{fig: entropy_surfaces} shows the approximated entropy across the design space for the first six iterations of BITS for GAPS. Figure~\ref{fig: entropyviters} shows the maximum entropy (minimum information) found for 30 search iterations.
 
Figure~\ref{fig: entropy_surfaces} demonstrates that the BITS for GAPS successfully identifies the maximum entropy in the design space across iterations. At iteration one (Fig.~\ref{fig: entropy_surfaces}a), the entropy is greatest where training data are scarce (high temperature, \proh-rich). All figures show that the most uncertain design points lie at the temperature extremes. Taken as a whole, this sampling pattern makes sense. Moreover, a GP is an interpolative method, so information (entropy) would be low (high) at the extremes of the design space and regions where data are scarce. 

Figure~\ref{fig: entropyviters} shows the evolution of the maximum posterior predictive entropy and the corresponding minimum information over successive BITS iterations. The maximum entropy, taken over all evaluated candidate models at each iteration, decreases as the algorithm explores the design space, indicating a progressive reduction in model uncertainty. Conversely, the minimum information, defined as the negative of the maximum entropy, increases over iterations, reflecting the cumulative information gain achieved through sequential sampling. Together, these trends demonstrate that BITS for GAPS effectively drives the surrogate model toward more informative and confident predictions as the search proceeds.

\subsubsection{BITS for GAPS improves predictive error.}
\begin{figure}
    \centering
    \begin{subfigure}{0.49\textwidth}
        \centering
        \includegraphics[width = \textwidth]{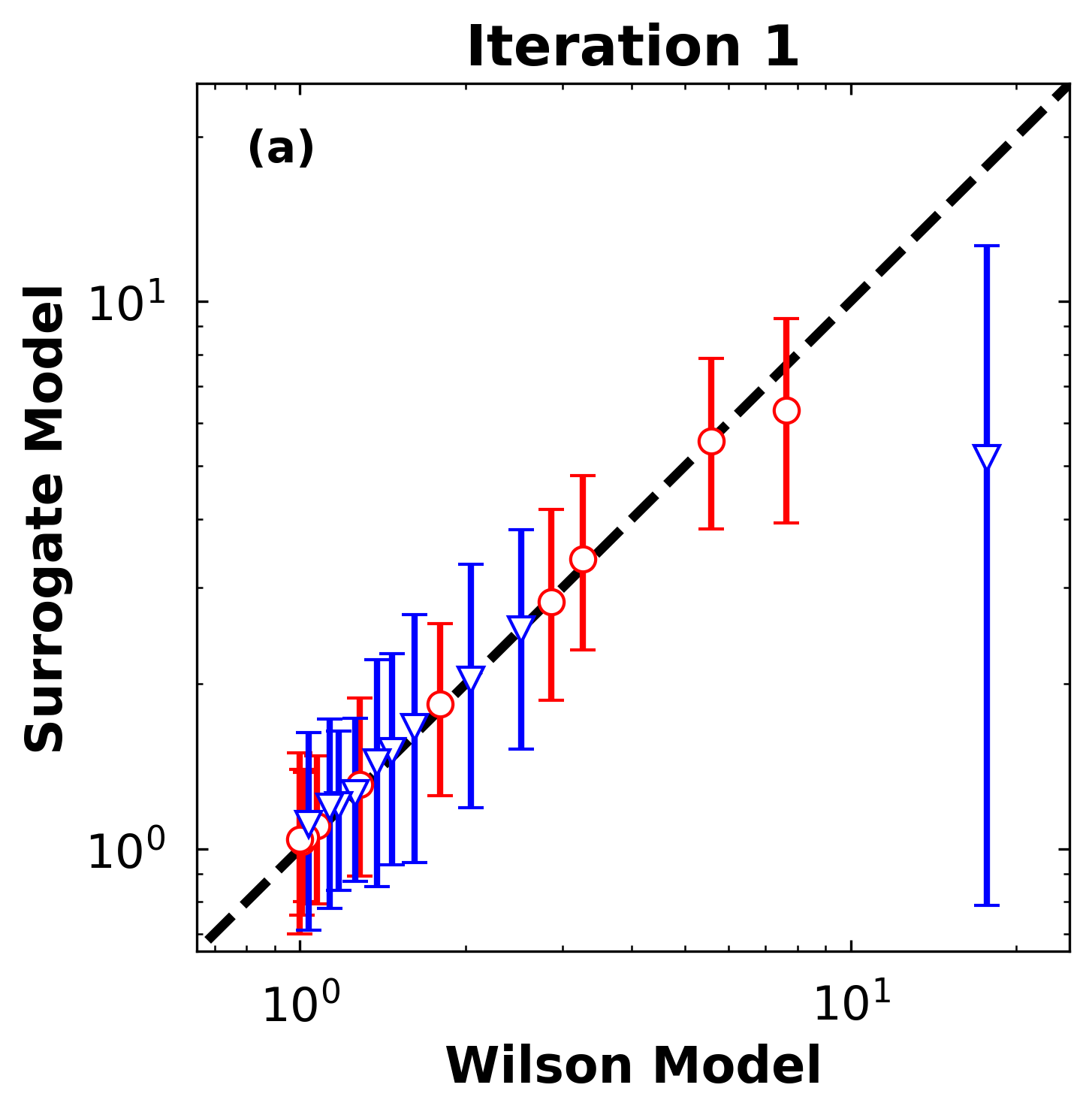}
        \caption*{}
    \end{subfigure}
    \begin{subfigure}{0.49\textwidth}
        \centering
        \includegraphics[width = \textwidth]{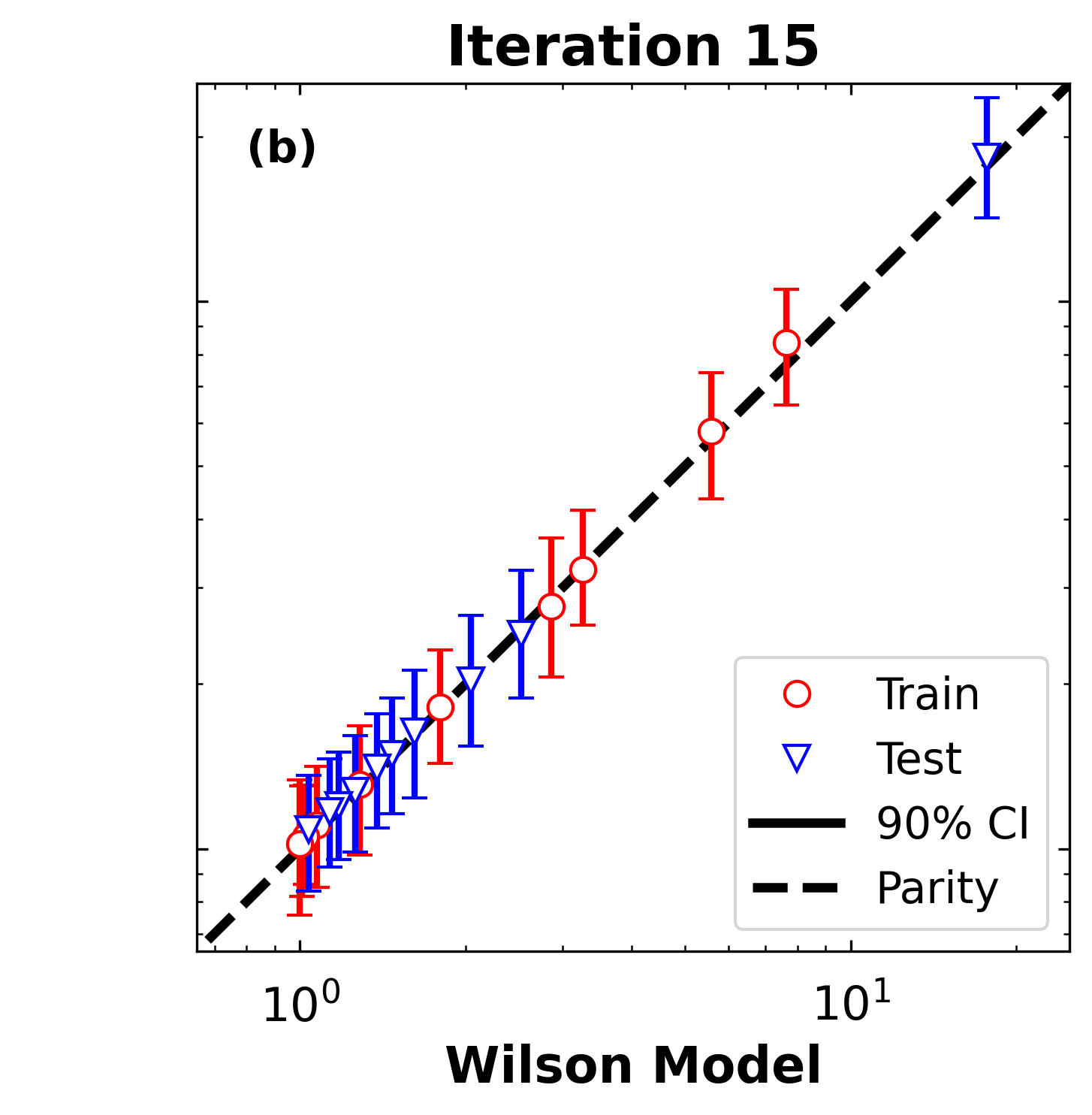}
        \caption*{}
    \end{subfigure}
        \begin{subfigure}{0.49\textwidth}
        \centering
        \includegraphics[width = \textwidth]{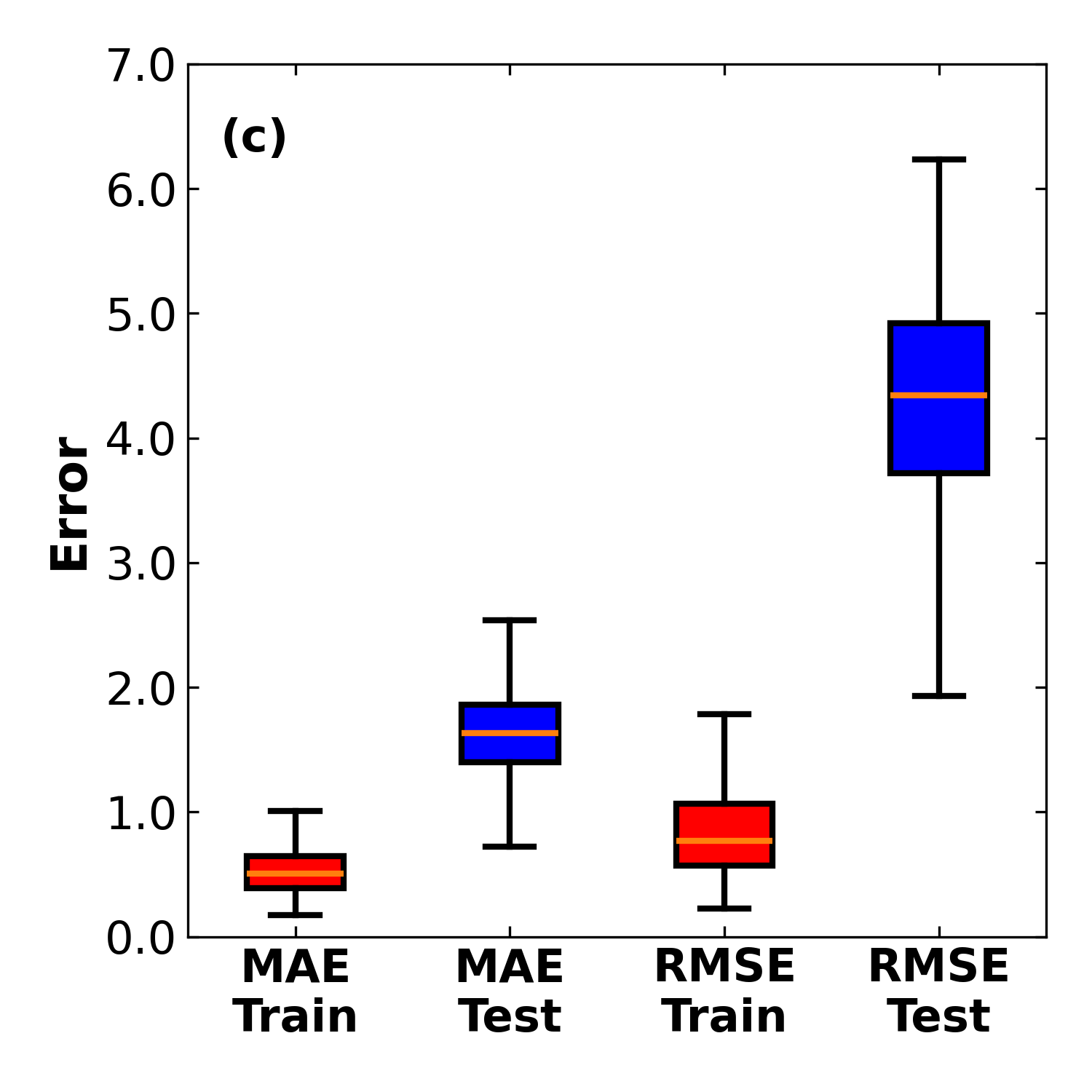}
        \caption*{}
    \end{subfigure}
    \begin{subfigure}{0.49\textwidth}
        \centering
        \includegraphics[width = \textwidth]{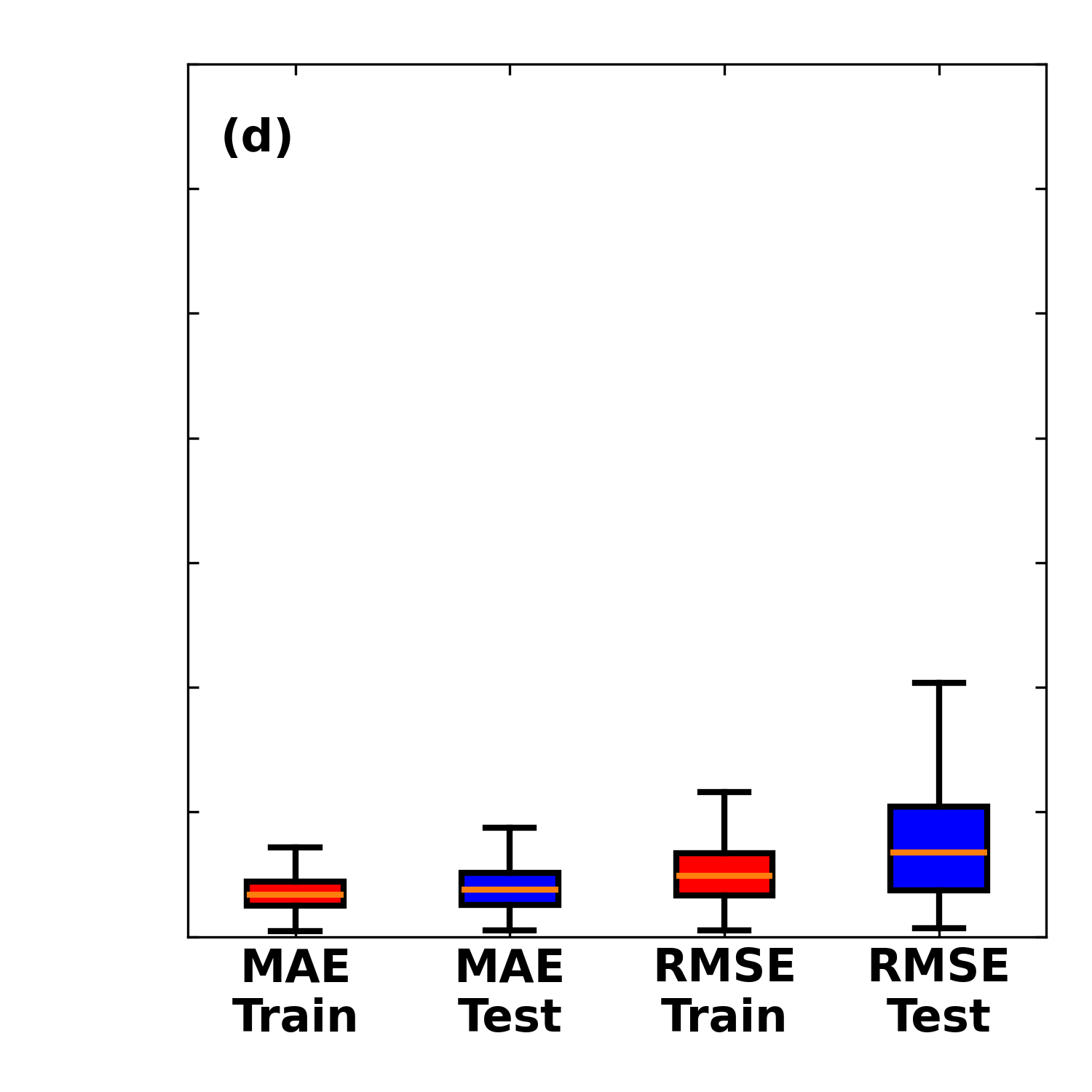}
        \caption*{}
    \end{subfigure}
    \caption{Surrogate model performance at early and late stages of BITS for GAPS. Left column (a, c): Results from iteration 1. Right column (b, d): Results from iteration 15. (a–b) Parity plots comparing surrogate model predictions and Wilson (ground truth) model evaluations for training (red circles) and test (blue downward triangles) datasets. Error bars represent a 90\% credible interval (CI). The dashed black line indicates the ideal parity (1:1) line. (c–d) Box plots showing the distribution of mean absolute error (MAE) and root mean square error (RMSE) for training (red) and test (blue) datasets.}
    \label{fig: gpvalidation}
\end{figure}

Figure~\ref{fig: gpvalidation} presents the predictive performance of the GP surrogate model at both early and late stages of BITS for GAPS. For qualitative assessment, Figures~\ref{fig: gpvalidation}a and~\ref{fig: gpvalidation}b display parity plots comparing GP predictions to the ground truth Wilson outputs at iterations one and 15, respectively. For quantitative evaluation, Figures~\ref{fig: gpvalidation}c and~\ref{fig: gpvalidation}d show box-and-whisker plots of the mean absolute error (MAE) and root mean square error (RMSE) for both the training and testing datasets.

Figures~\ref{fig: gpvalidation}a and~\ref{fig: gpvalidation}b demonstrate how BITS for GAPS progressively improves the accuracy of the GP surrogate model across iterations. At iteration one (Fig.~\ref{fig: gpvalidation}a), the GP underpredicts the ground truth values in the higher output range of the Wilson (ground truth) model, causing noticeable deviations from the parity line. This outcome aligns with our expectations: by imposing ideal mixing through the GP prior mean, we encode the belief that, in the absence of data, the activity coefficient should approach unity. When forced to extrapolate in under-sampled regions, the GP naturally reverts to this prior. By iteration 15 (Fig.~\ref{fig: gpvalidation}b), the updated GP predictions align much more closely with the parity line, indicating improved accuracy and reduced bias.

Figures~\ref{fig: gpvalidation}c and~\ref{fig: gpvalidation}d demonstrate the convergence in error across iterations. Box plots of the MAE and RMSE reveal a reduction in test set errors between iterations one and 15, indicating improved extrapolation beyond the initially sampled design space. Additionally, training set errors and uncertainty decrease slightly. The uncertainty across the test and train sets becomes more uniform by iteration 15, highlighting increased stability and reduced sensitivity to posterior uncertainty as the model gains information.

\subsubsection{BITS for GAPS corrects the systematic bias in the surrogate model.}
\begin{figure}
    \centering
    \begin{subfigure}{0.49\textwidth}
        \centering
        \includegraphics[width = \textwidth]{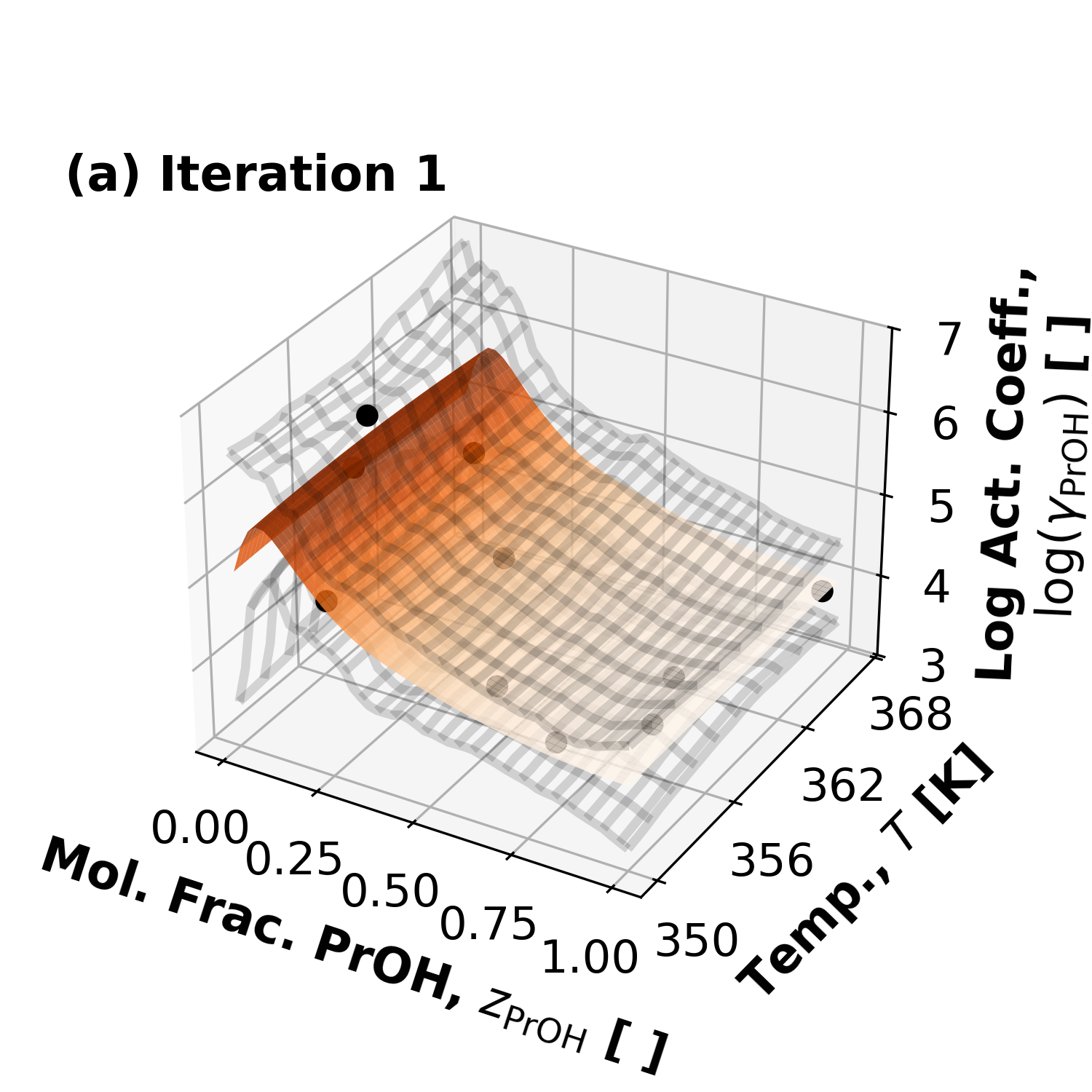}
        \caption*{}
    \end{subfigure}
    \begin{subfigure}{0.49\textwidth}
        \centering
        \includegraphics[width = \textwidth]{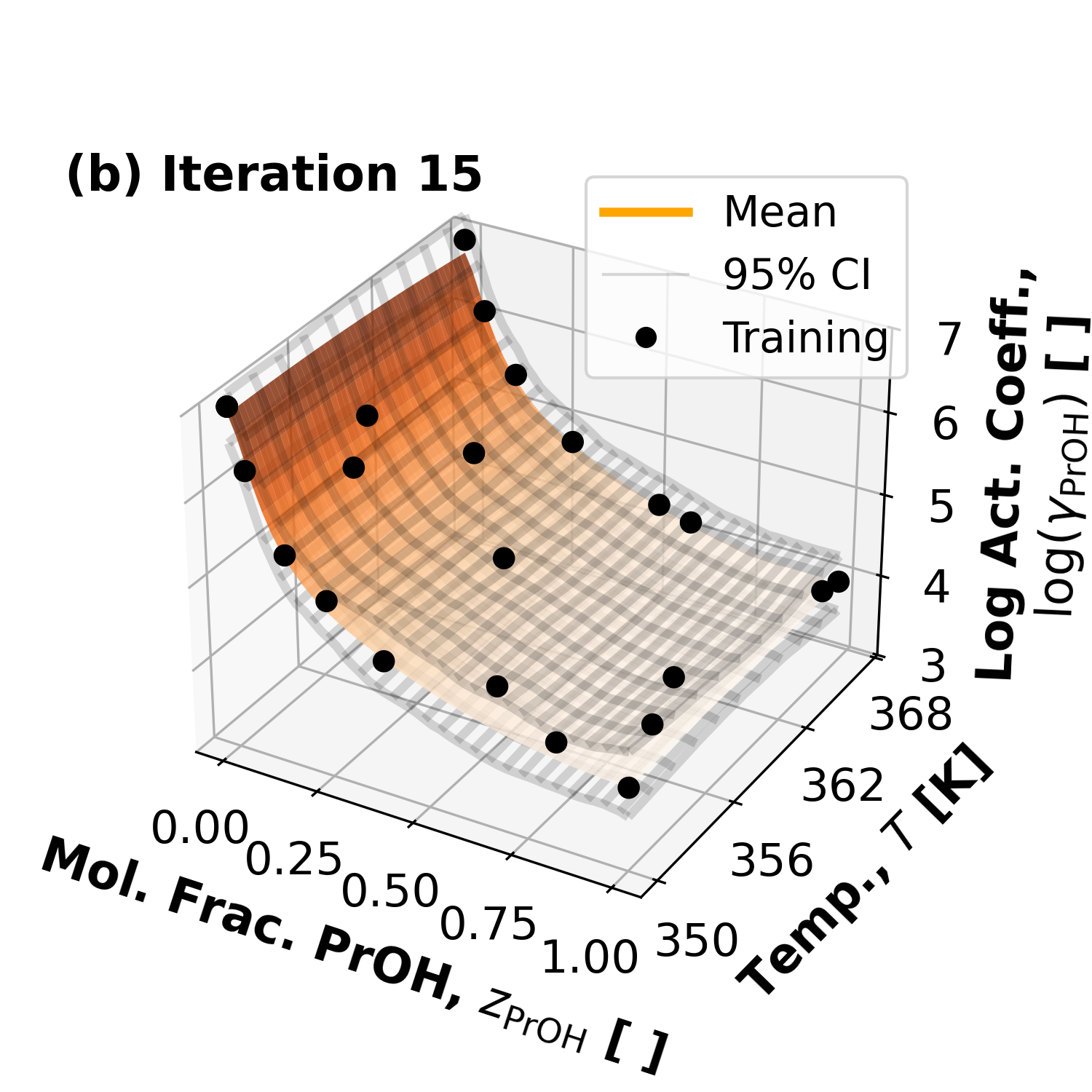}
        \caption*{}
    \end{subfigure}
    \caption{Evolution of the activity coefficient surrogate surface over BITS for GAPS
    iterations. (a) GP posterior surface after iteration 1, and (b) after iteration 15,
    showing the predicted \proh log activity coefficient, $\log(\actcoeff_{\proh})$ [~\,], as a
    function of mole fraction of \proh, $\molefrac_\proh$ [~\,], and temperature, $T$ [K]. The orange surface represents the posterior mean, the grey wireframes indicate the 95\%
    credible interval (CI), and black circles denote training data.}
    \label{fig: gppost2d}
\end{figure}

Figure~\ref{fig: gppost2d} presents the activity coefficient posterior surface at two representative stages of the iterative sampling. This visualization provides a qualitative basis for assessing how the surrogate model evolves across iterations. 

\begin{figure}
    \centering
    \begin{subfigure}{0.4\textwidth}
        \centering
        \includegraphics[width = \textwidth]{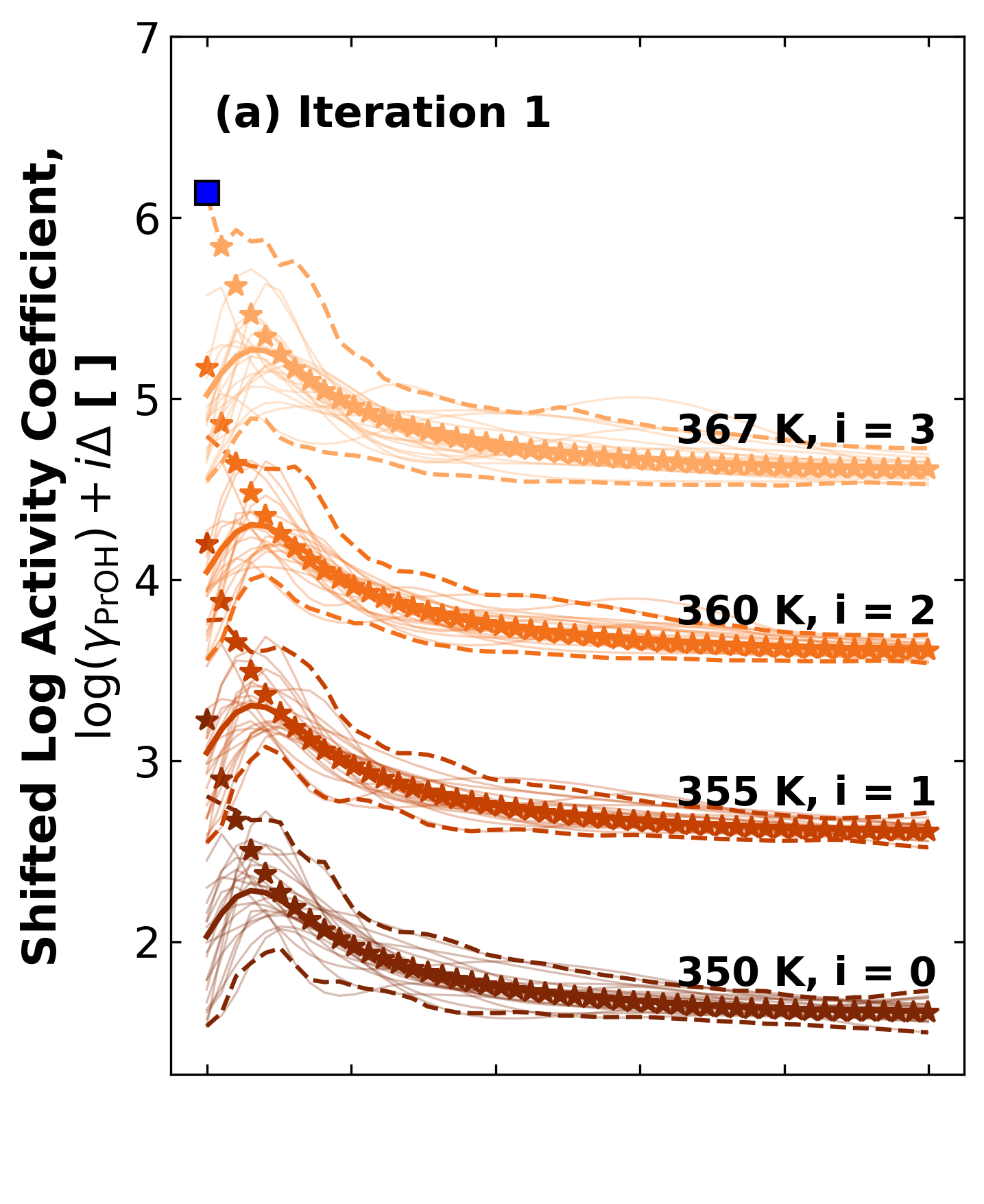}
        \caption*{}
    \end{subfigure}
    \begin{subfigure}{0.4\textwidth}
        \centering
        \includegraphics[width = \textwidth]{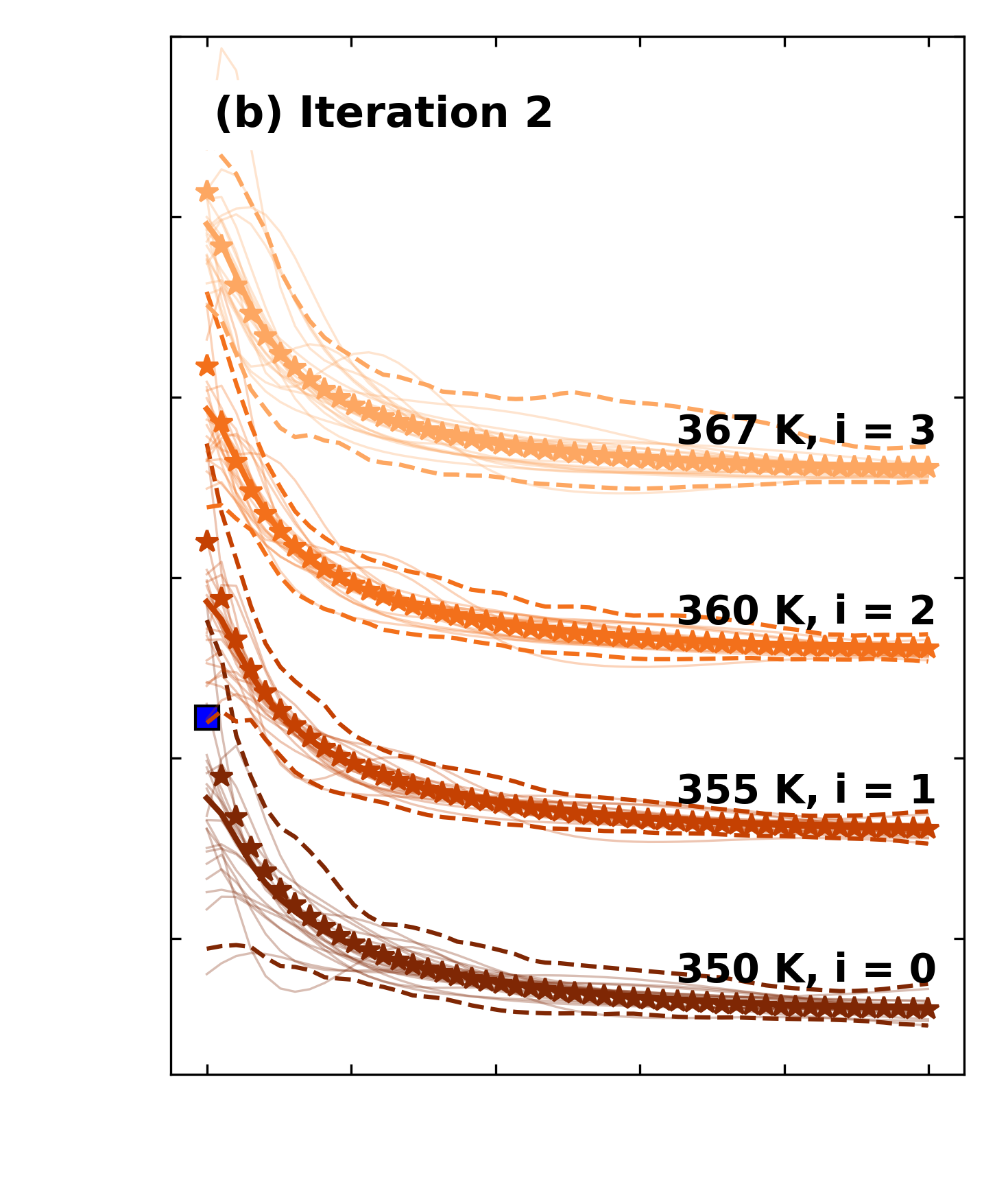}
        \caption*{}
    \end{subfigure}
    \begin{subfigure}{0.4\textwidth}
        \centering
        \includegraphics[width = \textwidth]{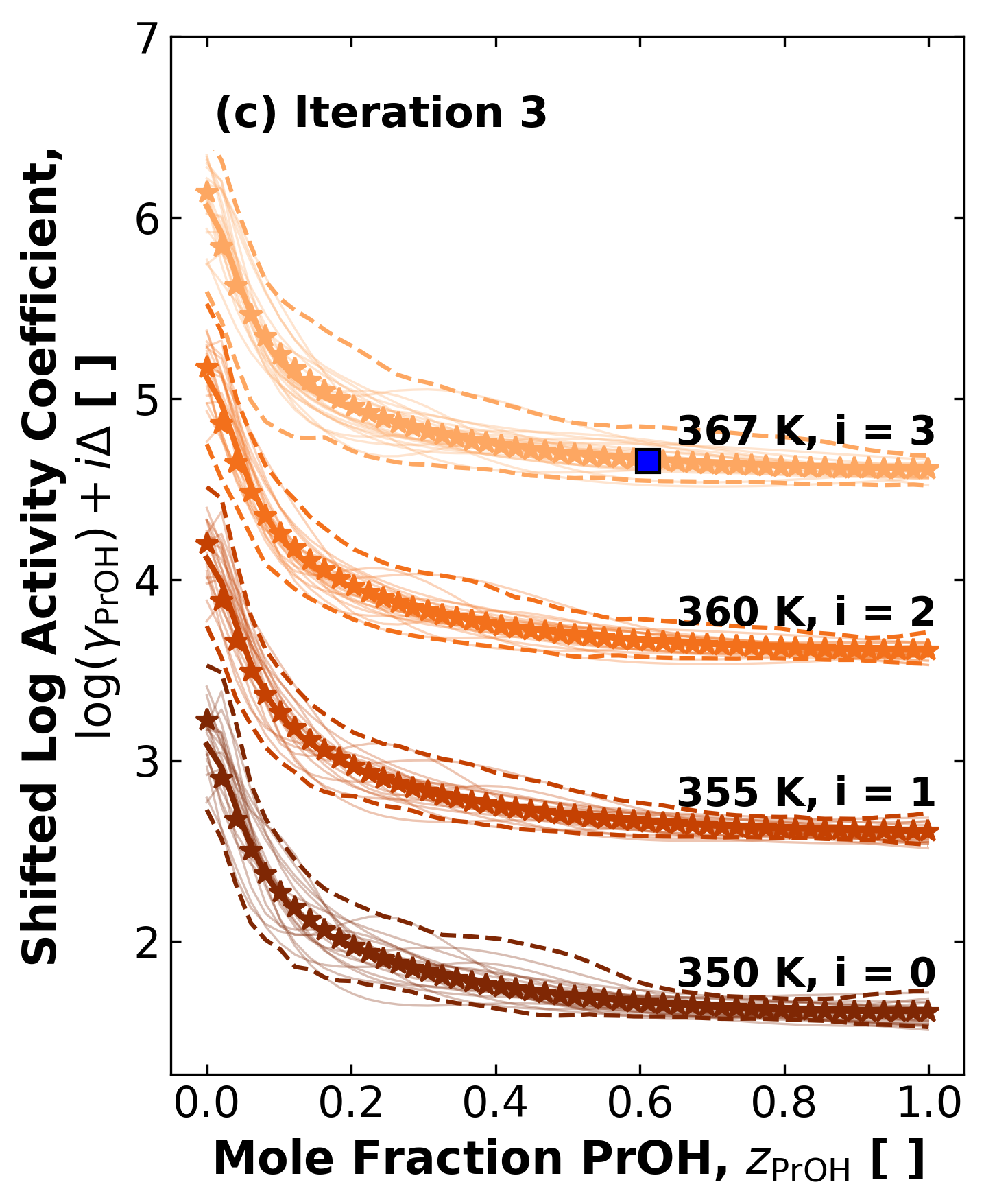}
        \caption*{}
    \end{subfigure}
    \begin{subfigure}{0.4\textwidth}
        \centering
        \includegraphics[width = \textwidth]{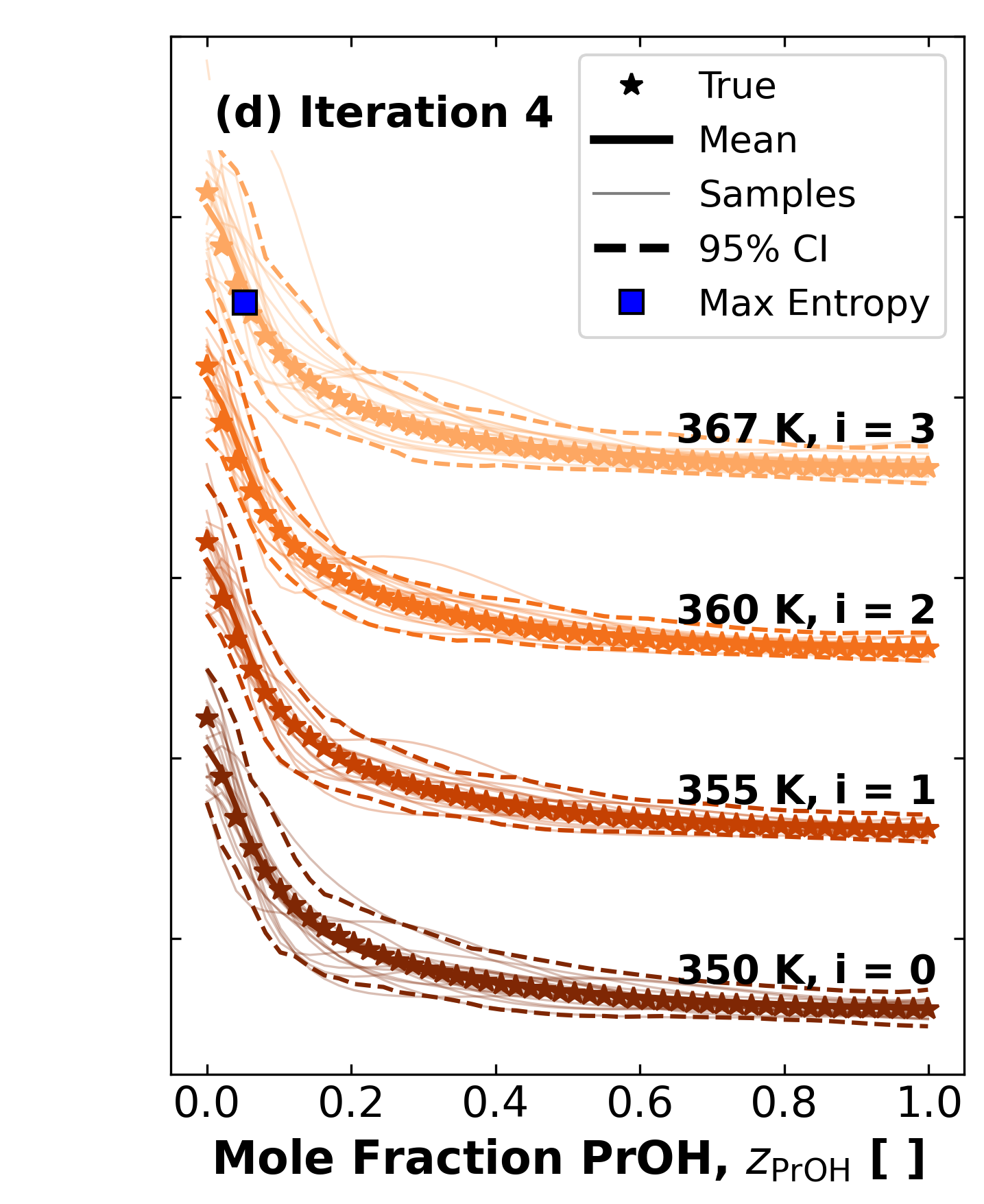}
        \caption*{}
    \end{subfigure}
    \caption{Isotherms of the activity coefficient models. Panels (a)–(d) correspond to iterations 1 through 4. The vertical axis shows the predicted logarithm of the activity coefficient for PrOH with an offset, $\log(\gamma_{\mathrm{PrOH}}) + i\Delta \,[~\,]$, as a function of the mole fraction of PrOH, $z_{\mathrm{PrOH}}~[~\,]$. Different shades of orange and indices for offsets $i$ represent distinct isotherms (350–367 K). Solid lines indicate the surrogate model posterior mean, while dashed lines denote the 95\% credible interval (CI). Star markers represent the ground truth Wilson model, and the blue square highlights the maximum entropy acquisition point selected at each iteration.}
    \label{fig: gppost1d}
\end{figure}

To further illustrate the evolution of the activity coefficient surrogates' predictive behavior, Figure~\ref{fig: gppost1d} presents one-dimensional slices (isotherms) of draws from the activity coefficient posterior. These plots complement the 2D surface (Fig.~\ref{fig: gppost2d}). The inclusion of individual posterior samples, credible intervals, and ground truth values enables a direct assessment of prediction accuracy and uncertainty calibration over time.

Figure~\ref{fig: gppost1d} highlights how the GP surrogate decreases systematic bias as BITS for GAPS progresses. At iteration one (Fig.~\ref{fig: gppost1d}a), the predicted means (solid lines) notably deviate from the Wilson (ground truth) model values, particularly at the dilute limit where the activity coefficient is largest. This discrepancy is accompanied by wider CIs, indicating greater uncertainty. In contrast, by iteration four (Fig.~\ref{fig: gppost1d}a), the predicted mean aligns much more closely with the Wilson model values across the full range of mole fractions. The CIs have also narrowed substantially, reflecting increased model confidence.

\subsubsection{Hybrid model enables distillation system design.}
\begin{figure}
    \centering
    \begin{subfigure}{0.49\textwidth}
        \centering
        \includegraphics[width = \textwidth]{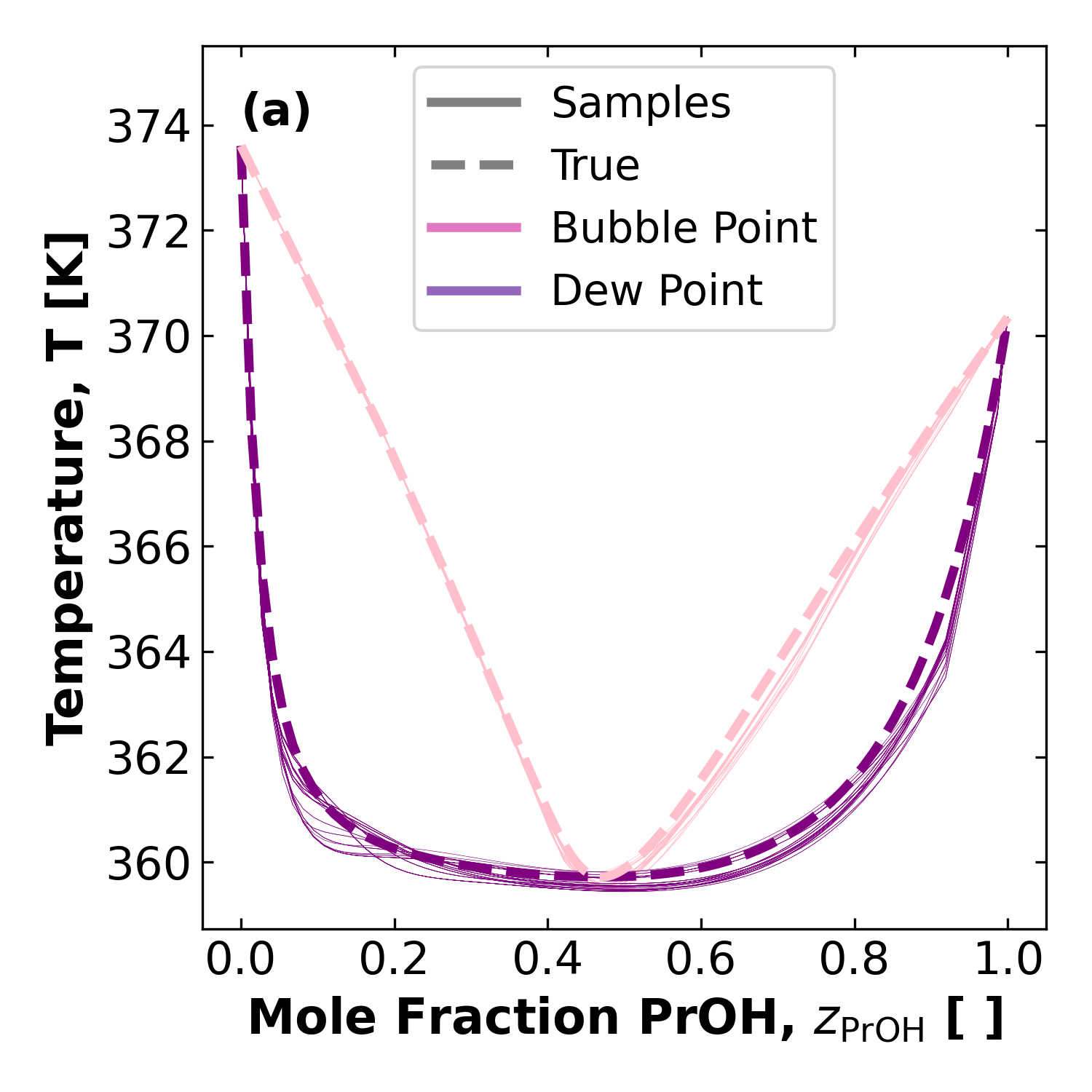}
        \caption*{}
        \label{fig: txy}
    \end{subfigure}
    \begin{subfigure}{0.49\textwidth}
        \centering
        \includegraphics[width = \textwidth]{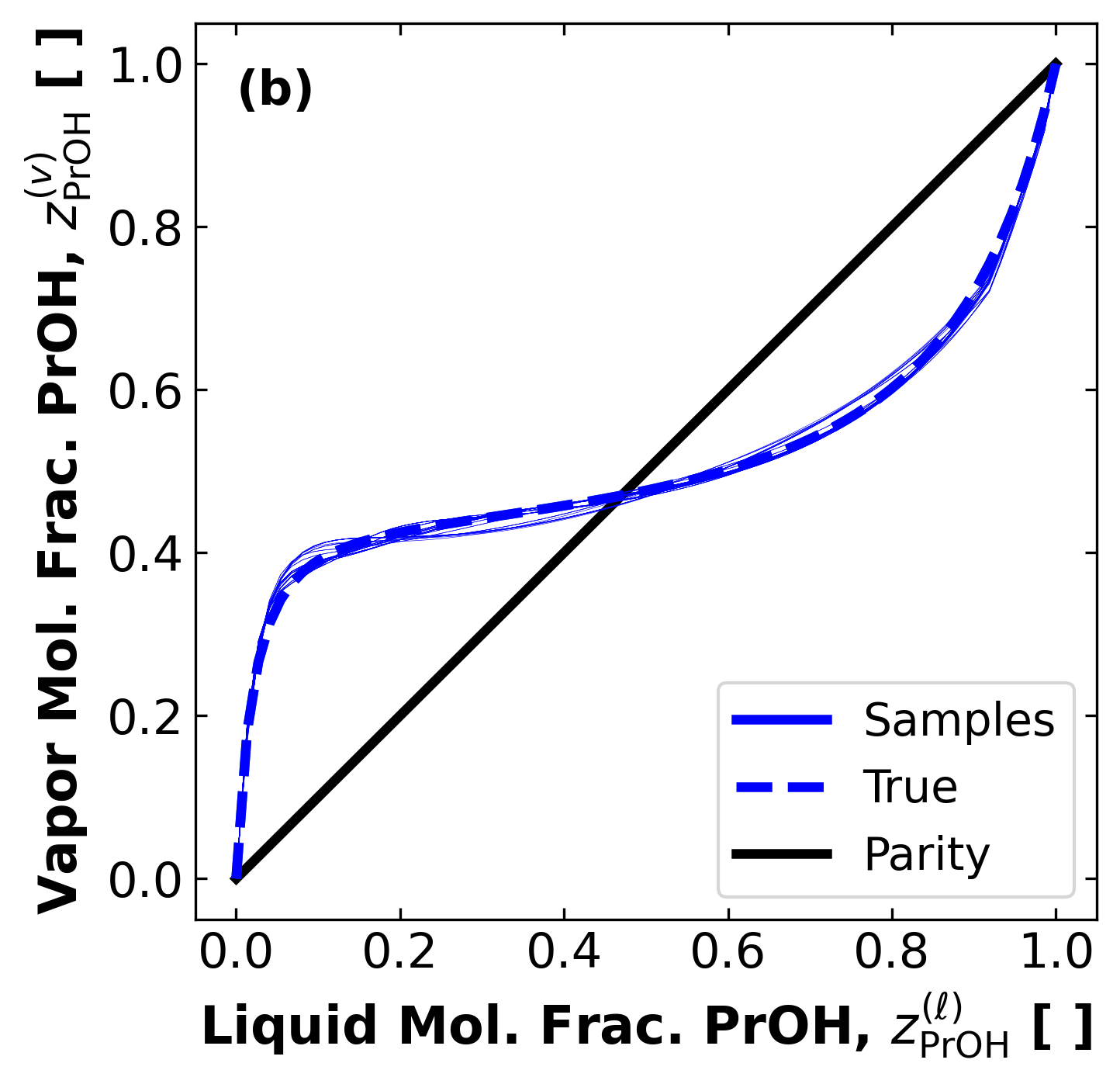}
        \caption*{}
        \label{fig: xy}
    \end{subfigure}
    \caption{VLE phase diagrams for \water--\proh system. (a) Dew point (purple) and bubble point (pink) temperature, \temp [K], vs. overall mole fraction of \proh, $\molefrac_{\proh}$ [~\,]. (b) Vapor mole fraction of \proh, $\molefrac_{\proh}^{(v)}$ [~\,], vs. liquid mole fraction of \proh, $\molefrac_{\proh}^{(\ell)}$ [~\,], with a parity line (black). In both figures, dashed lines indicate the ground truth, and narrow solid lines represent the solutions to the optimization problem under 50 realizations of the posterior mean.}
    \label{fig: phasediagrams}
\end{figure}

\begin{figure}[ht]
    \centering
    \begin{subfigure}{0.4\textwidth}
        \centering
        \includegraphics[width=\textwidth]{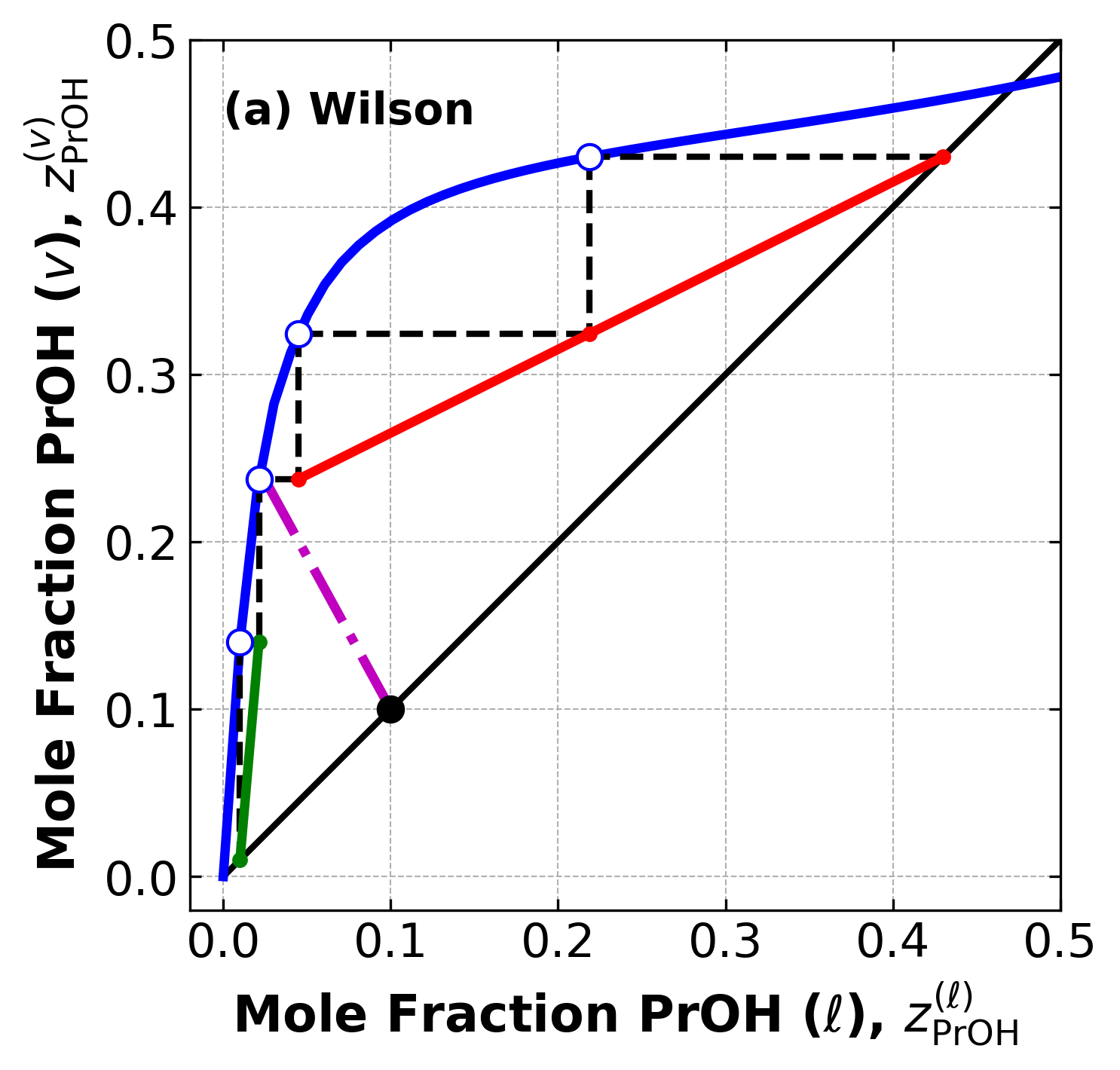}
        \caption*{}
    \end{subfigure}
    \begin{subfigure}{0.4\textwidth}
        \centering
        \includegraphics[width=\textwidth]{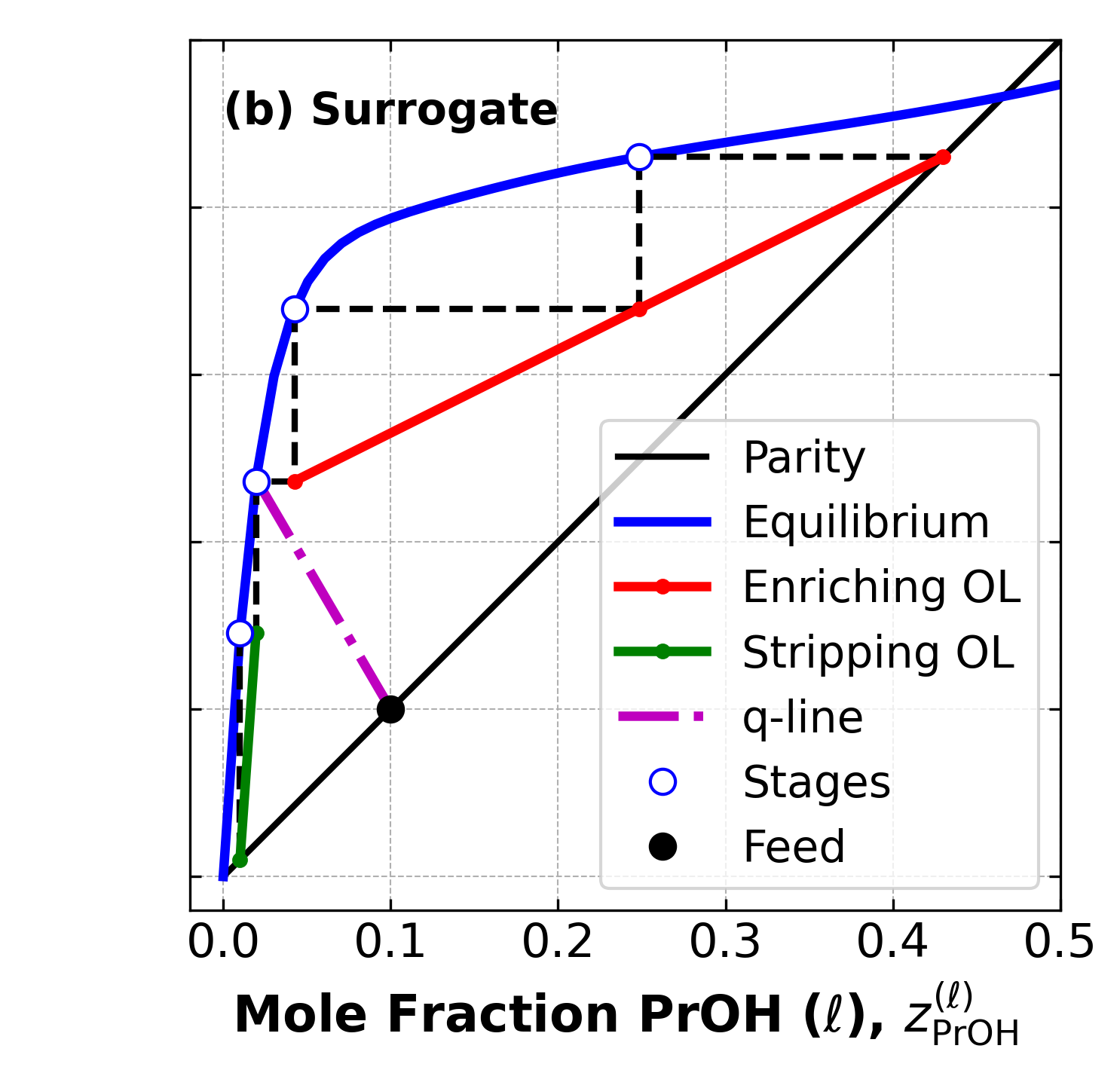}
        \caption*{}
    \end{subfigure}
    
    \begin{subfigure}{\textwidth}
        \centering
        \begin{tabular}{c c c c c}
            \hline
            \textbf{Stage} & \multicolumn{2}{c}{\textbf{Liquid Mole Fraction \proh}} & \multicolumn{2}{c}{\textbf{Vapor Mole Fraction \proh}} \\
            \cline{2-3} \cline{4-5}
             & Wilson & Surrogate & Wilson & Surrogate \\
            \hline
            1 & 0.22 & 0.25 & 0.43 & 0.43 \\
            2 & 0.05 & 0.05 & 0.32 & 0.34 \\
            3 & 0.03 & 0.03 & 0.24 & 0.24 \\
            4 & 0.02 & 0.02 & 0.14 & 0.15 \\
            \hline
        \end{tabular}
        \caption*{(c)}
    \end{subfigure}

    \caption{McCabe–Thiele diagrams for distillation design using 
    (a) the ground-truth Wilson activity coefficient model, 
    (b) the activity coefficient surrogate, and 
    (c) comparison of equilibrium mole fractions of \proh in liquid and vapor phases across theoretical stages. 
    The red and green lines represent the enriching and stripping operating lines, respectively. The blue curve shows the VLE, while the black diagonal line indicates the parity line. Dashed lines illustrate the theoretical equilibrium stages, and the circular markers indicate the compositions at each stage. The magenta dashed-dotted line shows the q-line.}
    \label{fig: distillation}
\end{figure}

After 15 iterations, the surrogate model posterior was used to inform distillation system design. Figure~\ref{fig: phasediagrams} shows temperature-composition and vapor-liquid composition phase envelopes constructed using the GP surrogate and Wilson (ground truth) model. Figure~\ref{fig: distillation} shows the McCabe-Thiele diagram of the GP Surrogate and Wilson model.

Figures~\ref{fig: phasediagrams} and~\ref{fig: distillation} show excellent agreement between the surrogate and ground truth model. Figure~\ref{fig: distillation} shows that both models yield to the same column design: four theoretical stages with the feed introduced on stage three. The stage compositions, however, are not always identical—the enriching section and the vicinity of the feed stage show small shifts, reflecting the surrogate’s local prediction errors (Fig.~\ref{fig: distillation}c).

\subsubsection{Markov Chain Monte Carlo diagnostics affirm the inference task is well-posed.}
\begin{figure} 
    \centering \includegraphics[width=\linewidth]{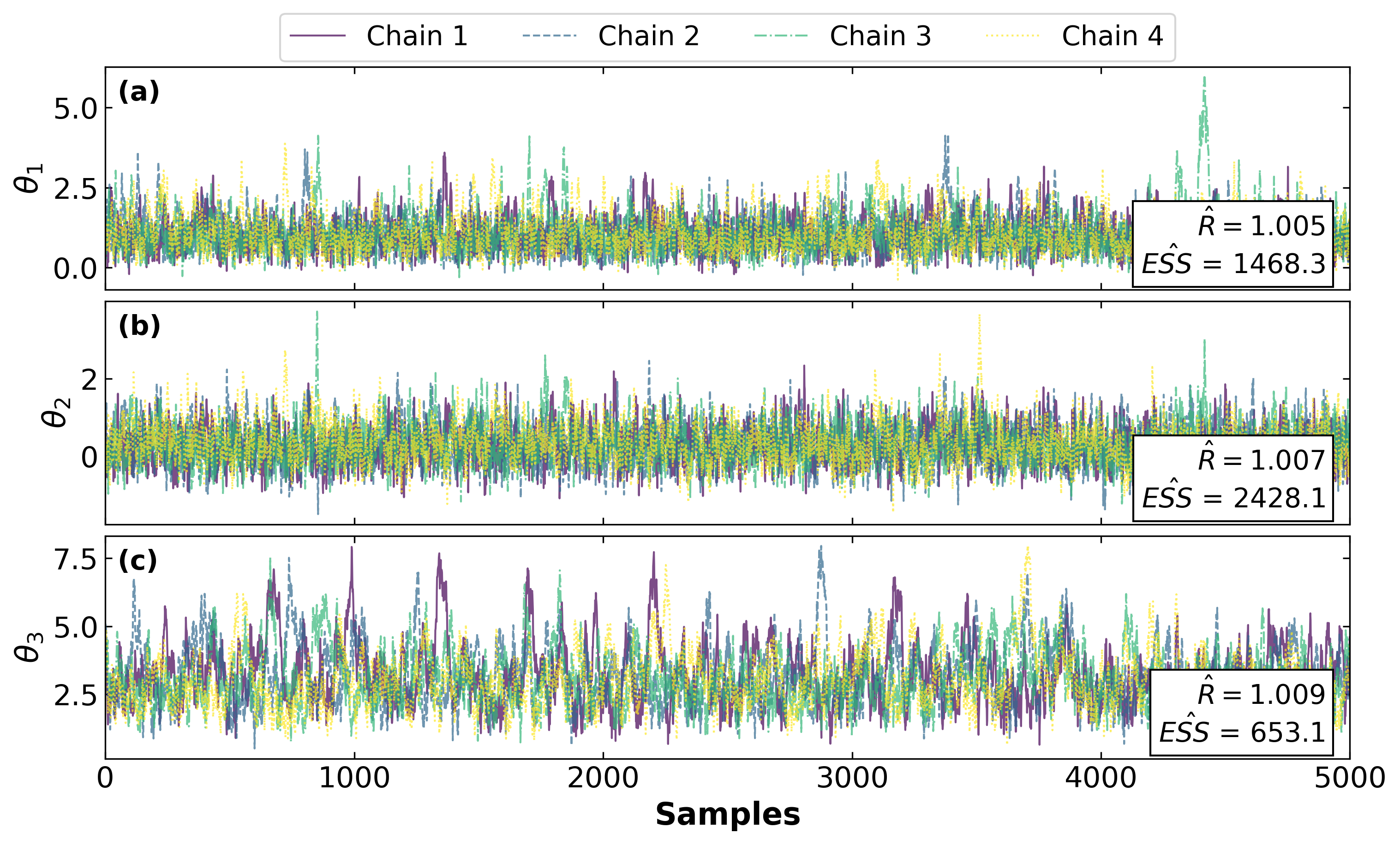} 
    \caption{Trace plots for latent hyperparameters: (a) kernel standard deviation $\hyperparam_1$, (b) mole fraction length scale $\hyperparam_2$, and (c) temperature length scale $\hyperparam_3$. Colors and linestyles denote different Hamiltonian Monte Carlo chains. Gelman–Rubin statistics ($\hat{R}$) are reported in the bottom right for each trajectory.} 
    \label{fig: trace} 
\end{figure}

\begin{figure}
    \centering
    \includegraphics[width=\linewidth]{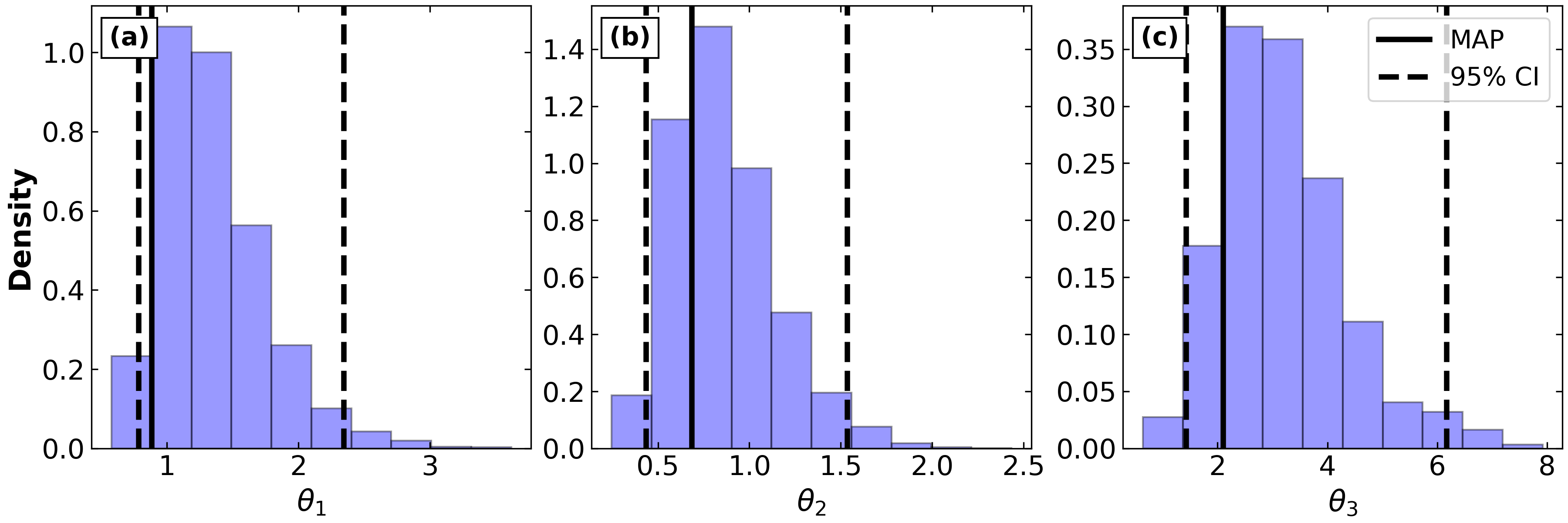}
    \caption{Marginal posterior distributions of the latent hyperparameters: (a) kernel standard deviation $\hyperparam_1$, (b) mole fraction length scale $\hyperparam_2$, and (c) temperature length scale $\hyperparam_3$. Solid black lines indicate the \textit{maximum a posteriori} (MAP) values; dashed black lines denote the 95\% credible interval (CI).}
    \label{fig: marginals}
\end{figure}

\begin{figure}
    \centering
    \includegraphics[width=\linewidth]{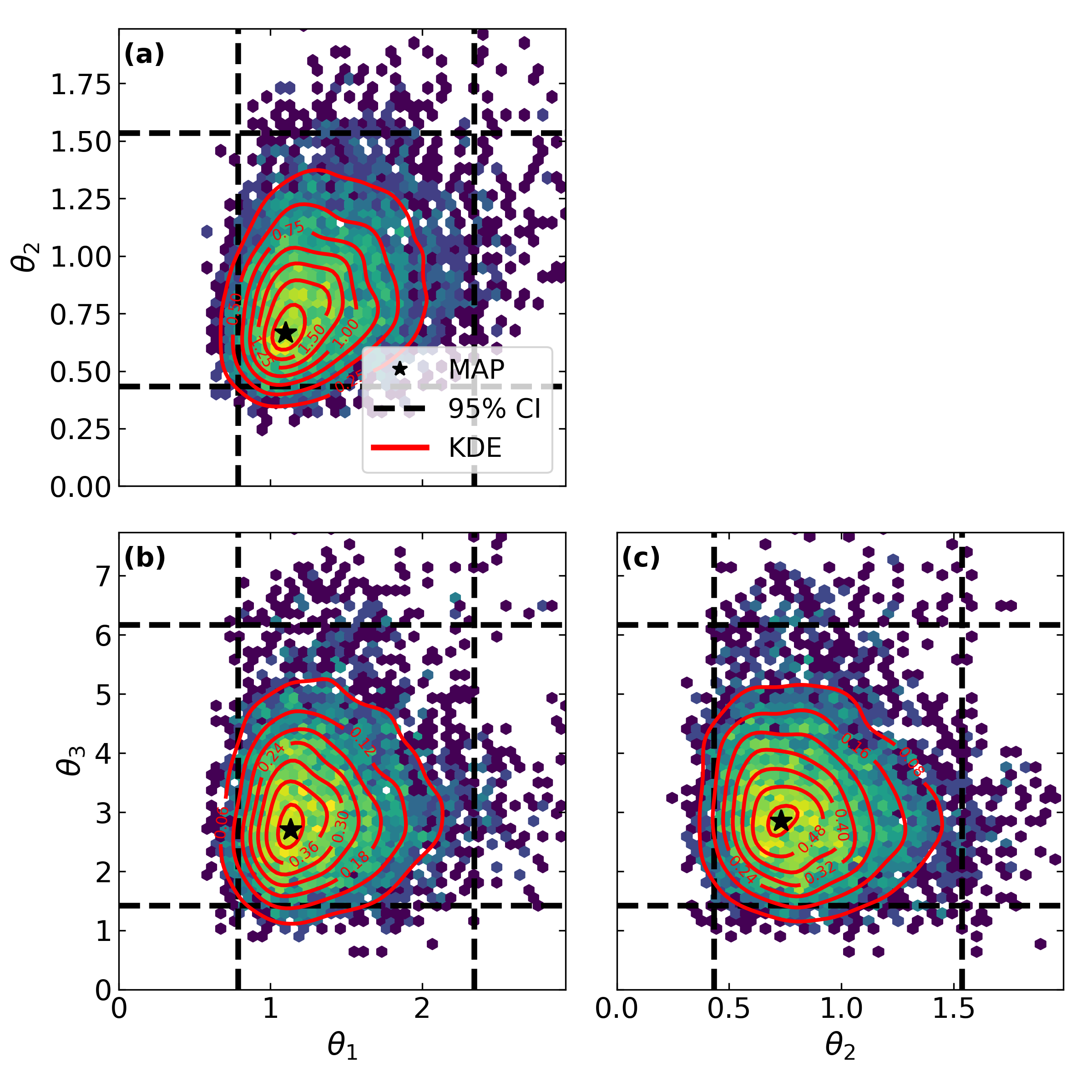}
    \caption{Pairwise marginal distributions of the latent hyperparameters: (a) mole fraction length scale $\hyperparam_2$ vs. kernel standard deviation $\hyperparam_1$, (b) temperature length scale $\hyperparam_3$ vs. kernel standard deviation $\hyperparam_1$, and (c) temperature length scale $\hyperparam_3$ vs. mole fraction length scale $\hyperparam_2$. Black stars indicate the \textit{maximum a posteriori} (MAP) estimates; dashed black lines represent the 95\% credible interval (CI); red contours denote kernel density estimates (KDEs).}
    \label{fig: jointmarginals}
\end{figure}

For completeness, we present and discuss the MCMC diagnostics. Figure~\ref{fig: trace} presents trace plots from HMC sampling of the latent GP hyperparameters. Figure~\ref{fig: marginals} presents histograms of the hyperparameter posterior samples from the first chain of the HMC sampler. Figure~\ref{fig: jointmarginals} shows joint marginal samples from the hyperparameter posterior. KDEs quantify the density for visualization purposes.

The trace plots in Figure~\ref{fig: trace} indicate adequate mixing and convergence to the posterior distribution. For each hyperparameter, the chains explore a similar region of parameter space and exhibit frequent transitions between values, avoiding long periods of stasis. The chains appear well-overlapped, suggesting minimal autocorrelation and good exploration of the posterior. In all cases, the Gelman–Rubin statistic was less than 1.1, indicating that the chains likely converged to the target posterior distribution. The effective sample sizes were 1468, 2428, and 653 for the three hyperparameters, respectively, indicating that the posterior was reasonably well explored, with some parameters exhibiting slower mixing due to stronger posterior correlations.

The posterior distributions in Figure~\ref{fig: marginals} suggest that the hyperparameters are sufficiently well-constrained to support continued use of the predictive model. All posterior distributions appear to be unimodal. The inferred hyperparameters in Figure~\ref{fig: marginals} affirm physically meaningful characteristics of prior beliefs placed on the hyperparameters. In particular, the relative magnitudes of the length scales (Figures~\ref{fig: marginals}b and~\ref{fig: marginals}c) suggest that the response surface varies comparably with respect to the scaled \proh mole fraction and the normalized temperature. While similar posterior magnitudes are numerically convenient in the transformed input space, these results align with prior physical intuition when interpreted in the original units. Specifically, if one were to reverse the input transformations, the shorter effective length scale in the mole fraction dimension indicates that the activity coefficient changes more rapidly with composition than with temperature. This observation is consistent with known thermodynamic behavior of the \water--\proh mixture. The activity coefficient of \proh exhibits sharp deviations from ideality at low mole fractions. In contrast, the influence of temperature on intermolecular interactions is more gradual and typically becomes significant only near infinite dilution~\cite{murti1958, perry1973}.

Figure~\ref{fig: jointmarginals} reveals a sufficiently concentrated posterior landscape to support reliable inference. In particular, there are no strong degeneracies across the joint distributions, and the MAP estimates fall within well-defined, high-density regions. The joint distribution of the kernel standard deviation and the mole fraction length scale ($\hyperparam_1$ vs. $\hyperparam_2$); Fig.~\ref{fig: jointmarginals}a) shows a mild positive correlation, suggesting some coupling between these parameters, though not to a degree that would indicate degeneracy.

\FloatBarrier

\section{Conclusions} 
\label{sec: conclusions}
This work introduces BITS for GAPS, a framework for information-theoretic sequential design in a Bayesian hierarchical GP setting. Existing information-theoretic experimental design methods and entropy-based acquisition functions for GPs often use fixed or point-estimated hyperparameters. In contrast, BITS for GAPS propagates hyperparameter uncertainty into the acquisition function using hierarchical Bayesian modeling. Explicitly addressing hyperparameter uncertainty motivates the main methodological contributions of BITS for GAPS: a new data-acquisition objective and corresponding analytical approximations.

We demonstrate the framework on a vapor–liquid equilibrium case study by constructing a hierarchical GP surrogate for latent activity coefficients and embedding it in a hybrid distillation model via extended Raoult’s law. This example illustrates how to incorporate partial physical knowledge into a probabilistic surrogate and how entropy-based design can guide data acquisition in regions of high model uncertainty. Over successive iterations, the surrogate yields more consistent predictions and physically meaningful uncertainty quantification, thereby supporting downstream analyses such as phase-envelope construction and theoretical-stage estimation. A limitation of this study is that these findings are based on the empirical results from the numerical example. Future work may quantify the performance of BITS for GAPS for other case studies.

The primary contribution of this work is methodological: our focus is on the hierarchical GP approach, which generalizes standard GP models and reduces to them when hyperparameter uncertainty is removed. BITS for GAPS complements standard GP data acquisition strategies and is best suited to cases with significant hyperparameter uncertainty or prior knowledge. 

Entropy-based acquisition functions are computationally expensive, especially for high-dimensional input spaces or with large datasets. The proposed closed-form Taylor approximation helps address these computational challenges. Additional strategies to improve the scalability of GP inference (e.g., inducing points) may be considered as future work. Similar, BITS for GAPS could be extended to parallel or batch experiment design. Further analysis may clarify the tightness and behavior of the proposed entropy bounds and explore how BITS for GAPS performs with standard GP iterative sampling methods on a case-by-case basis. Overall, BITS for GAPS provides a foundation for principled information-theoretic design in hierarchical surrogate models and enables uncertainty-aware data acquisition in hybrid physical–statistical systems.


\section{Acknowledgments}
The authors acknowledge support from the National Science Foundation via Award CBET-1917474.

\section{Declaration of Competing Interest}
The authors declare that they have no known competing financial interests or personal relationships that could have appeared to influence the work reported in this paper.

\section{Data Availability}
Data for this work are available upon request.

\section{Declaration of generative AI and AI-assisted technologies in the manuscript preparation process.}
During the preparation of this work, the authors used ChatGPT (versions GPT-4 and GPT-5, developed by OpenAI) to assist with editing the manuscript text and debugging code. After using this tool, the authors reviewed and revised the content as necessary and take full responsibility for the final version of the manuscript.

\clearpage
\bibliographystyle{elsarticle-num-names}
\bibliography{ref.bib}

\newpage
\appendix

\setcounter{section}{1}
\renewcommand{\thesection}{SI Section-\arabic{section}}

\renewcommand{\thesubsection}{SI-\arabic{subsection}}

\setcounter{equation}{0}
\renewcommand{\theequation}{SI-\arabic{equation}}

\setcounter{figure}{0}
\renewcommand{\thefigure}{SI\arabic{figure}}

\setcounter{table}{0}
\renewcommand{\thetable}{SI-\arabic{table}}

\setcounter{footnote}{0}

\setcounter{page}{1}
\renewcommand{\thepage}{SI-\arabic{page}}

\nolinenumbers
{\Large
\begin{center}
Supplementary Information\\ BITS for GAPS: Bayesian Information-Theoretic Sampling for hierarchical GAussian Process Surrogates
\end{center}
}

\begin{center}
Kyla D. Jones, Alexander W. Dowling\footnote{corresponding author: adowling@nd.edu}

\emph{Department of Chemical and Biomolecular Engineering, University of Notre Dame, Notre Dame, IN 46556, USA}
\end{center}

\begin{center}
    \today
\end{center}
\subsection{Absolute moments of the standard normal distribution.}
\label{subsection: propositionproofsi}
Let $\stdnormrv_* \sim \mathcal{N}(0,1)$. We compute the absolute moment $\Exp{|\stdnormrv_*|^{\taylorterms+1}}$ in closed form. By symmetry of the standard normal density,
\begin{gather*}
    \Exp{|\stdnormrv_*|^{\taylorterms+1}} = \frac{1}{\sqrt{2\pi}}\int_{-\infty}^\infty |\stdnormrv_*|^{\taylorterms+1} e^{-\stdnormrv_*^2/2}\,\d\stdnormrv_* = \frac{2}{\sqrt{2\pi}}\int_0^\infty \stdnormrv_*^{\taylorterms+1}e^{-\stdnormrv_*^2/2}\,\d\stdnormrv_*.
\end{gather*}
Let $\usub=\stdnormrv_*^2/2$, so that $\d\usub = \stdnormrv_*\d\stdnormrv_*$. Substituting yields
\begin{gather*}
    \Exp{|\stdnormrv_*|^{\taylorterms+1}} = \frac{2}{\sqrt{2\pi}}\int_0^\infty (2\usub)^{\taylorterms/2}e^{-\usub}\,\d\usub = \frac{2^{\taylorterms/2 + 1}}{\sqrt{2\pi}}\int_0^\infty \usub^{\taylorterms/2}e^{-\usub}\,\d\usub.
\end{gather*}
Recalling the definition of the gamma function, 
\begin{gather*}
    \Gamma(t) = \int_0^\infty \usub^{t-1}e^{-\usub}\d \usub,
\end{gather*}
we obtain
\begin{gather*}
    \Exp{|\stdnormrv_*|^{\taylorterms+1}} = \frac{2^{\taylorterms/2 + 1}}{\sqrt{2\pi}} \Gamma\left(\frac{\taylorterms+2}{2}\right).
\end{gather*}

\subsection{Derivation of cross-overlap.}
\label{subsection: crossoverlapsi}
We derive the cross-overlap $\crossoverlap_{\sampleindex, \sampleindex'}$, defined as the integral of the product of two univariate Gaussian p.d.f.'s. Let
\begin{gather*}
    p_\sampleindex(\latentfxn_*) = %
    \frac{1}{\sqrt{2\pi \posteriorstd_\sampleindex^2}}%
    \exp\left[-\frac{1}{2} \left(\frac{\latentfxn_* - \posteriormean_\sampleindex}{\posteriorstd_\sampleindex}\right)^2\right],
    \quad
    p_\sampleindex(f) = %
    \frac{1}{\sqrt{2\pi \posteriorstd_{\sampleindex'}^2}}%
    \exp\left[-\frac{1}{2} \left(\frac{\latentfxn_* - \posteriormean_{\sampleindex'}}{\posteriorstd_{\sampleindex'}}\right)^2\right].
\end{gather*}
The product of the two densities is 
\begin{gather*}
    p_\sampleindex(\latentfxn_*) p_{\sampleindex'}(\latentfxn_*)%
    = \frac{1}{\sqrt{2\pi \posteriorstd_\sampleindex^2 \posteriorstd_{\sampleindex'}^2}}%
    \exp\left( -\frac{1}{2} \left[\left(\frac{\latentfxn_* - \posteriormean_\sampleindex}{\posteriorstd_\sampleindex}\right)^2 + \left(\frac{\latentfxn_* - \posteriormean_{\sampleindex'}}{\posteriorstd_{\sampleindex'}}\right)^2\right]\right).
\end{gather*}
%
%
Expanding the quadratic terms yields
\begin{gather*}
    \left(\frac{\latentfxn_* - \posteriormean_\sampleindex}{\posteriorstd_\sampleindex}\right)^2 - \left(\frac{\latentfxn_* - \posteriormean_{\sampleindex'}}{\posteriorstd_{\sampleindex'}}\right)^2%
    =
     \left(\frac{1}{\posteriorstd_\sampleindex^2} + \frac{1}{\posteriorstd_{\sampleindex'}^2}\right)\latentfxn_*^2 - 2 \left(\frac{\posteriormean_\sampleindex}{\posteriorstd_\sampleindex^2} + \frac{\posteriormean_{\sampleindex'}}{\posteriorstd_{\sampleindex'}^2}\right)\latentfxn_*  + \left(\frac{\posteriormean_\sampleindex^2}{\posteriorstd_\sampleindex^2} + \frac{\posteriormean_{\sampleindex'}^2}{\posteriorstd_{\sampleindex'}^2} \right)
\end{gather*}
Define the coefficients 
\begin{gather*}
    A := \frac{1}{\posteriorstd_\sampleindex^2} + \frac{1}{\posteriorstd_{\sampleindex'}^2}, %
    \quad %
    B := \frac{\posteriormean_\sampleindex}{\posteriorstd_\sampleindex^2} + \frac{\posteriormean_{\sampleindex'}}{\posteriorstd_{\sampleindex'}^2}.
\end{gather*}
Then the product density can be written as
\begin{gather*}
    p_\sampleindex(\latentfxn_*) p_{\sampleindex'}(\latentfxn_*)%
    = \frac{1}{\sqrt{2\pi \posteriorstd_\sampleindex^2 \posteriorstd_{\sampleindex'}^2}}%
    \exp\left[ -\frac{1}{2} \left( A \latentfxn_*^2 - 2B \latentfxn_* + \frac{\posteriormean_\sampleindex^2}{\posteriorstd_\sampleindex^2} + \frac{\posteriormean_{\sampleindex'}^2}{\posteriorstd_{\sampleindex'}^2}  \right)\right].
\end{gather*}
%
%
Completing the square in $\latentfxn_*$,
\begin{gather*}
    A \latentfxn_*^2 - 2B \latentfxn_*%
    =  A\left(\latentfxn_* - \frac{B}{A}\right)^2 - \frac{B^2}{A}.
\end{gather*}
Substituting this expression gives
\begin{gather*}
    p_\sampleindex(\latentfxn_*) p_{\sampleindex'}(\latentfxn_*)%
    =
    \frac{1}{\sqrt{2\pi \posteriorstd_\sampleindex^2 \posteriorstd_{\sampleindex'}^2}}%
    \exp\left[%
    -\frac{A}{2}%
    \left(%
    \latentfxn_* - \frac{B}{A}%
    \right)^2%
    \right]%
    \exp \left [%
    -\frac{1}{2}%
    \left(%
    \frac{\posteriormean_\sampleindex^2}{\posteriorstd_\sampleindex^2}%
    + \frac{\posteriormean_{\sampleindex'}^2}{\posteriorstd_{\sampleindex'}^2}%
    - \frac{B^2}{A}
    \right)
    \right].
\end{gather*}
Introduce 
\begin{gather*}
    m:=\frac{B}{A}=\frac{\posteriormean_\sampleindex/\posteriorstd_\sampleindex^2 + \posteriormean_{\sampleindex'}/\posteriorstd_{\sampleindex'}^2}{1/\posteriorstd_\sampleindex^2 + 1/\posteriorstd_{\sampleindex'}^2},%
    \quad %
    s^2 :=A^{-1} = \left(\frac{1}{\posteriorstd_\sampleindex^2} + \frac{1}{\posteriorstd_{\sampleindex'}^2} \right)^{-1}.
\end{gather*}
Then
\begin{gather*}
    p_\sampleindex(\latentfxn_*) p_{\sampleindex'}(\latentfxn_*)%
    =\frac{1}{\sqrt{2\pi \posteriorstd_\sampleindex^2 \posteriorstd_{\sampleindex'}^2}}%
    \exp\left[%
    -\frac{1}{2}\left(\frac{\latentfxn_* - m}{s}\right)^2\right] \exp\left[-\frac{1}{2}\left(\frac{\posteriormean_\sampleindex^2}{\posteriorstd_\sampleindex^2}+\frac{\posteriormean_{\sampleindex'}^2}{\posteriorstd_{\sampleindex'}^2} - \frac{B^2}{A}\right)%
    \right].
\end{gather*}
%
%
The entropy cross overlap is defined as
\begin{gather*}
       \crossoverlap_{\sampleindex, \sampleindex'}%
       := \int p_\sampleindex(\latentfxn_*) p_{\sampleindex'}(\latentfxn_*) \, \d \latentfxn_*. 
\end{gather*}
Substituting the expression above yields
\begin{gather*}
    \crossoverlap_{\sampleindex, \sampleindex'} = %
    \frac{1}{\sqrt{2\pi \posteriorstd_\sampleindex^2 \posteriorstd_{\sampleindex'}^2}} \exp\left[-\frac{1}{2}\left(\frac{\posteriormean_\sampleindex^2}{\posteriorstd_\sampleindex^2}+\frac{\posteriormean_{\sampleindex'}^2}{\posteriorstd_{\sampleindex'}^2} - \frac{B^2}{A}\right)\right] \int \exp\left[-\frac{1}{2}\left(\frac{\latentfxn - m}{s}\right)^2\right] \,\d \latentfxn.
\end{gather*}
The remaining integral is the normalization integral of a Gaussian density,
\begin{gather*}
    \int \exp\left[-\frac{1}{2}\left(\frac{\latentfxn_* - m}{s}\right)^2\right] \,\d \latentfxn_* = \sqrt{2\pi s^2}.
\end{gather*}
Using $s^2 = A^{-1}$ gives
\begin{align*}
    \crossoverlap_{\sampleindex, \sampleindex'}%
    = \frac{1}{\sqrt{2\pi (\posteriorstd_\sampleindex^2 + \posteriorstd_{\sampleindex'}^2)}} \exp\left[-\frac{1}{2}\left(\frac{\posteriormean_\sampleindex^2}{\posteriorstd_\sampleindex^2}+\frac{\posteriormean_{\sampleindex'}^2}{\posteriorstd_{\sampleindex'}^2} - \frac{B^2}{A}\right)\right].
\end{align*}
%
%
We now simplify the term involving $B^2/A$. First note that
\begin{gather*}
    B^2 = \left(\frac{\posteriormean_\sampleindex}{\posteriorstd_\sampleindex^2} + \frac{\posteriormean_{\sampleindex'}}{\posteriorstd_{\sampleindex'}^2} \right)^2 = \frac{\posteriormean_\sampleindex^2}{\posteriorstd_\sampleindex^4} + \frac{\posteriormean_{\sampleindex'}^2}{\posteriorstd_{\sampleindex'}^4} + 2 \frac{\posteriormean_\sampleindex \posteriormean_{\sampleindex'}}{\posteriorstd_\sampleindex^2 \posteriorstd_{\sampleindex'}^2},
\end{gather*}
while
\begin{gather*}
    A = \frac{1}{\posteriorstd_\sampleindex^2} + \frac{1}{\posteriorstd_{\sampleindex'}^2} = \frac{\posteriorstd_\sampleindex^2 + \posteriorstd_{\sampleindex'}^2}{\posteriorstd_\sampleindex^2 \posteriorstd_{\sampleindex'}^2}.
\end{gather*}
Hence
\begin{gather*}
    \frac{B^2}{A} = \left( \frac{\posteriormean_\sampleindex^2}{\posteriorstd_\sampleindex^4} + \frac{\posteriormean_{\sampleindex'}^2}{\posteriorstd_{\sampleindex'}^4} + 2 \frac{\posteriormean_\sampleindex \posteriormean_{\sampleindex'}}{\posteriorstd_\sampleindex^2 \posteriorstd_{\sampleindex'}^2} \right) \frac{\posteriorstd_\sampleindex^2 \posteriorstd_{\sampleindex'}^2}{\posteriorstd_\sampleindex^2 + \posteriorstd_{\sampleindex'}^2} = \frac{\posteriormean^2_\sampleindex \posteriorstd_{\sampleindex'}^2 / \posteriorstd_\sampleindex^2 +\posteriormean^2_{\sampleindex'} \posteriorstd_\sampleindex^2/\posteriorstd_{\sampleindex'}^2 + 2\posteriormean_\sampleindex \posteriormean_{\sampleindex'}}{\posteriorstd_\sampleindex^2  + \posteriorstd_{\sampleindex'}^2}.
\end{gather*}
Now we expand the remaning term in the exponent
\begin{gather*}
    \frac{\posteriormean_\sampleindex^2}{\posteriorstd_\sampleindex^2} + \frac{\posteriormean_{\sampleindex'}^2}{\posteriorstd_{\sampleindex'}^2} %
    = \frac{\posteriormean_\sampleindex^2(\posteriorstd_\sampleindex^2 + \posteriorstd_{\sampleindex'}^2)/\posteriorstd_\sampleindex^2 + \posteriormean_{\sampleindex'}^2(\posteriorstd_\sampleindex^2 + \posteriorstd_{\sampleindex'}^2)/\posteriorstd_{\sampleindex'}^2}{\posteriorstd_\sampleindex^2  + \posteriorstd_{\sampleindex'}^2}
    = \frac{\posteriormean_\sampleindex^2 + \posteriormean_\sampleindex^2 \posteriorstd_{\sampleindex'}^2/\posteriorstd_\sampleindex^2 + \posteriormean_{\sampleindex'}^2 +\posteriormean_{\sampleindex'}^2 \posteriorstd_{\sampleindex}^2/\posteriorstd_{\sampleindex'}^2}{\posteriorstd_\sampleindex^2  + \posteriorstd_{\sampleindex'}^2}.
\end{gather*}
The exponent can be simplified as
\begin{gather*}
    \frac{\posteriormean_\sampleindex^2}{\posteriorstd_\sampleindex^2} + \frac{\posteriormean_{\sampleindex'}^2}{\posteriorstd_{\sampleindex'}^2} - \frac{B^2}{A} %
    = \frac{\posteriormean_\sampleindex^2 -2\posteriormean_\sampleindex \posteriormean_{\sampleindex'} + \posteriormean_{\sampleindex'}^2}{\posteriorstd_\sampleindex^2  + \posteriorstd_{\sampleindex'}^2}
    = \frac{(\posteriormean_\sampleindex - \posteriormean_{\sampleindex'})^2}{\posteriorstd_\sampleindex^2  + \posteriorstd_{\sampleindex'}^2}.
\end{gather*}
%
%
Combining the above expressions, the entropy cross overlap between the two Gaussian densities is
\begin{gather*}
    \crossoverlap_{\sampleindex, \sampleindex'} = \int p_\sampleindex(\latentfxn_*)\,p_{\sampleindex'}(\latentfxn_*)\,\d \latentfxn_* = %
    \frac{1}{\sqrt{2\pi (\posteriorstd_\sampleindex^2 + \posteriorstd_{\sampleindex'}^2)}} \exp\left(-\frac{1}{2}\frac{(\posteriormean_\sampleindex - \posteriormean_{\sampleindex'})^2}{\posteriorstd_\sampleindex^2  + \posteriorstd_{\sampleindex'}^2}\right).
\end{gather*}
This quantity is proportional to a Gaussian density with mean $\posteriormean_{\sampleindex'}$ and variance $\posteriorstd_\sampleindex^2 + \posteriorstd_{\sampleindex'}^2$.
\subsection{Software Requirements}
For completeness, this section lists all of the packages installed in the conda environment used to generate these results on a MacBook Pro M2 2022.
\label{subsection: softreqs}
\input{\FigurePath requirements_items_prefixed}

\subsection{Binary Distillation Model and Solution Procedure}
\label{subsection: mccabethiele}

\paragraph{Scope and assumptions}
We model a steady-state, binary, staged distillation column with a total condenser and a partial reboiler, operating under constant molar overflow. Each stage is ideal, adiabatic, and at vapor--liquid equilibrium, with negligible pressure drop. A single feed enters the column at stage \(n_F\) with molar flow rate \(F\) and light-key mole fraction \(x_F\).

\paragraph{Model specification and unknowns}
The model is specified by: total number of equilibrium stages \(n\) (stage 1 = top tray, stage \(n\) = reboiler), reflux ratio \(R\), distillate composition \(x_D\), bottoms composition \(x_W\), feed flow \(F\), feed composition \(x_F\), feed stage \(n_F\), and feed quality \(q\).  

The unknowns are the stagewise liquid and vapor flow rates \((L_i,V_i)\) and light-key mole fractions \((x_i,y_i)\) for \(i=1,\dots,n\), along with condenser and reboiler flows: distillate \(D\), reflux \(L_0\), vapor to the condenser \(V_1\), vapor from the reboiler \(V_{n+1}\), and bottoms \(W\).

\paragraph{Governing equations}
For each stage \(i=1,\dots,n\), the total and component material balances are:
\begin{equation*}
L_{i-1} + V_{i+1} - L_i - V_i =
\begin{cases}
F, & i=n_F,\\
0, & \text{otherwise},
\end{cases}
\end{equation*}
\begin{equation*}
x_{i-1}\,L_{i-1} + y_{i+1}\,V_{i+1} - x_i\,L_i - y_i\,V_i  =
\begin{cases}
x_F\,F, & i=n_F,\\
0, & \text{otherwise}.
\end{cases}
\end{equation*}

The vapor--liquid equilibrium relation on each stage is given by:
\[
y_i = \phi(x_i),
\]
where \(\phi(\cdot)\) is obtained from tabulated \((x,y)\) data at the operating pressure \(P\), interpolated to arbitrary \(x\) (e.g., monotone cubic interpolation). 

Under constant molar overflow, internal flows are constant except at the feed stage:
\begin{gather*}
L_i - L_{i-1} =
\begin{cases}
- q\,F, & i=n_F,\\
0, & \text{otherwise},
\end{cases} 
\end{gather*}

\begin{align*}
V_{i+1} - V_i &=
\begin{cases}
(1-q)\,F, & i=n_F,\\
0, & \text{otherwise}.
\end{cases}
\end{align*}

The condenser and reboiler satisfy:
\begin{align*}
L_0 &= R\,D, & V_1 &= L_0 + D, & L_n &= V_{n+1} + W, \\
x_D &= x_0, & x_W &= x_n.
\end{align*}

\end{document}

%% file: Figures/requirements_items_prefixed.tex
\begin{center}
\footnotesize
\begin{longtable}{@{}ll@{}}
\toprule
\textbf{Package} & \textbf{Version / Build} \\
\midrule
\endfirsthead
\toprule
\textbf{Package} & \textbf{Version / Build} \\
\midrule
\endhead
\bottomrule
\endfoot
absl-py & 2.1.0=py39hca03da5\_0 \\
assimulo & 3.5.2=py39hcc55131\_0 \\
astunparse & 1.6.3=py\_0 \\
atomicwrites & 1.4.0=py\_0 \\
autograd & 1.7.0=pyhd8ed1ab\_0 \\
blas & 1.0=openblas \\
brotli & 1.1.0=hb547adb\_1 \\
brotli-bin & 1.1.0=hb547adb\_1 \\
brotli-python & 1.0.9=py39h313beb8\_8 \\
bzip2 & 1.0.8=h99b78c6\_7 \\
c-ares & 1.33.1=hd74edd7\_0 \\
ca-certificates & 2024.9.24=hca03da5\_0 \\
cached-property & 1.5.2=py\_0 \\
certifi & 2024.8.30=py39hca03da5\_0 \\
charset-normalizer & 3.3.2=pyhd3eb1b0\_0 \\
check\_shapes & 1.1.1=pyhd8ed1ab\_0 \\
cloudpickle & 3.0.0=py39hca03da5\_0 \\
contourpy & 1.2.1=py39h48c5dd5\_0 \\
cycler & 0.12.1=pyhd8ed1ab\_0 \\
cython & 0.29.37=py39hf3050f2\_0 \\
decorator & 5.1.1=pyhd3eb1b0\_0 \\
deprecated & 1.2.13=py39hca03da5\_0 \\
dm-tree & 0.1.7=py39h313beb8\_1 \\
dropstackframe & 0.1.1=pyhd8ed1ab\_0 \\
flatbuffers & 24.3.25=h313beb8\_0 \\
fonttools & 4.53.1=py39hfea33bf\_0 \\
freetype & 2.12.1=hadb7bae\_2 \\
gast & 0.5.3=pyhd3eb1b0\_0 \\
giflib & 5.2.2=h93a5062\_0 \\
gmp & 6.3.0=h7bae524\_2 \\
google-pasta & 0.2.0=pyhd3eb1b0\_0 \\
gpflow & 2.9.2=pyhd8ed1ab\_0 \\
grpcio & 1.62.2=py39h047a24b\_0 \\
h5py & 3.11.0=nompi\_py39h534c8c8\_102 \\
hdf5 & 1.14.3=nompi\_hec07895\_105 \\
icu & 75.1=hfee45f7\_0 \\
idna & 3.7=py39hca03da5\_0 \\
imageio & 2.36.0=pypi\_0 \\
importlib-metadata & 7.0.1=py39hca03da5\_0 \\
importlib-resources & 6.4.4=pyhd8ed1ab\_0 \\
importlib\_resources & 6.4.4=pyhd8ed1ab\_0 \\
joblib & 1.4.2=py39hca03da5\_0 \\
js2py & 0.74=py39hca03da5\_0 \\
julia & 0.6.2=pypi\_0 \\
juliacall & 0.9.23=pypi\_0 \\
juliapkg & 0.1.13=pypi\_0 \\
keras & 3.5.0=pypi\_0 \\
kiwisolver & 1.4.5=py39hbd775c9\_1 \\
krb5 & 1.21.3=h237132a\_0 \\
lark & 1.1.2=py39hca03da5\_0 \\
lcms2 & 2.16=ha0e7c42\_0 \\
lerc & 4.0.0=h9a09cb3\_0 \\
libabseil & 20240116.2=cxx17\_h313beb8\_0 \\
libaec & 1.1.3=hebf3989\_0 \\
libblas & 3.9.0=23\_osxarm64\_openblas \\
libbrotlicommon & 1.1.0=hb547adb\_1 \\
libbrotlidec & 1.1.0=hb547adb\_1 \\
libbrotlienc & 1.1.0=hb547adb\_1 \\
libcblas & 3.9.0=23\_osxarm64\_openblas \\
libclang & 18.1.1=pypi\_0 \\
libcurl & 8.9.1=hfd8ffcc\_0 \\
libcxx & 18.1.8=h5a72898\_4 \\
libdeflate & 1.21=h99b78c6\_0 \\
libedit & 3.1.20230828=h80987f9\_0 \\
libev & 4.33=h1a28f6b\_1 \\
libffi & 3.4.2=h3422bc3\_5 \\
libgfortran & 5.0.0=13\_2\_0\_hd922786\_3 \\
libgfortran5 & 13.2.0=hf226fd6\_3 \\
libgrpc & 1.62.2=h9c18a4f\_0 \\
libhwloc & 2.11.1=default\_h7685b71\_1000 \\
libiconv & 1.17=h0d3ecfb\_2 \\
libjpeg-turbo & 3.0.0=hb547adb\_1 \\
liblapack & 3.9.0=23\_osxarm64\_openblas \\
libnghttp2 & 1.58.0=ha4dd798\_1 \\
libopenblas & 0.3.27=openmp\_h517c56d\_1 \\
libpng & 1.6.43=h091b4b1\_0 \\
libprotobuf & 4.25.3=hbfab5d5\_0 \\
libre2-11 & 2023.09.01=h7b2c953\_2 \\
libsqlite & 3.46.0=hfb93653\_0 \\
libssh2 & 1.11.0=h7a5bd25\_0 \\
libtiff & 4.6.0=hf8409c0\_4 \\
libwebp-base & 1.4.0=h93a5062\_0 \\
libxcb & 1.16=hc9fafa5\_1 \\
libxml2 & 2.12.7=h01dff8b\_4 \\
libzlib & 1.3.1=hfb2fe0b\_1 \\
llvm-openmp & 18.1.8=hde57baf\_1 \\
markdown & 3.4.1=py39hca03da5\_0 \\
markdown-it-py & 2.2.0=py39hca03da5\_1 \\
markupsafe & 2.1.3=py39h80987f9\_0 \\
matplotlib & 3.9.2=py39hdf13c20\_0 \\
matplotlib-base & 3.9.2=py39h1398496\_0 \\
mdurl & 0.1.0=py39hca03da5\_0 \\
metis & 5.1.0=h13dd4ca\_1007 \\
ml-dtypes & 0.3.2=pypi\_0 \\
mpfr & 4.2.1=h1cfca0a\_2 \\
multipledispatch & 0.6.0=py39hca03da5\_0 \\
munkres & 1.1.4=pyh9f0ad1d\_0 \\
namex & 0.0.7=py39hca03da5\_0 \\
ncurses & 6.5=h7bae524\_1 \\
numpy & 1.26.4=py39h3b2db8e\_0 \\
numpy-base & 1.26.4=py39ha9811e2\_0 \\
openjpeg & 2.5.2=h9f1df11\_0 \\
openssl & 3.3.1=h8359307\_3 \\
opt\_einsum & 3.3.0=pyhd3eb1b0\_1 \\
optree & 0.12.1=py39h48ca7d4\_0 \\
packaging & 24.1=pyhd8ed1ab\_0 \\
pandas & 2.2.2=py39h998126f\_1 \\
pharmapy & 0.0.1=dev\_0 \\
pillow & 10.4.0=py39h3baf582\_0 \\
pip & 24.2=pyhd8ed1ab\_0 \\
protobuf & 4.25.3=py39h8472c4a\_0 \\
pthread-stubs & 0.4=h27ca646\_1001 \\
pybind11-abi & 4=hd3eb1b0\_1 \\
pygments & 2.15.1=py39hca03da5\_1 \\
pyjsparser & 2.7.1=py39hca03da5\_0 \\
pyparsing & 3.1.4=pyhd8ed1ab\_0 \\
pysocks & 1.7.1=py39hca03da5\_0 \\
python & 3.9.19=hd7ebdb9\_0\_cpython \\
python-dateutil & 2.9.0=pyhd8ed1ab\_0 \\
python-flatbuffers & 24.3.25=py39hca03da5\_0 \\
python-tzdata & 2024.1=pyhd8ed1ab\_0 \\
python\_abi & 3.9=5\_cp39 \\
pytz & 2024.1=pyhd8ed1ab\_0 \\
qhull & 2020.2=h420ef59\_5 \\
re2 & 2023.09.01=h4cba328\_2 \\
readline & 8.2=h92ec313\_1 \\
regex & 2024.7.24=py39h80987f9\_0 \\
requests & 2.32.3=py39hca03da5\_0 \\
rich & 13.7.1=py39hca03da5\_0 \\
scikit-learn & 1.5.1=py39h46d7db6\_0 \\
scipy & 1.13.1=py39hd336fd7\_0 \\
semver & 3.0.2=pypi\_0 \\
setuptools & 72.2.0=pyhd8ed1ab\_0 \\
six & 1.16.0=pyh6c4a22f\_0 \\
snappy & 1.2.1=h313beb8\_0 \\
suitesparse & 7.8.1=hf6fcff2\_0 \\
sundials & 7.1.1=h252a1ed\_0 \\
tabulate & 0.9.0=py39hca03da5\_0 \\
tbb & 2021.12.0=h420ef59\_3 \\
tensorboard & 2.16.2=pypi\_0 \\
tensorboard-data-server & 0.7.0=py39ha6e5c4f\_1 \\
tensorflow & 2.16.2=pypi\_0 \\
tensorflow-estimator & 2.17.0=cpu\_py39h9ff499c\_0 \\
tensorflow-io-gcs-filesystem & 0.37.1=pypi\_0 \\
tensorflow-macos & 2.16.2=pypi\_0 \\
tensorflow-probability & 0.24.0=pypi\_0 \\
termcolor & 2.1.0=py39hca03da5\_0 \\
tf-keras & 2.17.0=pypi\_0 \\
threadpoolctl & 3.5.0=py39h33ce5c2\_0 \\
tk & 8.6.13=h5083fa2\_1 \\
tornado & 6.4.1=py39hfea33bf\_0 \\
typing-extensions & 4.11.0=py39hca03da5\_0 \\
typing\_extensions & 4.11.0=py39hca03da5\_0 \\
tzdata & 2024a=h0c530f3\_0 \\
tzlocal & 5.2=py39hca03da5\_0 \\
unicodedata2 & 15.1.0=py39h0f82c59\_0 \\
urllib3 & 2.2.2=py39hca03da5\_0 \\
werkzeug & 3.0.3=py39hca03da5\_0 \\
wheel & 0.44.0=pyhd8ed1ab\_0 \\
wrapt & 1.14.1=py39h1a28f6b\_0 \\
xorg-libxau & 1.0.11=hb547adb\_0 \\
xorg-libxdmcp & 1.1.3=h27ca646\_0 \\
xz & 5.2.6=h57fd34a\_0 \\
zipp & 3.20.0=pyhd8ed1ab\_0 \\
zstd & 1.5.6=hb46c0d2\_0
\end{longtable}
\end{center}